%% file: ijca93rjcorr.tex
\documentstyle[11pt]{article}
\input defn

\input bghkmac


\newcommand{\worldsd}[1]{{\it worlds}_N^{d,\epsvec,#1}}
\newcommand{\nworldsd}[1]{{\it \#worlds}_N^{d,\epsvec,#1}}
\newcommand{\nworldsepsv}[1]{{\it \#worlds}_N^{\epsvec,#1}}
\newcommand{\PrNepsv}[1]{{\Pr}_N^{\epsvec,#1}}

\begin{document}
 
\title{From Statistical Knowledge Bases to Degrees of Belief%
\thanks{%
A preliminary version of this paper appeared in the
International Joint Conference on Artificial Intelligence,
1993~[BGHK93]. This version is essentially identical to one that appears
in {\em Artificial Intelligence} {\bf 87}:1--2, 1996, pp.~75--143.
Some of this work was performed
while Adam Grove was at Stanford University and at IBM Almaden Research
Center, and while Daphne Koller was at U.C.\ Berkeley and at IBM Almaden
Research Center.
This research has been supported in part by the Canadian
Government through their NSERC and IRIS programs,  by the Air Force
Office of Scientific Research (AFOSR) under Contract F49620-91-C-0080,
by an IBM Graduate Fellowship, and by a University of California
President's Postdoctoral Fellowship.  The United States Government is
authorized to reproduce and distribute reprints for governmental
purposes. }}
 
\author{
Fahiem Bacchus\\
Computer Science Dept.\\
University of Waterloo\\
Waterloo, Ontario\\
Canada, N2L 3G1\\
fbacchus@logos.uwaterloo.ca
\and
Adam J. Grove\\
NEC Research Institute\\
4 Independence Way\\
Princeton, NJ 08540\\
grove@research.nj.nec.com
\and
Joseph Y.\ Halpern\\
IBM Almaden Research Center\\
650 Harry Road\\
San Jose, CA 95120--6099\\
halpern@almaden.ibm.com
\and
Daphne Koller\\
Computer Science Department\\
Stanford University\\
Stanford, CA 94305\\
koller@cs.stanford.edu
}

\date{}
\maketitle
 
\nocite{BGHKnonmon}
 
\vspace{-4ex}
  \begin{abstract}
An intelligent agent will often be uncertain about various properties
of its environment, and when acting in that environment it will
frequently need to quantify its uncertainty. For example, if the agent
wishes to employ the expected-utility paradigm of decision theory to
guide its actions,
it
will need to assign degrees of belief
(subjective probabilities) to various assertions. Of course, these
degrees of belief should not be arbitrary, but rather should be based
on the information available to the agent.
This paper describes one
approach for inducing degrees of belief from very rich knowledge bases,
that can include information about particular individuals,
statistical correlations, physical laws, and default rules.
We call our approach
the {\em random-worlds\/} method. The
method is based on the principle of
indifference: it treats all of the worlds the agent considers possible
as being equally likely.
It
is able to integrate qualitative
default reasoning with quantitative probabilistic reasoning by
providing a language in which both types of information can be easily
expressed. Our results show that a number of desiderata that
arise in direct inference (reasoning from statistical information to
conclusions about individuals) and default reasoning follow directly
{from} the semantics of random worlds. For example, random worlds
captures important patterns of reasoning such as specificity,
inheritance, indifference to irrelevant information, and default
assumptions of independence. Furthermore, the expressive power of the
language used and the intuitive semantics of
random worlds
allow the method to deal
with problems that are beyond the scope
of many other non-deductive reasoning systems.
  \end{abstract}

\section{Introduction}
Consider an agent with a knowledge base, $\KB$,
who has to make decisions about
its actions in the world.
For example, a doctor may need to decide on a
treatment for a particular patient, say Eric.
The doctor's knowledge
base might contain information of different types, including:
statistical information, e.g., ``80\% of patients
with jaundice have hepatitis''; first-order information,
e.g., ``all patients with hepatitis have jaundice''; default information,
e.g., ``patients with hepatitis typically have a fever''; and
information about the particular patient at hand, e.g., ``Eric has
jaundice''.
In most cases, the knowledge base will
not contain complete information about a particular individual.
For example, the doctor may be uncertain
about the exact disease that Eric has.  Since
the efficacy of a treatment will almost certainly depend on the disease,
it is important for the doctor to be able to quantify the relative
likelihood of various possibilities.
More generally, to apply standard tools for decision making such
as {\em decision theory} (see, \eg \cite{LuceR,Savage}),
an agent must assign
probabilities, or {\em degrees of belief}, to
various events.
For example, the doctor may wish to assign a
degree of belief to an event such as ``Eric has
hepatitis''.
This paper describes one particular method that allows such
an agent
to use its knowledge base to assign degrees of belief in a
principled manner; we call this method the {\em random-worlds method}.

There has been a great deal of work addressing aspects of this
general problem.  Two large bodies of work that are particularly relevant
are the work on {\em direct inference\/}, going back to Reichenbach
\cite{Reichenbach:Theory.Of.Probability},
and the various approaches to {\em
nonmonotonic
reasoning}.  Direct inference deals with the problem of
deriving degrees of belief from statistical information, typically by
attempting to find
a suitable {\em reference class\/} whose statistics
can be used
to determine the degree of belief.
For instance, a suitable reference class for the patient Eric might
be the class of all patients with jaundice.
While direct inference is concerned with statistical knowledge,
the field of nonmonotonic reasoning,
on the other hand,
deals mostly with knowledge bases that
contain default rules.
As we shall argue,
none of the systems proposed for either reference-class reasoning
or nonmonotonic reasoning
can deal adequately with the large
and complex knowledge bases we are interested in.
In particular,
none can handle rich knowledge bases
that may contain first-order,
default, and statistical information.
The random-worlds approach, on the other hand, can deal with such complex
knowledge bases, and handles
several paradigmatic problems in both nonmonotonic and reference-class
reasoning remarkably well.
 
We now provide a brief overview of the random-worlds approach.
We assume that the information in the knowledge base is expressed in
a variant of the language
introduced by Bacchus \cite{Bacchus}.  Bacchus's language augments
first-order logic by allowing statements of
the form
$\cprop{\Hep(x)}{\Jaun(x)}{x} = 0.8$, which says that 80\% of patients
with jaundice have hepatitis.  Notice,
however, that in finite models this
statement has the (probably unintended) consequence that the number
of patients with jaundice is a multiple of 5.  To avoid this problem,
we use approximate equality rather than equality, writing
$\cprop{\Hep(x)}{\Jaun(x)}{x} \aeq 0.8$, read ``approximately 80\% of
patients with jaundice have hepatitis''.
Intuitively, this says that the proportion of jaundiced patients with
hepatitis is
close to 80\%: \ie within some tolerance
$\eps$ of 0.8.

Not only does the use of approximate equality solve the problem of
unintended consequences, it has another significant advantage:
it lets us express default information.  We interpret a statement
such as ``Birds typically fly'' as expressing the statistical
assertion
that ``Almost all birds fly''.  Using approximate
equality, we can represent this as
$\cprop{\Fly(x)}{\Bird(x)}{x} \aeq 1$.
This interpretation is closely related to various approaches
applying probabilistic semantics to nonmonotonic logic; see
Pearl~\cite{Pearl90} for an overview
of these approaches, and Section~\ref{GMPsec} for further discussion.

Having described the language in which our knowledge base is
expressed, we now need to decide how to assign degrees of belief given
a knowledge base.
Perhaps the most widely used framework for assigning degrees
of belief (which are essentially subjective probabilities) is the
Bayesian paradigm.  There, one assumes a space of possibilities and a
probability distribution over this space (the {\em prior\/}
distribution), and calculates {\em posterior\/} probabilities by
conditioning on what is known (in our case, the knowledge base).
{T}o use this approach, we must
specify the space of possibilities and the distribution over it.
In Bayesian reasoning, there is relatively little
consensus as to how this should be done in general.
Indeed, the usual philosophy is that these decisions are
subjective.  The difficulty of making these decisions
seems to have been an important reason for
the historic unpopularity of the Bayesian approach in symbolic AI \cite{MH}.
 
Our approach is different.  We assume that the $\KB$
contains all the knowledge the agent has, and we allow a very
expressive language so as to make this assumption reasonable.
This assumption means that any knowledge the agent has that could
influence the prior distribution is already included in the $\KB$.
As a consequence, we give a single uniform construction of a
space of possibilities and a distribution over it.
Once we have this
probability space, we can use the Bayesian approach:
{T}o compute the
probability of an assertion $\phi$ given $\KB$, we condition on $\KB$, and
then compute the probability of $\phi$ using the resulting posterior
distribution.
 
So how do we choose the probability space?
One general strategy, discussed by Halpern \cite{Hal4}, is to give
semantics to degrees of belief in terms of a probability distribution over
a set of {\em possible worlds}, or first-order models.
This semantics clarifies the distinction between statistical
assertions and degrees of belief.
As we suggested above, a statistical assertion such as
$\cprop{\Hep(x)}{\Jaun(x)}{x} \aeq 0.8$ is true or false in
a particular world, depending on how many jaundiced patients have
hepatitis in that world.
On the other hand, a degree of
belief
is neither true nor false in a particular world---it has
semantics only with respect to the entire set of possible worlds and a
probability distribution over them.  There is no necessary connection
between the information in the agent's $\KB$ and the distribution
over worlds that determines her degrees of belief.
However, we clearly want
there to be some connection. In particular,
we want the agent to base her degrees of beliefs on her information
about the world, including her statistical information.
As this paper shows, the random-worlds method is a powerful technique for
accomplishing this.
 
{T}o define our probability space, we have to choose an
appropriate set of possible worlds.
Given some domain of individuals, we stipulate that
the set of worlds is simply the set of all first-order models over
this domain.  That is, a possible world corresponds to a particular
way of interpreting the symbols in the agent's vocabulary over the domain.
In our context, we can assume that the ``true world'' has a finite domain,
say of size $N$.
In fact, without loss of generality, we assume that the domain is
$\{1,\ldots,N\}$.
 
Having defined the probability space (the set of possible worlds), we
must construct a probability distribution over this set. For this, we
give perhaps the simplest possible definition: we assume that all the
possible worlds are equally likely (that is, each world has the same
probability). This can be viewed as an application of the {\em
principle of indifference}.  Since
we are assuming that all the agent knows is incorporated in her
knowledge base, the agent has no {\em a priori\/} reason to prefer one
world over the other. It is therefore reasonable to view all worlds as
equally likely.
Interestingly, the principle of indifference (sometimes also called
the {\em principle of insufficient reason\/}) was originally promoted
as part of the very definition of probability when the field was
originally formalized by Jacob Bernoulli and others; the principle was
later popularized further and applied with considerable success by
Laplace.  (See \cite{Hacking2} for a historical discussion.)  It
later fell into disrepute as a general definition of probability,
largely because of the existence of paradoxes that arise when the
principle is applied to infinite
or continuous probability spaces.
We claim, however, that
the principle of indifference can be a natural and effective way of
assigning degrees of belief in certain contexts, and in particular,
in the context where we restrict our attention to a finite collection
of worlds.
 
Combining our choice of possible worlds with the principle of
indifference, we obtain our prior distribution.  We can now induce a
degree of belief in $\phi$ given $\KB$ by conditioning on $\KB$ to
obtain a posterior distribution and then computing the probability of
$\phi$ according to this new distribution.  It is easy to see that,
since each world is equally likely,
the
degree of belief in $\phi$ given $\KB$ is the fraction of possible
worlds satisfying $\KB$ that also satisfy $\phi$.

One problem with the approach as stated so far is that, in general, we
do not know the domain size $N$.  Typically, however, $N$ is known to
be large.  We therefore approximate the degree of belief for the true
but unknown $N$ by computing the
limiting value of this degree of belief as $N$
grows large.  The result is our random-worlds method.
 
The key ideas in the approach are not new.  Many of them can be found
in the work of Johnson \cite{Johnson} and Carnap
\cite{Carnap,Carnap2}, although these authors focus on knowledge bases
that contain only first-order information, and for the most part
restrict their attention to unary predicates.
Related approaches have been used in the more recent works of
Shastri~\cite{Shastri} and of Paris and Vencovska~\cite{PV}, in the
context of a unary statistical language.  Chuaqui's recent work
\cite{chuaqui} is also relevant.  His work, although technically quite
different from ours, shares the idea of basing a %
theory of
probabilistic reasoning upon notions of indifference and symmetry.
The works of Chuaqui and
Carnap investigate very different issues from those we examine in this
paper.  For example, Carnap, and others who later continued to develop
his ideas, were very much interested in inductive learning
(especially the problem of learning universal laws).
While we believe the question of learning is very
important (see Section~\ref{learning}), we have largely concentrated
on understanding (and generalizing) the process of going from
statistical information and default rules to inferences about
particular individuals.  Many of the new results we describe reflect
this different emphasis.
 
Having defined the method, how do we judge its reasonableness?
Fortunately, as we mentioned, there are two large bodies of work on
related problems
{from} which we can draw guidance: reference-class
reasoning and default reasoning.  While none of the solutions
suggested for these problems seems entirely adequate, the years
of research have resulted in some strong intuitions regarding what
answers are intuitively reasonable for certain types of queries.
Interestingly, these intuitions often lead to identical desiderata.
In particular, most systems (of both types) espouse some form of
preference for more specific information and the ability to ignore
irrelevant information.
We show that the random-worlds approach
satisfies these desiderata.
In fact, in the case of random worlds, these properties follow from two
much general theorems.  We prove that, in those cases where there is
a specific piece of statistical information that should ``obviously''
be used to determine a degree of belief, random worlds does in fact
use this information.  The different desiderata, such as a preference
for more specific information and an indifference to irrelevant
information follow as easy corollaries.
We also show that random worlds provides reasonable answers in many other
contexts, not covered by the standard specificity and irrelevance
heuristics.
Thus, the random-worlds method is indeed a powerful
one, that can deal with rich knowledge bases and still produce
the answers that people have identified as being the most appropriate
ones.
 
The rest of this paper is organized as follows.  In the next two
sections, we outline some of the major themes and problems in the work
on reference classes and on default reasoning.  Since one of our major
claims is that the random-worlds approach solves many of these
problems, this will help set our work in context.  In
Section~\ref{formalism}, we describe the random-worlds method in
detail.  In Section~\ref{results}, we state and prove a number of
general theorems about the properties of the approach, and show
how various desiderata follow from these theorems.
In Section~\ref{GMPsec} we discuss the
problem of calculating degrees of belief.  Using results from
\cite{GHK2}, we demonstrate a close connection between random worlds and
{\em maximum entropy\/} in the case of unary knowledge bases.  Based
on this connection, we show that in
many cases of interest a maximum-entropy computation can be used to
calculate an agent's degree of belief.  Furthermore, we show that the
maximum-entropy approach to default reasoning considered in \cite{GMP}
can be embedded in our framework.  Finally, we discuss some possible
criticisms
and limitations
of the random-worlds method in Section~\ref{discussion} and
the possible impact of the method in Section~\ref{conclusions}.

\section{Reference classes}
\label{refclass}
Strictly speaking, the only necessary relationship between objective
knowledge about frequencies and proportions on the one hand and
degrees of belief on the other hand is the simple mathematical fact
that they both obey the axioms of probability. But in practice we
usually hope for a deeper connection: the latter should be based on
the former in some ``reasonable'' way.  Of course, the random-worlds
approach that we are advocating is precisely a theory of how this
connection can be made.
But our approach is far from the first to attempt to connect
statistical information and degrees of belief.  Most of the earlier
work is based on the idea of finding a suitable {\em reference class}.
In this section, we review some of this work and show why we believe
that this approach, while it has some intuitively reasonable
properties, is inadequate as a general methodology.  (See also
\cite{BGHKci} for further discussion of this issue.)  We go into some
detail here, since the issues that arise provide some motivation for
the results that we prove later regarding our approach.
 
\subsection{The basic approach}
The earliest sophisticated attempt at clarifying the connection between
objective statistical knowledge and degrees of belief, and the basis for
most subsequent proposals, is due to Reichenbach
\cite{Reichenbach:Theory.Of.Probability}. Reichenbach describes the idea as
follows:
  \begin{quotation}
    ``If we are asked to find the probability holding for an individual future
    event, we must first incorporate the case in a suitable {\em reference
    class\/}. An individual thing
    or event may be incorporated in many reference
    classes\ldots{} . We then proceed by considering the narrowest
    [smallest]
    reference
    class for which suitable statistics can be compiled.''
  \end{quotation}
Although not stated explicitly in this quote, Reichenbach's approach was to
equate the degree of belief in the individual event with the
statistics from the chosen reference class. As an example, suppose
that we want to determine a probability (\ie a degree
of belief) that Eric, a particular patient with jaundice, has the disease
hepatitis. The particular individual Eric is a member of the class of all
patients with jaundice.  Hence, following Reichenbach, we can use the class
of all such patients as a reference class, and assign a degree of belief
equal to our statistics concerning the frequency of hepatitis among this
class. If we know that this frequency is 80\%, then we would assign a
degree of belief of $0.8$ to the assertion that Eric has hepatitis.
 
Reichenbach's approach consists of (1)
the postulate that we use the statistics from a particular reference
class to infer a degree of belief with the same numerical value, and
(2) some guidance as to how to choose this reference class from
a number of competing reference classes.
We consider each point in turn.
 
In general, a {\em reference class\/} is simply a set of
domain individuals%
\footnote{These ``individuals''
might be complex objects
(such as sequences of coin tosses) depending
on what we take as primitive in our ontology.}
that contains the particular individual about whom we wish to reason
and for which we have
``suitable statistics''.
In our framework, we may take
the set of individuals satisfying a formula
$\psi(x)$ to be a reference class.
The requirement that the particular
individual $c$ we wish to reason about belongs to the class is
then represented
by the logical assertion $\psi(c)$.%
\footnote{Although the examples in this section deal with reasoning
about single individuals, in general both reference-class reasoning
and random worlds can be applied to queries such as ``Did Eric infect
Tom'', which involve reasoning about a number of individuals
simultaneously. In such cases the reference classes will consist of
sets of {\em tuples\/} of individuals.}
But what does the phrase ``suitable
statistics'' mean?
Suppose for now we take a ``suitable statistic'' to be a closed
interval that is nontrivial, \ie that is not $[0,1]$, in which the
proportion or frequency lies.
More precisely, consider some query $\phi(c)$, where $\phi$ is some logical
assertion and $c$ is a constant, denoting some individual in the
domain. Then, under this interpretation,
$\psi(x)$ is a reference class for this query if
we know both $\psi(c)$ and $\cprop{\phi(x)}{\psi(x)}{x} \in
[\alpha,\beta]$, for some nontrivial interval $[\alpha,\beta]$.  That
is, we know that $c$ has property $\psi$, and that among the class of
individuals that possess property $\psi$, the proportion that also
have property $\phi$ is between $\alpha$ and $\beta$.
If we decide that this is the appropriate reference class then,
using Reichenbach's approach, we would
conclude $\beliefprob{\phi(c)} \in [\alpha, \beta]$, i.e., the
probability (degree of belief) that $c$ has property $\phi$ is between
$\alpha$ and $\beta$. Note that the appropriate reference class
for the query $\phi(c)$ depends both on
the formula $\phi(x)$ and on the individual $c$.
 
Given a query $\phi(c)$, there will in general be many
reference classes that are arguably appropriate for it.
For example, suppose we know both $\psi_1(c)$ and $\psi_2(c)$,
and we have two pieces of statistical information:
$\cprop{\phi(x)}{\psi_1(x)}{x} \in [\alpha_1,\beta_1]$ and
$\cprop{\phi(x)}{\psi_2(x)}{x} \in [\alpha_2,\beta_2]$.
In this case both $\psi_1(x)$ and $\psi_2(x)$ are reference classes for
$\phi(c)$  and, depending on the values of the $\alpha$'s and $\beta$'s,
they could assign conflicting degrees of belief to $\phi(c)$. The
second part of Reichenbach's approach is intended to deal with the
problem of how to choose a single reference class from a set of possible
classes. Reichenbach recommended preferring the
narrowest (\ie the smallest, or most specific) class.
In this example, if we know $\forall x \, (\psi_1(x) \rimp
\psi_2(x))$,
so
that the class $\psi_1(x)$ is a subset of the class $\psi_2(x)$, then,
using Reichenbach's approach, we would
take the statistics from the more specific reference class
$\psi_1(x)$ and conclude that $\beliefprob{\phi(c)} \in
[\alpha_1,\beta_1]$.

These two parts of Reichenbach's approach---using statistics taken from
a class as a degree of belief about an individual and preferring
statistics from more specific
classes---are generally reasonable and
intuitively compelling when applied to
simple examples. Of course, even on the simplest examples
Reichenbach's strategy cannot be said to be ``correct'' in any absolute
sense. Nevertheless, it is impressive that
there is such widespread agreement as to the
reasonableness of the answers. As we show later,
the random-worlds approach agrees with both aspects of Reichenbach's
approach when applied to simple (and
uncontroversial) examples. Unlike
that approach, however, the random-worlds approach derives these
intuitive answers from more basic principles.
As a result, it is able to deal well with more complex examples that
defeat Reichenbach's approach.
 
Despite its successes,
Reichenbach's approach has
several serious problems.
For one thing, defining what counts as a ``suitable statistic'' is
not easy.  For another, it is clear that the principle of preferring
more specific information
rarely suffices to deal with
the cases that arise with a rich knowledge base.
Nevertheless,
much of the work on connecting statistical information and degrees of
belief, including that of Kyburg
\cite{Kyburg:Reference.Class,Kyburg:Statistical.Inference} and of Pollock
\cite{Pollock:Induction}, has built on Reichenbach's ideas of reference
classes
by elaborating the manner in which choices are made between reference classes.
As a result, these later approaches all suffer from a similar set of
difficulties,
which we now discuss.
 
\subsection{Identifying reference classes}
\label{identify.ref}
 
Recall that we took a reference class to be simply a set for which we
have ``suitable statistics''.  But if any set of
individuals whatsoever can potentially serve as a reference class
then problems arise.
Assume we know $\Jaun(\Eric)$ and
$\cprop{\Hep(x)}{\Jaun(x)}{x} \aeq 0.8$.  In this case $\Jaun(x)$ is a
legitimate reference class for the query $\Hep(\Eric)$.  Therefore, we
would like to conclude that $\beliefprob{\Hep(\Eric)} = 0.8$.  But
$\Eric$ is also a member of the {\em more specific\/} class of
jaundiced patients without hepatitis together with $\{\Eric\}$ (\ie
the class defined by the formula $(\Jaun(x) \land \neg \Hep(x)) \lor x
= \Eric$).
If there are quite a few jaundiced patients without hepatitis, then
we have excellent statistics for
the proportion of patients in this class with hepatitis: it is
approximately 0\%.  Thus, the conclusion that
$\beliefprob{\Hep(\Eric)} = 0.8$ is disallowed by
the rule instructing us to use the most specific reference class.
In fact, it seems that we can almost always find a more specific class
that will give a different and intuitively incorrect answer.
This example suggests that
we cannot take an arbitrary set of individuals to be a
reference class; it must satisfy additional criteria.

Kyburg and Pollock deal with this difficulty
by placing restrictions on the set of allowable reference classes
that, although different, have the effect
of disallowing {\em disjunctive reference classes}, including
the problematic class described above.
This approach suffers from
two deficiencies.
First, as Kyburg himself has observed
\cite{Kyburg:Statistical.Inference}, these restrictions do
not eliminate the problem completely.
Furthermore, restricting the set of allowable reference
classes may prevent us from making full use of the information we have.
For example, the genetically
inherited disease Tay-Sachs (represented by the predicate $\TS$)
appears only in babies of two distinct populations: Jews of
east-European extraction ($\EEJ$), and French-Canadians from a certain
geographic area ($\FC$). Within
the afflicted population, Tay-Sachs occurs in $2\%$ of the babies.  The
agent might represent this fact using the statement
$\cprop{\TS(x)}{\EEJ(x) \lor \FC(x)}{x} = 0.02$.
However, if disjunctive reference classes are disallowed, then the
agent would not be able to use this information in reasoning.
 
It is clear that if one takes the reference-class approach to generating
degrees of belief, some restrictions on what constitutes a legitimate
reference class are inevitable.
Unfortunately, it seems that the current approaches to this problem are
inadequate. The random-worlds
approach does not depend on the notion of a reference class,
and so is not forced to confront this issue.

\subsection{Competing reference classes}
\label{competing}
 
Even if the problem of defining the set of ``legitimate'' reference
classes can be resolved, the reference-class approach must still
address the problem of choosing the ``right''
class out of the set of legitimate ones.
The solution to this problem has typically been to posit a collection of
rules indicating when one reference class should be preferred over
another.
The basic criterion is the one we already mentioned:~choose the most
specific
class.  But even in the cases to which this
specificity rule applies, it is not always appropriate.  Assume, for
example, that we know that between 70\% and 80\% of birds chirp
and that between 0\% and 99\% of magpies chirp.
If Tweety is a magpie, the specificity
rule would tell us to use the more specific reference class, and
conclude that $\beliefprob{\Chirps(\Tweety)} \in [0,0.99]$.  Although the
interval $[0,0.99]$ is certainly not trivial, it is not very
meaningful.  Had the $0.99$ been a 1, the interval would have been
trivial,
and we could have then ignored this
class and used the
more detailed statistics of $[0.7,0.8]$ derived from the class of
birds.
 
The knowledge base above might be appropriate for someone who knows
little about magpies, and so feels less confidence in his statistics
for magpies than in his statistics for the class of birds as a whole.
But since $[0.7,0.8] \subseteq [0,0.99]$, we know nothing that
indicates that magpies are actually different from birds in general
with respect to chirping.  There is an alternative intuition that says
that if the statistics for the less specific reference class (the
class of birds) are more precise, and they do not contradict the
statistics for the more specific class (magpies), then we should use
them. That is, we should conclude that $\beliefprob{\Chirps(\Tweety)}
\in [0.7,0.8]$.  This intuition is captured and generalized in
Kyburg's {\em strength rule}.
 
Unfortunately, neither the specificity rule nor its extension by
Kyburg's strength rule are adequate in most cases.  In typical
examples, the agent generally has several incomparable classes
relevant to the problem, so that neither rule applies.
Reference-class systems such as Kyburg's and Pollock's simply give no
useful answer in these cases.  For example, suppose we know that Fred
has high cholesterol and is a heavy smoker, and that 15\% of people
with high cholesterol get heart disease.  If this is the only suitable
reference class, then (according to all the systems)
$\beliefprob{\HeartDisease(\Fred)} = 0.15$.  On the other hand,
suppose we then acquire the additional information that 9\% of heavy
smokers develop heart disease (but still have no nontrivial
statistical information about the
class of people with
both attributes).  In this case, neither class is the single right
reference class, so approaches that rely on finding a single reference
class generate a trivial range for the degree of belief that Fred will
contract
heart disease in this case. For example, Kyburg's system will generate
the interval $[0,1]$ for the degree of belief.
 
Giving up completely in the face of conflicting evidence seems to us
to be inappropriate. The entire enterprise of generating degrees of
belief is geared to providing the agent with some guidance for its
actions (in the form of degrees of belief) when deduction is
insufficient to provide a definite answer. That is, the aim is to
generate {\em plausible\/} inferences. The presence of conflicting
information does not mean that the agent no longer needs guidance.
When we have several competing reference classes, none of which
dominates the others according to specificity or any other rule that
has been proposed, then the degree of belief should most reasonably be
some {\em combination\/} of the corresponding statistical values.  As
we show later, the random-worlds approach does indeed combine the
values from conflicting reference classes in a reasonable way, giving
well-motivated answers even when the reference-class approach would
fail.

\subsection{Other types of information}
We have already pointed out the problems that arise with the
reference-class approach if more than one reference class bears on a
particular problem.  A more subtle problem is encountered in cases where
there is relevant information that is not in the form of a reference
class.  We have said that for $\psi(x)$ to be a reference class for a
query about $\phi(c)$ we must know $\psi(c)$ and have some statistical
information about $\cprop{\phi(x)}{\psi(x)}{x}$. However, it is not
sufficient to consider only the query $\phi(c)$.  Suppose we also know
$\phi(c) \dimp \sigma(c)$ for some other formula $\sigma$. Then we would
want $\beliefprob{\phi(c)} = \beliefprob{\sigma(c)}$.  But this implies
that all of the reference classes for $\sigma(c)$ are relevant as well,
because anything we can infer about $\beliefprob{\sigma(c)}$ tells us
something about $\beliefprob{\phi(c)}$.  Both Pollock
\cite{Pollock:Induction} and Kyburg \cite{Kyburg:Reference.Class} deal
with this by considering all of the reference classes for any formula
$\sigma$ such that $\sigma(c) \dimp \phi(c)$ is known. However, they do
not consider the case where it is known that $\sigma(c) \rimp \phi(c)$,
which implies that $\beliefprob{\sigma(c)} \leq \beliefprob{\phi(c)}$,
nor the case where it is known that $\phi(c) \rimp \sigma(c)$, which
implies that $\beliefprob{\sigma(c)} \geq \beliefprob{\phi(c)}$.  Thus,
if we have a rich theory about $\phi(c)$ and its implications, it can
become very hard to locate all of the possible reference classes or even
to define what qualifies as a possible reference class.
 
\subsection{Discussion}

A comparison between random worlds and reference-class approaches
can be made in terms of the use of local versus global
information. The reference-class approach is predicated on the
assumption that
we can always focus on
a
single
piece of information,
the statistics over a single reference class,
that summarizes all the relevant information in the knowledge base.
A strategy based on identifying a single relevant (``local'') datum
can offer great efficiency, but of course we should not expect this
to be a general substitute for the use of all the (``global'') information
we have available.
In this sense, the difficulties encountered by the reference-class
approach are not surprising.  When generating degrees of belief from a
rich knowledge base, it will not always be possible to find a single
reference class that captures all of the relevant information.
 
It is important to remember that although the notion of a
reference class seems intuitive, it arises as part
of one proposed {\em solution\/} strategy for the problem of computing
degrees of belief.  The notion of a reference classes is not part of
the description of the problem, and there is
no reason for it to necessarily be part of the solution.  Indeed, as we
have tried to argue, making it part of the solution can lead to
more problems than it solves.
 
Our approach
makes no attempt to locate a single local
piece of information (a reference class). Thus, all of the problems
described above that arise from trying locate the ``right'' reference class
vanish.  Rather, it uses a semantic
construction that takes into account all of the information in the
knowledge base in a uniform manner.
As we shall see,
the random-worlds
approach generates answers that agree with the reference-class
approach in those special cases where there is a single
appropriate reference class.
However, it continues to give reasonable answers in many
situations where no
single local piece of information suffices.
Furthermore, these answers are obtained directly from the
simple semantics of
random worlds, with no {\em ad hoc\/} rules and assumptions.
 
\section{Default reasoning}
\label{nonmon}
 
One main claim of this paper is that the random-worlds method of inference,
coupled with our statistical interpretation of defaults,
provides a well-motivated
and successful system of default reasoning.
Evaluating such a claim is hard because there are many,
sometimes rather vague,
criteria for success that one can consider.
In particular, not all criteria are appropriate for all default reasoning
systems:  Different applications (such as some of the ones outlined in
\cite{McCarthy:Applications.of.Circumscription}) require different
interpretations for a default rule, and therefore need to satisfy
different desiderata.
Nevertheless, there are certain desiderata that have gained
acceptance as measures for the success of a new nonmonotonic reasoning
system.
Some are general properties of nonmonotonic inference (see Section~\ref{KLM}).
Most, on the other hand, involve getting the ``right'' answers to
a small set of standard examples (more often than not involving a bird
called ``Tweety'').  As we claim at the end of this section, this has made
an ``objective'' validation of proposed systems difficult, to say the
least.
In this section, we survey some of the desired properties for default
reasoning and the associated problems and issues.  Of course, our survey
cannot be comprehensive.  The areas we consider are  the semantics of
defaults, basic properties of default inference, inheritance and
irrelevance,
expressive power, and the lottery paradox.
 
\subsection{Semantics of defaults}
It is possible to discuss some properties of default
reasoning systems in an extremely abstract fashion
(see Section~\ref{KLM}), but for other properties we need to make
some assumptions about the type of system being considered.
In particular, we consider systems that incorporate some notion of
a {\em default rule}, which we now explain.
In general, a default rule is an expression that has the form $A(x)
\default B(x)$, whose intuitive interpretation is that if $A$ holds for some
individual $x$ then typically (normally, usually, probably, etc.) $B$
holds for that individual.%
\footnote{We use $\default$ for a default implication, reserving
$\rimp$ for standard material implication.}
While the syntax actually used differs
significantly from case to case, most default reasoning systems have some
construct of this type. For instance, in Reiter's
{\em default logic\/} \cite{reiter} we would
write
$$\frac{A(x)\ :\ B(x)}{B(x)} $$
while in a {\em circumscriptive\/} framework \cite{McC}, we might use
$$ \forall x \, (A(x) \land \neg \Ab(x) \rimp  B(x))$$
while circumscribing $\Ab(x)$.
Theories based on first-order conditional
logic \cite{Delgrande:Conditional.Logic.Revised}
often do use the syntax $A(x) \default B(x)$.
As we said in the introduction,
in the random worlds framework this default is captured using
the statistical assertion $\cprop{B(x)}{A(x)}{x} \aeq 1$.

While most systems of default inference have a notion of a default rule, not
all of them address the issue of what the rule {\em means}.  In particular,
while all systems describe how a default rule should be used, some do not
ascribe semantics (or ascribe only unintuitive semantics) to such rules.
Without a good, intuitive semantics for defaults it becomes
very difficult to judge the reasonableness of a collection
of default statements.
For example, as we mentioned above, one standard reading
of $\phi \default \psi$ is ``$\phi$'s are typically $\psi$'s''.
Under this reading, the pair of defaults $A \default B$ and
$A \default \neg B$
should be inconsistent.
In approaches such as Reiter's default logic,
$A \default B$ and $A \default \neg B$ can be simultaneously adopted;
they are not ``contradictory'' because there is no relevant notion
of contradiction.
In contrast, our approach does give semantics to defaults.
In fact, we use a single logic and semantics that covers first-order
information, default information, and statistical information.
Such an approach enables us, among other things,
to verify the consistency  of a collection of
defaults and to see whether a default follows logically from a
collection of defaults. Of other existing theories, those based on
conditional
or modal logic come closest to achieving this
(see \cite{Boutilier:Phd.Thesis} for
further discussion of this point).

\subsection{Properties of default inference}\label{KLM}
As we said, default reasoning systems have typically been measured by
testing them on a number of important examples.  Recently, a few tools have
been developed that improve upon this approach. Gabbay
\cite{Gabbay:nonmon.inference} (and later Makinson
\cite{Makinson:nonmon.inference} and Kraus, Lehmann, and Magidor
\cite{KLM}) introduced the idea of investigating the input/output relation
of a default reasoning system, with respect to certain general properties
that such an inference relation might possess. Makinson
\cite{Makinson:survey} gives a detailed survey of this work.
 
The idea is simple. Fix a theory of default reasoning
and let $\KB$ be some knowledge base appropriate to this theory.
Suppose $\phi$ is a default conclusion reached from
$\KB$ according to the particular default approach being considered.
In this case, we write $\KB \dentails \phi$.
The relation $\dentails$ clearly depends on the default theory being
considered.  It is necessary to assume in this context that
$\KB$ and $\phi$ are both expressed in the same logical
language,
and that the language
has a notion of valid implication.
Thus, for example, if we are considering default logic or
$\epsilon$-semantics, we must assume that the defaults are fixed (and
incorporated
into the notion of $\dentails$) and that both $\KB$ and
$\phi$ are first-order or propositional formulas.  Similarly, in the case of
circumscription, the circumscriptive policy must also be fixed and
incorporated into $\dentails$. (See also the discussion at the beginning
of Section~\ref{nonmon.inheritance}.)

With this machinery we can state a few desirable properties of
default theories in a way that is independent of the (very diverse)
details of such theories.
There are five properties of $\dentails$
that have been viewed as being particularly desirable \cite{KLM}:
  \begin{itemize}
    \item{\em Right Weakening.} If $\phi \rimp \psi$ is logically valid and
          $\KB \dentails \phi$, then $\KB \dentails \psi$.
    \item {\em Reflexivity.} $\KB \dentails \KB$.
    \item {\em Left Logical Equivalence.} If $\KB \dimp \KBp$ is
          logically valid, then $\KB \dentails \phi$ if and only if $\KBp \dentails
          \phi$.
    \item {\em Cut.} If $\KB \dentails \theta$ and $\KB \land \theta \dentails
          \phi$ then $\KB \dentails \phi$.
    \item {\em Cautious Monotonicity.} If $\KB \dentails \theta$ and $\KB
          \dentails \phi$ then $\KB \land \theta \dentails \phi$.
  \end{itemize}
While it is beyond the scope of this paper to defend these criteria
(see \cite{KLM}), we do want to stress
{\em Cut\/} and {\em Cautious Monotonicity},
since they will be useful in our later results.  They tell us that we
can safely add to $\KB$ any conclusion $\theta$ that we can derive from
$\KB$, where ``safely'' is interpreted to mean that the set of conclusions
derivable (via $\dentails$) from $\KB \land \theta$ is precisely
the same as that derivable from $\KB$ alone.
 
As shown in \cite{KLM}, numerous other conditions can be derived
{from} these properties. For example, we can prove:
\begin{itemize}
\item{\em And.} If $\KB\dentails \phi$ and $\KB\dentails \psi$ then
  $\KB \dentails \phi \land \psi$.
\end{itemize}
Other plausible
properties, however,
do not follow from these basic five. For example, the following
property captures reasoning by cases:
\begin{itemize}
\item {\em Or.} If $\KB \dentails \phi$ and $\KBp \dentails \phi$,
  then $\KB \lor \KBp \dentails \phi$.
\end{itemize}
 
Perhaps the most interesting property that does not follow from the
basic five properties is what has been called {\em Rational
  Monotonicity\/} \cite{KLM}.  Note that the property of (full)
monotonicity, which we do not want, says that $\KB \dentails \phi$
implies $\KB \land \theta \dentails \phi$, no matter what $\theta$ is.
It has been argued
that default reasoning should satisfy the same property in those cases
where $\theta$ is ``irrelevant'' to the connection between $\KB$ and
$\phi$.  While it is difficult to characterize ``irrelevance'',
one situation where we may believe that $\theta$
should not affect the conclusions we can derive from $\KB$ is if
$\theta$ is not implausible given $\KB$, \ie if it is not the case
that $\KB \dentails \neg \theta$ (see Section~\ref{nonmon.inheritance}
for an example). The following property asserts that monotonicity
holds when adding such a formula $\theta$ to our knowledge base:
\begin{itemize}
\item {\em Rational Monotonicity.} If $\KB \dentails \phi$ and it is
  not the case that $\KB \dentails \neg \theta$, then $\KB \land
  \theta \dentails \phi$.
\end{itemize}

{\em Rational Monotonicity\/} is a fairly
strong property, and is certainly not universally
agreed upon (see \cite{Makinson:survey} for a discussion, and some
weakened versions).  However,
several people, notably Lehmann and Magidor \cite{LehMag}, have argued
strongly for the desirability of this principle. One advantage of {\em
Rational Monotonicity\/} is that it covers some fairly
noncontroversial patterns of reasoning involving property inheritance.
We explore this further in the next section.
As is demonstrated in Section~\ref{KLMresults}, our
approach satisfies a slightly weakened version of
 {\em Rational Monotonicity\/}.

The set of properties we have discussed provides a simple, but useful,
system for classifying default theories. There are certainly
applications in which some of the properties are inappropriate;
Reiter's default logic is still popular even though it does not
satisfy {\em Cautious Monotonicity}, {\em Or},
or {\em Rational Monotonicity}
\cite{Makinson:survey}.  (We briefly discuss one of the consequent
disadvantages of default logic in the next section.) Nevertheless,
many people would argue
that the five core properties given above
constitute a reasonable, if incomplete, set of desiderata for
mainstream default theories.
 
\subsection{Specificity and inheritance}
\label{nonmon.inheritance}
As we have pointed out, systems of default reasoning have particular
mechanisms for expressing default rules. A collection of such rules
(perhaps in conjunction with other information)
forms a default theory (or default knowledge base).
For example, a particular default theory $\KBdef$ might contain the
default ``$A$'s are typically $B$'s''; we denote this by writing
$(A(x) \default B(x)) \in \KBdef$. A default theory $\KBdef$ is used
by a default reasoning system in order to reason from various premises
to default conclusions.
For example, a theory $\KBdef$ containing the above default might infer
$B(c)$ from $A(c)$.  Let $\dentailssub{def}$
indicate the input/output relationship generated by a
particular
default
reasoning system that uses $\KBdef$.  Thus,
$A(c) \dentailssub{def} B(c)$ indicates
that
this
default reasoning system is able to
conclude $B(c)$ from the premise $A(c)$ using the
default theory $\KBdef$.
In this
section we examine some additional properties
we might like $\dentailssub{def}$ to
satisfy.
 
Clearly, the presence of a default rule in a theory does not
necessarily mean that the associated default reasoning system will (or should)
apply that rule to any particular individual.
Nevertheless, unless
something special is known about that individual, the following seems
to be an obvious requirement for any default reasoning system:
  \begin{itemize}
    \item {\em Direct Inference for Defaults.}
          If
          $(A(x) \default B(x)) \in \KBdef\,$ and $\KBdef\,$ contains no
          assertions mentioning $c$, then $A(c) \dentailssub{def} B(c)$.
  \end{itemize}
This requirement has been previously discussed by Poole \cite{Poole91},
who called it the property of {\em Conditioning}. We have chosen a
different name that relates the property more directly to earlier
notions arising in work on direct inference.
 
We view {\em Direct Inference for Defaults\/}
as stating a (very weak) condition for how a default theory should
behave on some of the simpler
problems
involving hierarchies of classes and default properties.
Consider the following standard example, in which our
default
knowledge base $\KBfly$ is
  \begin{tabbing}
    XXXX \= \kill
    \> $\Bird(x) \default \Fly(x)$ \\
    \> $Penguin(x) \default \neg \Fly(x)$,  \\
    \> $\forall x \, (\Penguin(x) \rimp \Bird(x))$.
  \end{tabbing}
Should Tweety the penguin inherit the property of flying from the
class of birds, or the property of not flying from the class of
penguins? For any system satisfying {\em Direct
Inference for Defaults\/} we must have $\Penguin(\Tweety) \dentailssub{fly}
\lnot\Fly(\Tweety)$.
So long as the system treats universals in a reasonable
manner, this will be equivalent to
$\Penguin(\Tweety) \land
\Bird(\Tweety) \dentailssub{fly} \lnot\Fly(\Tweety)$. Thus we see that if a
system satisfies {\em Direct Inference for Defaults}, then it
automatically satisfies a form of
{\em specificity\/}---the preference for more
specific defaults. Specificity in default reasoning is, of course,
directly related to
the preference for more specific subsets that we saw in
the context of reference-class reasoning. Specificity
is one of the least controversial desiderata in default reasoning.
 
In approaches such as default reasoning or circumscription, the most
obvious encoding of these defaults satisfies neither
{\em Direct Inference for Defaults} nor specificity.
However, default logic and circumscription are certainly powerful enough for us
to be able to arrange specificity if we wish.  For example, in default
logic, this
can be done by means of {\em non-normal defaults\/} \cite{ReitCris}.
There is a cost to doing this, however: adding a default rule can require
that all older default rules be reexamined, and possibly changed, to
enforce the desired precedences.

{\em Direct Inference for Defaults\/} is a weak principle, since in most
interesting cases there is no default that fits the case at
hand perfectly. Suppose we learn that Tweety is a yellow penguin.
Should we still conclude that Tweety does not fly?  That is, should we
conclude
$\Penguin(\Tweety) \land \Yellow(\Tweety)
\dentailssub{fly} \lnot\Fly(\Tweety)$?
Most people would say
we should, because we have
been given no reason to suspect that yellowness is relevant to flight.
In other words, in the absence of more specific information about yellow
penguins we should use the most specific superclass for which we do have
knowledge, namely penguins.
The {\em inheritance\/} property, \ie
the ability to inherit defaults from superclasses,
is a second criterion for successful default reasoning,
and is not provided by {\em Direct Inference for Defaults}.
 
In some sense, we can view {\em Rational Monotonicity\/} as providing a
partial solution to this problem \cite{LehMag}.  If a nonmonotonic reasoning
system satisfies {\em Rational Monotonicity\/} in addition to {\em
Direct Inference for Defaults\/} then it does
achieve inheritance in a large number of examples.  For instance,
we have already observed that {\em Direct
Inference for Defaults} gives
$\Penguin(\Tweety) \dentailssub{fly} \neg\Fly(\Tweety)$, given $\KBfly$.
Since $\KBfly$ gives us no reason to believe that yellow penguins are
unusual, any reasonable default reasoning system would have
$\Penguin(\Tweety) \notdentailssub{fly} \neg \Yellow(\Tweety)$. From
these two statements, {\em Rational Monotonicity\/} allows us to conclude
$\Penguin(\Tweety) \land \Yellow(\Tweety) \dentailssub{fly}
\neg\Fly(\Tweety)$, as desired.
 
However, {\em Rational Monotonicity\/}
is still insufficient for inheritance reasoning in general.  Suppose
we add the default $\Bird(x) \default \Warmblooded(x)$
to $\KBfly$.
We would surely expect Tweety to be warm-blooded.
However, {\em Rational Monotonicity\/} cannot be applied here.
{To} see why, observe that $\Bird(\Tweety) \dentailssub{fly}
\Warmblooded(\Tweety)$, while we want to conclude that
$\Bird(\Tweety) \land \Penguin(\Tweety) \dentailssub{fly}
\Warmblooded(\Tweety)$.%
\footnote{In any system that treats universals reasonably,
this is clearly equivalent to the assertion we are really interested
in: $\Penguin(\Tweety) \dentailssub{fly} \Warmblooded(\Tweety)$.}
We could use {\em Rational Monotonicity\/} to go
{from} the first statement to the second, if we could show that
$\Bird(\Tweety) \notdentailssub{fly} \neg \Penguin(Tweety)$.  However,
most default reasoning systems do not support this statement.
In fact, since penguins are exceptional birds that do not fly, it
is not unreasonable to conclude the contrary, i.e., that $\Bird(Tweety)
\dentailssub{fly} \neg \Penguin(Tweety)$. Thus, {\em Rational
Monotonicity\/} cannot be used to conclude that $\Tweety$ the
penguin is warm-blooded.
 
It seems undesirable that if a subclass is exceptional in any one respect,
then inheritance of all other properties is blocked.
However, it can be argued
that this blocking of {\em inheritance to exceptional
subclasses\/} is reasonable. Since penguins are known to be
exceptional birds perhaps we should be cautious and not allow them to
inherit {\em any\/} of the normal properties of birds.
But even if we
accept
this argument, there are many
examples which demonstrate that the complete blocking of inheritance
to exceptional subclasses yields an inappropriately weak theory of
default reasoning.
For example, suppose we add to $\KBfly$ the default
$\Yellow(x) \default \Visible(x)$.
This differs from standard exceptional-subclass inheritance in that
yellow penguins are
not known to be
exceptional
members
of the class of yellow things. That is, while penguins are known to be
somewhat unusual birds (and so perhaps the normal properties of birds
should not be inherited), there is no reason to suppose that yellow
penguins are different from other yellow objects.
Nevertheless, {\em Rational Monotonicity\/} does not suffice even
in this less controversial case.  Indeed,
there are well-known systems that satisfy {\em
Rational Monotonicity\/} but cannot conclude that Tweety, the yellow
penguin, is easy to see
\cite{LehMag,Pearl:Z}.
This problem has been called the {\em drowning problem\/}
\cite{Asher:condworkshop,BCDLP:IJCAI}.
 
Theories of default reasoning have had considerable difficulty
in capturing an ability to inherit from superclasses that can deal
properly with all of these different cases.
In particular, the problem
of inheritance to exceptional subclasses has been the most difficult.
While some recent
propositional
theories have been more successful at dealing with exceptional
subclass inheritance
\cite{GMP,Geffner:thesis,Geffner:Conditional.Entailment}, they encounter
other difficulties,
which we discuss in the next section.
 
\subsection{Expressivity}
\label{nonmon.expressivity}
 
In the effort to discover basic techniques and principles for
default reasoning, people have often looked at weak languages
based on propositional logic. For instance, $\epsilon$-semantics and
variants \cite{Geffner:Framework.Reasoning.With.Defaults,GMP},
modal approaches such as autoepistemic logic \cite{Moore85},
and conditional logics \cite{Boutilier:Phd.Thesis},
are usually
considered in a propositional framework. Others,
such as Reiter's default
logic and Delgrande's conditional logic
\cite{Delgrande:Conditional.Logic.Revised},
use a first-order language, but with a
syntax that tends to decouple the issues of first-order reasoning and
default reasoning; we discuss this below. Of the better-known systems,
circumscription seems to have the ability, at least in principle,
of making the richest use of first-order logic.

It seems uncontroversial that, ultimately, a system of default
reasoning should be built around a powerful language.
Sophisticated knowledge representation systems
almost invariably use
languages with the expressive power of some large fragment
of first-order logic, if not much more.
It is
hard or impractical to encode the
knowledge we have about almost any interesting domain without
the expressive power provided by non-unary predicates and first-order
quantifiers.
We would also like to reason logically as well as by default
within the same system, and to allow perhaps even richer languages.

It
has not been easy to integrate first-order logic and defaults
completely.
In fact, one of the major contributions of our approach is
its ability to express both types of information in a single
language.
One
difficulty for other approaches
concerns ``open'' defaults, that are intended to apply to
all individuals. For instance, suppose we wish to make a general
statement that birds typically fly, and be able to use this when
reasoning about different birds. Let us examine how some existing
systems do this.
 
In propositional approaches, the usual strategy is to claim that there are
different types of knowledge (see, for example,
\cite{Geffner:Conditional.Entailment} and the references therein).
General defaults, such as
$\Bird \default \Fly$, are in one class. When we reason about an individual,
such as Tweety, its properties are described by knowledge in a different
class, the {\em context\/}. For Tweety, the context might be $\Bird \land
\Yellow$. In a sense, the symbol $\Bird$ stands for a general property when
used in a default and talks about Tweety (say) when it appears in the
context.
First-order approaches have more expressive power in this regard.
For example, Reiter's default logic
uses defaults with free variables, \eg $\Bird(x) \default \Fly(x)$.
That Tweety
is a bird can then be written $\Bird(\Tweety)$, which
seems much more natural. The default itself is treated essentially
as a schema, implying all substitution instances (such as $\Bird(\Tweety)
\default \Fly(\Tweety))$.
 
One example shows the difficulties  with both of these approaches.
Suppose we know that:
  \begin{tabbing}
    XXXX \= \kill
    \> Elephants typically like zookeepers. \\
    \> Fred is a zookeeper, but elephants typically do not like Fred.\\
    \> Clyde is an elephant.\\
    \> Eric is a zookeeper.
  \end{tabbing}
Using this information we can apply specificity to determine
reasonable answers to such
questions as ``Does Clyde
like Fred?'' (No) or
``Does Clyde like Eric'' (Yes).
But the propositional strategy of classifying knowledge seems to fail here.
Is ``Elephants typically do not like Fred'' a general default,
or an item of contextual knowledge? Since it talks about elephants
in general and also about one particular zookeeper, it does not fit
either category well. In a rich first-order language, there
is no clear-cut distinction between specific facts and general knowledge
(nor do we believe there should be one).

Next, consider the first-order substitutional approach. It is easy
to see that this does not work at all.
One substitution instance of
$$\Elephant(x) \land \Zookeeper(y) \default \Likes(x,y)$$
is
$$\Elephant(x) \land \Zookeeper(\Fred) \default \Likes(x,\Fred),$$
which will contradict the second default. Of course,
we could explicitly exclude Fred:
$$\Elephant(x) \land \Zookeeper(y) \land y \neq Fred \default \Likes(x,y).$$
However, explicit exclusion is similar to the process of explicitly
disabling less specific defaults, mentioned in the previous section.
Both
destroy the modularity of the knowledge base, i.e., the form of a
default becomes dependent on what other defaults are in the knowledge
base. Hence, these techniques are highly impractical for large
knowledge bases.

The zookeeper example is similar to an example given by Lehmann and
Magidor \cite{LehmannMagidor:TARK}.
However, the solution they suggest to this problem does not provide an
explicit interpretation for open defaults.
Rather, the
``meaning''
of an open default is implicitly determined by a set of
rules provided for manipulating such defaults.
These rules can cope with the zookeeper example, but the key step in
the application of these rules
is the use of {\em Rational Monotonicity}.
More precisely, Lehmann and Magidor's argument applies to systems
which, given the premise
$\Elephant(x) \land \Zookeeper(y)$, can infer by default that
$\Likes(x,y)$ (\ie $\Elephant(x) \land \Zookeeper(y) \dentails
\Likes(x,y)$), and yet cannot infer either $x \neq \Clyde$ or
$y \neq \Eric$. The latter certainly seem reasonable since we know nothing
whatsoever about Clyde or Eric.
Now, however, we can apply Rational Monotonicity twice, which
effectively allows us to {\em assume\/} (\ie add to the premises)
that $x = \Clyde \land y = \Eric$, while still concluding $\Likes(x,y)$.
Finally, Reflexivity, Right Weakening, and Left Logical Equivalence
can be used to justify substituting for $x$ and $y$; we obtain
$\Elephant(\Clyde) \land \Zookeeper(\Eric) \dentails \Likes(\Clyde,\Eric)$,
as desired.
The key point is that this
argument will typically fail for  $\Fred$,
because we do have reason to believe that Fred is unusual (and so,
in many systems, we could conclude by default that $y \neq \Fred$).
Thus, as we would hope, we cannot conclude that $\Likes(\Clyde,\Fred)$,
and in fact it is easy to argue analogously that we conclude
$\neg \Likes(\Clyde,\Fred)$ using the second default.
But while Rational Monotonicity helps in this example, we have,
in Section~\ref{nonmon.inheritance}, already seen its main failing:
it is easily blocked by ``irrelevant''
exceptionality.
For example,
if $\Eric$ is known to be exceptional in some way
(even one unrelated to zookeeping), then Lehmann and Magidor's approach
will not be able conclude that he is liked by Clyde. This is surely
undesirable.
 
Thus, it seems to be very hard to interpret generic (open) defaults
properly.
This is perhaps the best-known issue regarding the expressive power
of various approaches to default logic.
There are, of course, others;
we close by mentioning one.
 
Morreau \cite{Morreau:condworkshop} has discussed the usefulness of
being able to refer to ``the class of individuals satisfying a certain
default''.
For example, the assertion:
  \begin{tabbing}
    XXXX \= \kill
    \> Typically, people who normally go to bed late normally rise late.
  \end{tabbing}
refers to ``the class of people who normally go to bed late''.
The structure of this assertion is essentially:
  \begin{tabbing}
    XXXX \= \kill
    \> $(\Day(y) \default \Sleepslate(x,y)) \default
    (\Day(y') \default \Riseslate(x,y')).$
  \end{tabbing}
This is a default whose precondition and conclusion are descriptions
of people whose behaviors are themselves defined using defaults.
Such defaults appear to pose problems for most existing default theories.
Reiter's default logic cannot express such defaults. And while some
theories of conditional logic (for example, those of
\cite{Delgrande:Conditional.Logic.Revised,Boutilier:Phd.Thesis}) can
express this example, they are as yet incapable of generating
reasonable inferences from nested defaults of this type.
Circumscription, on the other hand,
could perhaps
be configured to
cope with this example, but precisely how this could be accomplished is not
obvious to us.
We also note that the example has many variants. For instance, there
is clearly a difference between
the above default and the one ``Typically, people who go to bed late
rise late (\ie the next
morning)''; formally, the latter statement could be written:
  \begin{tabbing}
    XXXX \= \kill
    \> $(\Day(y) \land \Sleepslate(x,y)) \default
    \Riseslate(x,\Nextday(y))),$
  \end{tabbing}
There are also other variations. We would like to express and reason
correctly with them all.
The real issue here is that we need to define various properties
of individuals, and while many of these properties can be expressed
in first-order logic, others need to refer to defaults explicitly.
This argues, yet again, that it is a mistake to have a different language
for defaults than the one used for other knowledge.

\subsection{The lottery paradox}
\label{lottery.discuss}
 
The {\em lottery paradox\/} (\cite{Kyburg:Rational.Belief})
addresses the issue of how different default
conclusions interact. It provides a challenging test of the
intuitions and semantics of any default reasoning system.
There are a number of
issues raised by this paradox; we consider three here.
 
First, imagine that
a large number $N$ of people buy tickets to a lottery in which there is
only one winner.
For a particular person $c$, it seems sensible to conclude
by default that $c$ does not win the lottery.
But we can argue this way for
every individual, which seems to
contradict the fact that someone definitely will win.
Of course some theories, such as those based on propositional languages,
do not have enough expressive power to even state this version
of the problem.
Among theories that can state it, there would seem to be several options.
Clearly, one solution is to deny that default conclusions
are closed under arbitrary conjunction,
\ie to give up on the {\em And Rule\/}.
But aside from explicitly
probabilistic theories, we are not aware
of work taking this approach (although the existence of multiple
extensions in theories such as Reiter's is certainly related).
Without logical closure, there is a danger of being too dependent
on merely syntactic features of a problem.
Another solution is to prevent a theory from reasoning about all $N$
individuals at once \cite{EKP:scope}.
Finally, one can simply deny that $\neg\Winner(c)$ follows
by default. Circumscription, for instance, does this:  The standard
representation of the problem would result in multiple extensions, such
that for each individual $c$, there is one extension where $c$ is the
winner.  While this seems reasonable, circumscription only allows us to
conclude things that hold in all extensions; thus, we would not be able
to conclude $\neg \Winner(c)$.
The problem with these ``solutions''
is that the lottery problem seems
to be an
extremely reasonable application of default reasoning: if you buy a lottery
ticket you {\em should\/}
continue your life under the assumption that you will not win.

Second, a closely related issue is
raised by
Lifschitz's list of benchmark problems \cite{Lifschitz.bench}.
Suppose we have a default, for instance $\Ticket(x) \default
\neg\Winner(x)$, and no other knowledge.
Should $\forall x (Ticket(x) \rimp \neg\Winner(x))$ be a default
conclusion?  Likewise,
if we know $\Winner(c)$ but
consider it possible that the lottery has more than one winner,
should we nevertheless conclude that
$\forall x ((Ticket(x) \land x\neq c) \rimp \neg\Winner(x))$?
In circumscription, although not in many other theories, we get
both universal conclusions (as Lifschitz argues for).
The desire for these universal conclusions is certainly controversial; in
fact it seems that we often {\em expect\/} default rules to
have some exceptions.
However, as Lifschitz observes, there is a technical difficulty in
following this latter intuition:
How can we conclude from the default $\Ticket(x) \default
\neg \Winner(x)$ that, by default, each individual $c$ is not a
winner, and yet not also reach the universal conclusion that, by default,
no one wins?
The concern is that, in many systems, the latter  conclusion will be
logically entailed whether we wish it or not.
Because of its treatment of open defaults, Reiter's
default logic does not suffer from this
difficulty.
As we shall see, neither does the random-worlds approach.

Finally,
Poole \cite{Poole91} has considered a variant of the lottery paradox that
avoids entirely the issue of named individuals.  In his version, there is
a formula describing the types of birds we are likely to encounter, such
as:
$$\forall x (\Bird(x) \dimp (\Emu(x) \lor \Penguin(x) \lor \ldots
\lor \Canary(x))).$$
We then add to the knowledge base defaults such as
birds typically fly, but penguins typically do not fly, and we similarly
assert that every other species of bird is exceptional in some way. Now
suppose all we know is that $\Bird(\Tweety)$.
Can we conclude that Tweety flies? If we conclude that he can, then a similar
argument would also allow us to conclude that he is a typical bird in all
other respects.  But this would contradict the fact he must be
exceptional in some respect.  If we do not conclude that Tweety flies,
then the default ``Birds typically fly'' has been
effectively ignored.  Poole uses such examples to give an exhaustive
analysis of how various systems might react to the Lottery Paradox.
He shows that in any theory, some desideratum, such as closure under
conjunction or ``conditioning''
({\em Direct
inference for defaults\/}), must be sacrificed.
Perhaps the most interesting
``way out'' he discusses is the possibility of declaring that certain
combinations of defaults are inadmissible or inconsistent.  Is it really
reasonable to say that the class of birds is the union of subclasses all
of which are exceptional?  In many theories, such as Reiter's default
logic, there is nothing to prevent one from asserting this. But in a
theory which gives reasonable semantics to defaults, we may be able to
determine and justify the incompatibility of certain sets of defaults.
This, indeed, is how our approach avoids Poole's version of the lottery
paradox.
 
\subsection{Discussion}
 
In this section, we have presented a limited list of desiderata that seem
appropriate for a default reasoning system,
and have discussed some key problems and issues that must be resolved
by such a system.
While our list may be limited,
it is interesting to point out that there does not seem to be a single
default reasoning system that fulfills all these desiderata
in a satisfactory way.
Although we can (and do) show that random worlds does, in fact,
achieve all the requirements on this list, we would like to validate
random worlds in a more comprehensive fashion.
Unfortunately, to the best of our knowledge, there is (as
yet) no general framework for evaluating default reasoning systems.  In
particular, evaluation still tends to be on the level of ``Does
this theory solve these particular examples correctly?'' (see, for
example, the list of benchmark problems in \cite{Lifschitz.bench}).
While such examples are often important in identifying interesting
aspects of the problem and defining our intuitions in these cases,
they are clearly not a substitute for a comprehensive framework.  Had
there been such a framework, perhaps the drowning problem from
Section~\ref{nonmon.inheritance} would not have remained undiscovered for so
long.  While we do not attempt to provide such a general framework in
this paper, in Section~\ref{results} we prove a number of general
theorems concerning the random-worlds approach.  These theorems provide a
precise formulation of properties such as {\em Direct Inference for Defaults},
and show that they hold for random worlds.
Other properties such as
specificity and exceptional subclass inheritance
follow immediately from these theorems.
Thus, our proof that the random-worlds approach deals well with the
paradigm examples in default reasoning
follows from a general
theorem, rather than by a case-by-case analysis.
 
\section{The formalism}
\label{formalism}
 
\subsection{The language}
\label{language}
We are interested in a formal logical language that allows
us to express both statistical information and first-order
information. We therefore define a statistical language $\Laeq$,
which is a variant of a language designed by Bacchus \cite{Bacchus}.
For the remainder of the paper, let $\vocab$ be
a finite first-order vocabulary, consisting of predicate,
function,
and constant
symbols, and let $\X$ be a set of variables.%

Our statistical language augments standard first-order logic with a
form of statistical quantifier.  For a formula $\psi(x)$,
the term $\prop{\psi(x)}{x}$ is a {\em
proportion expression}.  It will be interpreted as a rational number
between 0 and 1, that represents the proportion of
domain elements satisfying $\psi(x)$.
We actually allow an arbitrary set of variables in the subscript
and in the formula $\psi$.
Thus, for example, $\prop{\Child(x,y)}{x}$ describes, for a fixed $y$,
the proportion of domain elements that are children of $y$;
$\prop{\Child(x,y)}{y}$ describes, for a fixed $x$, the proportion of
domain elements whose child is $x$; and $\prop{\Child(x,y)}{x,y}$
describes the proportion of pairs of domain elements that are in the
child relation.%

We also allow proportion expressions of the form
$\cprop{\psi(x)}{\theta(x)}{x}$, which we call {\em conditional
proportion expressions}. Such an expression is intended to denote
the proportion of domain elements satisfying
$\psi$ from among those elements satisfying $\theta$.
Finally, any rational number is also considered to be a proportion
expression, and the set of proportion
expressions is closed under addition and multiplication.
 
One important difference between our syntax and that of \cite{Bacchus}
is the use of {\em approximate equality\/} to compare proportion
expressions.
As we argued in the introduction,
exact comparisons are sometimes inappropriate. Consider a statement
such as ``80\% of patients with jaundice have hepatitis''. If this
statement appears in a knowledge base, it is almost certainly there as
a summary of a large pool of data.  It is clear that we do not mean
that {\em exactly\/} 80\% of all patients with jaundice have
hepatitis. Among other things, this would imply that the number of
jaundiced patients is a multiple of five, which is surely not
an intended implication.
We therefore use the approach described in
\cite{GHK2,KH92}, and compare proportion expressions using
(instead of $=$ and $\leq$) one of an infinite family of connectives
$\aeq_i$ and $\aleq_i$, for $i = 1,2,3\ldots{}$ (``$i$-approximately
equal'' or ``$i$-approximately less than or equal'').%
\footnote{In \cite{BGHK} the use of approximate
equality was suppressed in order to highlight other issues.}
For example, we can express the statement ``80\% of jaundiced patients
have hepatitis'' by the {\em proportion formula\/}
$\cprop{\Hep(x)}{\Jaun(x)}{x} \aeq_1 0.8$.
The intuition behind the semantics of approximate equality is that
each comparison should be interpreted using some small tolerance
factor to account for measurement error, sample variations, and so on.
The appropriate tolerance will differ for various
pieces of information, so our logic allows different
subscripts on the ``approximately equals'' connectives.  A formula such as
$\cprop{\Fly(x)}{\Bird(x)}{x} \aeq_1 1 \land \cprop{\Fly(x)}{\Bat(x)}{x}
\aeq_2 1$ says that both
$\cprop{\Fly(x)}{\Bird(x)}{x}$ and $\cprop{\Fly(x)}{\Bat(x)}{x}$ are
approximately $1$, but the notion of ``approximately'' may be different in
each case.
 
We can now give a recursive definition of the language $\Laeq$.
\dfn\label{Laeqdfn}
The set of {\em terms\/} in $\Laeq$
is the least set containing $\X$ and the constant symbols in $\vocab$
that is
closed under function application (so that if $f$ is a function
symbol in $\vocab$ of arity $r$, and $t_1, \ldots, t_r$ are terms,
then so is $f(t_1, \ldots, t_r)$).
 
\noindent
The set of {\em proportion expressions\/} is
the least set that
\begin{itemize}
  \item[(a)]
        contains the rational numbers,
  \item[(b)] contains {\em proportion terms\/} of the form
        $\prop{\psi}{X}$ and $\cprop{\psi}{\theta}{X}$, for formulas
        $\psi, \theta \in \Laeq$ and a finite set of variables $X
        \subseteq \X$, and
  \item[(c)] is closed under addition
        and multiplication.
\end{itemize}
The set of formulas in $\Laeq$ is the least set that
\begin{itemize}
  \item[(a)] contains {\em atomic
        formulas\/} of the form $R(t_1, \ldots, t_r)$, where $R$ is
        a predicate symbol in $\vocab \union \{=\}$ of arity $r$ and
        $t_1,\ldots,t_r$ are terms,
  \item[(b)] contains {\em proportion formulas\/} of the
        form $\zeta \aeq_i \zeta'$ and $\zeta \aleq_i \zeta'$, where
        $\zeta$ and $\zeta'$ are proportion expressions
        and $i$ is a natural number, and
  \item[(c)] is closed under
        conjunction, negation, and
        first-order quantification. \qed
\end{itemize}
\end{definition}
 
Notice that this definition allows
arbitrary nesting of quantifiers and proportion expressions.
In Section~\ref{statdefaults} we demonstrate the expressive power
of the language.
As observed in \cite{Bacchus}, the appearance of a variable $x$ in the
subscript of a
proportion expression binds the variable $x$ in the expression;
indeed, we can view $\prop{\cdot}{X}$ as a new type of quantification.
 
We now need to define the semantics of the logic.  As we shall see
below, most of the definitions are fairly straightforward. The two
features that cause problems are approximate comparisons and
conditional proportion expressions. We interpret the approximate
connective $\zeta \aeq_i \zeta'$ to mean that $\zeta$ is very close
to $\zeta'$.  More precisely, it is within some very small, but
unknown, tolerance factor. We formalize this using a {\em tolerance vector\/}
$\epsvec = \langle \epscom_1, \epscom_2, \ldots \rangle$, $\epscom_i
> 0$. Intuitively $\zeta \aeq_i \zeta'$ if the values of
$\zeta$ and $\zeta'$ are within $\epscom_i$ of each other.
(Note that,
although the use of tolerance vectors leads
to well-defined formal semantics, one might object that in practice
we generally will not know appropriate
tolerance
values.  We defer our response
to this objection to the next section.)

A difficulty arises when interpreting conditional proportion expressions
because we need
to deal with the problem of conditioning on an event of measure 0.
That is, we need to define semantics for
$\cprop{\psi}{\theta}{X}$ even when there are no assignments to
the variables in $X$ that would satisfy $\theta$.
When standard equality is used rather than approximate equality, this
problem is easily overcome. Following \cite{Hal4}, we can
eliminate conditional
proportion expressions altogether by viewing a statement
such as $\cprop{\psi}{\theta}{X} = \alpha$ as an abbreviation for
$\prop{\psi \land \theta}{X} = \alpha\prop{\theta}{X}$.  This
approach agrees with the standard interpretation of conditionals
if $\prop{\theta}{X} \ne 0$.
If $\prop{\theta}{X} = 0$, it enforces the convention that formulas
such as $\cprop{\psi}{\theta}{X} = \alpha$ or $\cprop{\psi}{\theta}{X} \le
\alpha$ are true
for any $\alpha$.
We used the same approach
in \cite{GHK2}, where we allowed approximate equality.
Unfortunately, as the following example shows, this interpretation of
conditional proportions can interact
in an undesirable way with the semantics
of approximate comparisons.
In particular,
this approach does not preserve the standard
semantics of conditional equality if $\prop{\theta}{X}$ is
{\em approximately\/} 0.
\xam\label{aeqproblem}
Consider the knowledge base:%
\footnote{We remark that, here and in our examples below, the actual
choice of subscript for $\aeq$ is unimportant.
However, we use different subscripts for different approximate
comparisons unless the tolerances for the different
measurements are known to be the same.}
$$\KB = (\prop{\Penguin(x)}{x} \aeq_1 0) \land
(\cprop{\Fly(x)}{\Penguin(x)}{x} \aeq_2 0).$$
We expect this to mean that the proportion of penguins is very small
(arbitrarily close to 0 in large domains), but also that the
proportion of fliers among penguins is also very
small. However, if we
attempt to
interpret conditional proportions as discussed above,
we obtain the knowledge base
$$\KBp = (\prop{\Penguin(x)}{x} \aeq_1 0) \land (\prop{\Fly(x) \land
\Penguin(x)}{x} \aeq_2 0 \cdot \prop{\Penguin(x)}{x}),
$$
which is equivalent to
$$
(\prop{\Penguin(x)}{x} \aeq_1 0) \land (\prop{\Fly(x) \land
\Penguin(x)}{x} \aeq_2 0).$$
This last formula simply asserts that  the proportion of penguins and the
proportion of flying penguins are both small, but says nothing about
the proportion of fliers among penguins.
In fact, the world where all penguins fly is
consistent with $\KBp$.  Clearly, the process of multiplying out
across an approximate connective does not preserve the
intended interpretation of the formulas.
\exam
 
Because of this problem, we cannot treat conditional proportions as
abbreviations and instead
have added
them as
primitive expressions in the language.
Of course, we now have to give them a semantics
that avoids the problem illustrated by Example~\ref{aeqproblem}.
We would like to maintain the
conventions used
when we had equality in the language.  Namely, in worlds where
$\prop{\theta(x)}{x} \ne 0$, we want
$\cprop{\phi(x)}{\theta(x)}{x}$ to denote
the fraction of elements satisfying $\theta(x)$ that also
satisfy $\phi(x)$.  In worlds where $\prop{\theta(x)}{x} = 0$,
we want
all formulas of the form $\cprop{\phi(x)}{\theta(x)}{x} \aeq_i
\alpha$ or $\cprop{\phi(x)}{\theta(x)}{x} \aleq_i \alpha$ to be true.
There are a number of ways of accomplishing this.  The
route we take is
perhaps not the simplest, but it introduces machinery that will
be helpful later.

We give semantics to the language $\Laeq$ by
providing a translation from formulas
in $\Laeq$ to formulas in a language $\Leq$ whose semantics is more
easily described.
The language $\Leq$ is essentially the language of
\cite{Hal4}, that uses true equality rather than approximate equality.
More precisely, the definition of $\Leq$ is identical to the definition
of $\Laeq$ given in Definition~\ref{Laeqdfn}, except that:
\begin{itemize}
  \item
        we use $=$ and $\le$ instead of $\aeq_i$ and $\aleq_i$,
  \item
        we allow the set of proportion expressions to include arbitrary real
        numbers (not just rational numbers),
  \item
        we do not allow conditional proportion expressions,
  \item
        we assume that $\Leq$ has a special family of variables
        $\vareps_i$, interpreted over the reals.
\end{itemize}
As we shall see, the variable $\vareps_i$ is used to interpret
the approximate equality connectives $\aeq_i$ and $\aleq_i$.
We view an expression in $\Leq$ that uses conditional
proportion expressions as an abbreviation for the
expression obtained by multiplying out.

The semantics for $\Leq$ is quite straightforward, and
follows the lines of \cite{Hal4}.
Recall that we give semantics to $\Leq$ in terms of {\em worlds\/}, or
finite first-order models.  For any natural number $N$, let
$\WN(\vocab)$
consist of all worlds with domain $D = \{1,\ldots, N\}$ over the vocabulary
$\vocab$.

Now, consider a world $\world \in \WN(\vocab)$,
a valuation $\val : \X \rightarrow \{1, \ldots, N \}$ for the variables
in $\X$, and a tolerance vector $\epsvec$.
We simultaneously assign to each proportion expression $\zeta$ a real
number $[\zeta]_{(\world,\val,\epsvec)}$ and to each formula $\xi$ a
truth value with respect to $(\world, \val, \epsvec)$.
Most of the clauses of the definition are completely standard, so we
omit them here.  In particular, variables are interpreted using
$\val$, each tolerance variable $\vareps_i$ is interpreted as denoting
the tolerance $\eps_i$, the predicates and constants are interpreted
using $\world$, the Boolean connectives and the first-order
quantifiers are defined in the standard fashion, and when interpreting
proportion expressions, the real numbers, addition, multiplication,
and $\le$ are given their standard meaning. It remains to interpret
proportion terms.  Recall that we eliminate conditional proportion
terms by multiplying out, so that we need to deal only with
unconditional proportion terms. If $\zeta$ is the proportion expression
$\prop{\psi}{x_{i_1},\ldots,x_{i_k}}$ (for $i_1 < i_2 < \ldots < i_k$),
then
$$
[\zeta]_{(\world,\val,\epsvec)} = \frac{1}{|D|^k}
\Bigl|\left\{(d_1,\ldots,d_k) \in D^k\ :\
(\world,\val[x_{i_1}/d_1,\ldots,x_{i_k}/d_k],\epsvec) \sat
\psi\right\}\Bigr|.
$$
Thus, if $\world \in \WN(\vocab)$, the proportion expression
$\prop{\psi}{x_{i_1},\ldots,x_{i_k}}$ denotes the fraction of the
$N^k$ $k$-tuples
of domain elements in $D$ that satisfy $\psi$
in the world $\world$.
For example, $[\prop{\Child(x,y)}{x}]_{(\world,\val,\epsvec)}$ is the
fraction of domain elements $d$ that are children of $\val(y)$.

We now show how a formula $\chi \in \Laeq$ can be associated with
a formula $\chi^* \in \Leq$.  We proceed as follows:
\begin{itemize}
  \item
        every proportion formula $\zeta \aleq_i \zeta'$
        in $\chi$ is (recursively) replaced by
        $\zeta - \zeta' \leq \vareps_i$,
  \item
        every proportion formula $\zeta \aeq_i \zeta'$ in $\chi$ is
        (recursively) replaced by
        the conjunction $(\zeta - \zeta' \leq \vareps_i) \land
        (\zeta' - \zeta \leq \vareps_i)$,
  \item finally, conditional proportion expressions are eliminated as in
        \cite{Hal4}'s
        semantics, by multiplying out.
\end{itemize}
This translation allows us to embed $\Laeq$
in $\Leq$.
Thus, for the remainder of the paper, we regard $\Laeq$ as a sublanguage
of $\Leq$.
We can now easily define the semantics of formulas
in $\Laeq$: For $\chi \in \Laeq$, we say that $(\world,\val,\epsvec) \sat
\chi$ iff $(\world,\val,\epsvec) \sat \chi^*$.
It is sometimes useful
to incorporate particular
values for the tolerances into the formula $\chi^*$.  Thus, let
$\chi[\epsvec]$ represent the formula that results from $\chi^*$ if
each variable $\vareps_i$ is replaced
by $\eps_i$,
its value according to
$\epsvec$.%
\footnote{Note that some of the tolerances $\eps_i$ may be irrational;
  it is for this reason that we allowed
arbitrary real numbers
in the proportion expressions of $\Leq$.}

Typically we are interested in closed sentences,
that is, formulas with no free variables.
In that case, it is not hard to show that the valuation plays
no role.  Thus, if $\chi$ is closed, we write $(\world, \epsvec)
\sat \chi$ rather than $(\world, \val,\epsvec) \sat \chi$.

\subsection{Degrees of belief}\label{degbelief}
As we explained in the introduction, we give semantics to degrees
of belief by considering all worlds of size $N$ to be equally likely,
conditioning on $\KB$, and then checking the probability of $\phi$ over
the resulting probability distribution.
In the previous section, we defined what it means for a sentence
$\chi$ to be satisfied in a world of size $N$ using a tolerance
vector $\epsvec$.
Given $N$ and $\epsvec$, we define $\nworlds(\chi)$ to be the number
of worlds in $\WN(\vocab)$ such that $(\world,\epsvec) \sat \chi$.
Since we are taking all worlds to be equally likely,
the degree of belief in $\phi$ given $\KB$
with respect to $\WN$ and $\epsvec$ is
$$
\prNw(\phi|\KB) =
\frac{\nworlds(\phi\land\KB)}{\nworlds(\KB)}.
$$
If $\nworlds(\KB) = 0$, this degree of belief is not well-defined.%
\footnote{
Strictly speaking, we should write
$\nworldsv{\vocab,\epsvec}_{N}(\chi)$ rather than
$\nworlds(\chi)$, since the number also depends on the choice of $\vocab$.
Indeed, we do so in the one place where this dependence matters
(Theorem~\ref{independent}).
The degree of belief is, however unaffected by expansions of the
vocabulary. That is, if $\vocab' \supset \vocab$ then the degree of
belief $\prNw(\phi|\KB)$ is the same under the vocabulary $\vocab'$ as
it is under $\vocab$.
}

Typically, we know neither $N$ nor $\epsvec$ exactly.  All we know is
that $N$ is ``large'' and that $\epsvec$ is ``small''. Thus, we would
like to take our {\em degree of belief\/} in $\phi$ given $\KB$ to be
$\lim_{\epsvec \tendsto \vec{0}}\, \lim_{N \tendsto \infty}\,
\prNw(\phi|\KB)$.  Notice that the order of the two limits over
$\epsvec$ and $N$ is important. If the limit $\lim_{\epsvec \tendsto
  \vec{0}}$ appeared last, then we would gain nothing by using
approximate equality, since the result would be equivalent to treating
approximate equality as exact equality.
 
This definition, however, is not sufficient; the limit may not
exist. We observed above that
$\prNw(\phi|\KB)$ is not always well-defined.  In particular, it may
be the case that for certain values of $\epsvec$, $\prNw(\phi|\KB)$ is
not well-defined for arbitrarily large $N$.
In order to deal with this problem of well-definedness, we define
$\KB$ to be {\em eventually consistent\/} if for all sufficiently small
$\epsvec$ and sufficiently
large $N$, $\nworlds(\KB) > 0$. Among other things, eventual
consistency implies
that the $\KB$ is satisfiable in finite domains of arbitrarily
large size.
For example, a $\KB$
stating that ``there are exactly 7 domain elements'' is not
eventually consistent.%
\footnote{Of course, in this case one probably would not want to consider
$\lim N \tendsto \infty$ anyway. If we are fortunate enough to
know the domain size, and it is reasonably small, we can simply
compute degrees of belief using the (known) fixed value of $N$.}
For the remainder of the paper, we assume that
all knowledge bases are eventually consistent.
 
Even if $\KB$ is eventually consistent, the limit may not exist.
For example, it may be the case that
for some $i$, $\prNw(\phi|\KB)$ oscillates
between $\alpha + \epscom_i$ and $\alpha - \epscom_i$
as $N$ gets large.  In this case, for any particular $\epsvec$, the
limit as $N$ grows will not exist.  However, it seems as if the limit
as $\epsvec$ grows small ``should'', in this case, be $\alpha$, since
the oscillations about $\alpha$ go to~0.  We avoid such problems by
considering the {\em lim sup\/} and {\em lim inf\/}, rather than the limit.
For any set $S \subset \reals$, the infimum of $S$, $\inf S$, is
the greatest lower bound of $S$. The {\em lim inf\/} of a sequence is the
limit of the infimums; that is,
$$\liminf_{N \tendsto \infty} a_N =
\lim_{N \tendsto \infty} \inf\{a_i: i>N\}.$$
The lim inf exists for any sequence bounded from below, even if the
limit does not. The {\em lim sup\/} is defined analogously, where
$\sup S$ denotes the least upper bound of $S$. If $\lim_{N \tendsto
\infty} a_N$ does exist, then $\lim_{N \tendsto \infty} a_N =
\liminf_{N \tendsto \infty} a_N = \limsup_{N \tendsto \infty} a_N$.
Since, for any $\epsvec$, the sequence $\prNw(\phi|\KB)$ is always
bounded from above and below, the lim sup and lim inf always exist.
Thus, we do not have to worry about the problem of nonexistence for
particular values of $\epsvec$.  We can now present the final form of
our definition.
 
\dfn
If
$$
\lim_{\epsvec \tendsto \vec{0}}\, \liminf_{N \tendsto \infty}\,
\prNw(\phi|\KB)
\mbox{~~and~~}
\lim_{\epsvec \tendsto \vec{0}}\, \limsup_{N \tendsto \infty}\,
\prNw(\phi|\KB)
$$
both exist and are equal, then
the {\em degree of belief in $\phi$ given $\KB$}, written
$\priw(\phi|\KB)$, is defined as the common limit;
otherwise $\priw(\phi|\KB)$ does not exist.
\end{definition}
 
We point out that, even using this definition, there are many cases
where the degree of belief does not exist.
However, as some of our examples show, in many situations the
nonexistence of a degree of belief can be understood intuitively,
and is sometimes related to the existence
of multiple extensions of a default theory. (See
Sections~\ref{statdefaults} and~\ref{competing.results} and \cite{GHK2}.)

We remark that Shastri \cite{Shastri} used a somewhat similar
approach to defining degrees of belief. His language does not
allow
the direct expression of statistical information,
but does
allow us to talk about the number of domain individuals that satisfy a
given predicate.  He then gives a definition of degree of belief
similar to ours.  Since he has no notion of approximate equality in
his language, and presumes a fixed domain size
(an assumption we wish to avoid),
he does not have to deal with limits as we do.
 
\subsection{Statistical interpretation for defaults}
\label{statdefaults}
 
As we mentioned in the introduction, there are many similarities between
direct inference from statistical information and default reasoning.
To capitalize on this observation, and to be able to
use random worlds as a default
reasoning system, we need to interpret defaults as statistical statements.
However, finding the appropriate statistical interpretation is not
straightforward.
For example, as is well known, if we interpret
``Birds typically fly'' as
``Most (\ie more than $50\%$ of) birds fly'',
then we get a default system that fails to satisfy some of the most
basic desiderata, such as the {\em And\/} rule, discussed in
Section~\ref{KLM}.
Using a higher fixed threshold in a straightforward way does not help. More
successfully,  Adams \cite{Adams}, and later Geffner and Pearl
\cite{Geffner:Framework.Reasoning.With.Defaults}, suggested an
interpretation of defaults based on ``almost all''.   In their framework,
this is done using {\em extreme probabilities\/}---conditional probabilities
that are arbitrarily close to 1: \ie within $1-\epsilon$ for some
$\epsilon$, and considering the limit as $\epsilon \rightarrow 0$.  The
basic system derived from this idea is called {\em $\epsilon$-semantics}.
Later, stronger systems
(that are able to make more inferences) based on the
same probabilistic idea were introduced (see Pearl \cite{Pearl90} for a
survey).

The intuition behind $\epsilon$-semantics and its extensions seems to be
statistical.  However, since the language used in these approaches
is propositional, this intuition cannot be expressed directly.
Indeed, these approaches typically make no distinction between
the statistical nature of the default and the degree of belief
nature of the default conclusion.
We are able to capture this intuition more directly in our approach,
since we can make this distinction explicitly.
Recall that we interpret a
statement such as ``Birds typically fly'' statistically, using the
approximate statement
$\cprop{\Fly(x)}{\Bird(x)}{x} \aeq_i 1$ for some $i$.
(Thus,
the use of an approximate connective to compare proportion expressions
is not purely a technical convenience.)  Clearly, we can view our
statistical interpretation of defaults as a generalization of the
extreme probabilities interpretation of defaults to the first-order
case.
The connection between our work and $\epsilon$-semantics extends
beyond the issue of representation: there is a deeper sense in which we can
view our approach
as the generalization of one of the extensions of $\epsilon$-semantics,
namely the maximum-entropy approach of Goldszmidt, Morris, and
Pearl \cite{GMP}, to the first-order setting.  This issue is
discussed in more detail in
Section~\ref{GMPsec}, where it is shown that this maximum-entropy
approach can be embedded in our framework.

Of course, the fact that our syntax is so rich allows us to express
a great deal of information that simply cannot be expressed in
any propositional approach.  We observed earlier
that a propositional
approach that distinguishes between default knowledge and contextual
knowledge has difficulty in dealing with the elephant-zookeeper example
(see Section~\ref{nonmon.expressivity}).
This example is easily dealt with in our framework.
\xam
\label{elephants}
The following knowledge base, $\KBel$, is a formalization of the
elephant-zookeeper example.
Recall, this problem concerns the defaults that (a) Elephants typically
like zookeepers, but (b) Elephants typically do not like Fred.
As discussed earlier, simply expressing this knowledge %
can be a challenge.
In our framework this example can be expressed as follows:
$$
\begin{array}{l}
  \cprop{\Likes(x,y)}{\Elephant(x) \land \Zookeeper(y)}{x,y} \aeq_1\, 1 \ \land\\
  \cprop{\Likes(x,\Fred)}{\Elephant(x)}{x} \aeq_2\, 0 \ \land \\
  \Zookeeper(\Fred)
  \land \Elephant(\Clyde) \land \Zookeeper(\Eric). \ \ \  \ \ \ \ \qed
\end{array}
$$
\end{example}
Furthermore, our interpretation of defaults
allows us to deal well with
interactions between first-order quantifiers and defaults.
\xam
\label{tall.parent}
We may know that people who have at least one tall parent
are typically tall.  This default can
be expressed in our language:
$$ \cprop{\Tall(x)}{\exists y \,
  (\Child(x,y) \land \Tall(y))}{x} \aeq_i 1.  \ \ \ \ \qed$$
\end{example}
We can
also define defaults over classes themselves defined using
default rules (as discussed by Morreau \cite{Morreau:condworkshop}).
\xam
\label{sleeplate}
In Section~\ref{nonmon.expressivity}, we discussed the problem
of expressing the nested default ``Typically, people who
normally go to bed late normally rise late.''  To express this default
we can simply use nested proportion statements:
The individuals who normally rise late are those who rise late most
days; they are the $x$'s satisfying
$\cprop{\Riseslate(x,y)}{\Day(y)}{y}$ $\aeq_1 1$. Similarly, the
individuals who normally go to bed late are the $x$'s satisfying
$\cprop{\Sleepslate(x,y')}{\Day(y')}{y'}$ $\aeq_2 1$.
Thus we can capture the default
by saying most $x$'s that go to bed late also rise late, as in
the knowledge base $\KBlate$:
$$
\Bigcprop{\cprop{\Riseslate(x,y)}{\Day(y)}{y} \aeq_1 1}
{\cprop{\Sleepslate(x,y')}{\Day(y')}{y'} \aeq_2 1}{x} \aeq_3 1.
$$
On the other hand, the related default that ``Typically, people who go to
bed late rise late (\ie the next morning)'' can be expressed as:
$$
\Bigcprop{\Riseslate(x,\Nextday(y))}
{\Day(y) \land \Sleepslate(x,y)}{x,y} \aeq_1 1,
$$
which is clearly different from the first default.
\exam

\section{Properties of random worlds}
\label{results}

We now show that the random-worlds method validates several
desirable reasoning patterns, including
essentially all of those discussed in
Sections~\ref{refclass} and~\ref{nonmon}.
It is worth noting that all of these reasoning patterns follow from
the basic definition of the random worlds method given in
Section~\ref{degbelief}; none of these patterns require any additional
structure to be added to the method.
We also note that all the results in this section hold for our
language in its full generality: the formulas can
contain arbitrary
function and predicate symbols (including non-unary predicates),
and have nested quantifiers
and proportion statements.
Finally, we note that the theorems we state are not the most general ones
possible. It is quite easy to construct examples for which the conditions of
the theorems do not hold, but random worlds still gives the intuitively
plausible answer.
We could find theorems that deal with additional cases,
although it seems to be fairly difficult to
find other results whose conditions are
easy to state and check, and yet cover an interestingly large class of
examples. We discuss this issue again in
Section~\ref{discussion.complexity}.

\subsection{Random worlds and default reasoning}
\label{KLMresults}
In this subsection, we focus on formulas which are assigned degree of
belief 1.
Given any knowledge base $\KB$ (which can, in particular, include
defaults using the statistical interpretation of
Section~\ref{statdefaults}),
we say that {\em $\phi$ is a default conclusion from $\KB$},
and write $\KB \rwent \phi$,
if $\priw(\phi|\KB) = 1$.
As we now show, the relation $\rwent$
satisfies all the basic properties of default inference
discussed in Section~\ref{KLM}.
We start by proving two somewhat more general results.

\pro
\label{lle}
If $\sat \KB \dimp \KBp$, then $\Prinf(\phi|\KB) =
\Prinf(\phi|\KBp)$ for all formulas $\phi$.%
\footnote{By $\Prinf(\phi|\KB) =
\Prinf(\phi|\KBp)$
we mean that either both degrees of belief exist
and have the same value, or neither exists.
Proposition~\ref{cmc} should be interpreted analogously.}
\epro
\prf
By assumption, precisely the same set of worlds satisfy
$\KB$ and $\KBp$.  Therefore, for all $N$ and $\epsvec$, $\prNw(\phi|\KB)$
and $\prNw(\phi|\KBp)$ are equal.
Therefore, the limits are also equal. \eprf
 
\pro\label{cmc}
If $\KB \rwent \theta$, then
$\priw(\phi|\KB) = \priw(\phi|\KB\land\theta)$
for any $\phi$. \epro
\prf
Fix $N$ and $\epsvec$. Then, by the standard properties of
conditional probability, we get
$$\prNw(\phi|\KB)
=  \prNw(\phi|KB\land\theta)\cdot\prNw(\theta|\KB)\; +
\prNw(\phi|KB\land\neg\theta)\cdot\prNw(\neg\theta|\KB).
$$%
By assumption, $\prNw(\theta|\KB)$ tends to 1 when we take limits,
so the first summand tends to
$\priw(\phi|\KB\land\theta)$. Since
$\prNw(\neg\theta|\KB)$ has limit 0 and
$\prNw(\phi|KB\land\neg\theta)$ is bounded, the second
summand tends to 0. The result follows. \eprf
 
\thm
\label{KLM.properties}
The relation $\rwent$ satisfies the properties of And,
Cautious Monotonicity, Cut, Left Logical Equivalence, Or, Reflexivity, and
Right Weakening.
 
\ethm
\prf
\begin{description}
  \item[{\it And:}]  As we mentioned
        in Section~\ref{KLM}, this follows from the other
        properties proved below.
  \item[{\it Cautious Monotonicity and Cut:}]
        These follow immediately from Proposition~\ref{cmc}.
 
  \item[{\it Left Logical Equivalence:}] Follows immediately
        from Proposition~\ref{lle}.
 
  \item[{\it Or:}]
        Assume $\priw(\phi|\KB) = \priw(\phi|\KBp) = 1$, so that
        $\priw(\neg \phi | \KB) = \priw(\neg \phi | \KBp) = 0$.
        Fix $N$ and $\epsvec$.  Then
        \begin{eqnarray*}
          \prNw(\neg \phi | \KB \lor \KBp)
          & = & \prNw(\neg \phi\land (\KB \lor \KBp) | \KB \lor \KBp) \\
          & \leq & \prNw(\neg \phi \land \KB | \KB \lor \KBp)\; +
          \prNw(\neg \phi \land \KBp | \KB \lor \KBp) \\
          & \leq & \prNw(\neg \phi|\KB) + \prNw(\neg \phi | \KBp).
        \end{eqnarray*}
      Taking limits, we conclude that
      $\Prinf(\neg \phi| \KB \lor \KBp) = 0$.  It follows that
      $(\KB \lor \KBp) \rwent \phi$.
  \item[{\it Reflexivity:}]
        Because we restrict
        our
        attention to $\KB$'s that are eventually consistent,
        $\priw(\KB|\KB)$ is well-defined. But then $\priw(\KB|\KB)$ is clearly
        equal to 1.
  \item[{\it Right Weakening:}]
        Suppose $\priw(\phi|\KB) = 1$.
        If $\sat \phi \rimp \phi'$, then the set of worlds satisfying $\phi'$ is a
        superset of the set of worlds satisfying $\phi$.  Therefore, for any $N$
        and $\epsvec$, $\prNw(\phi'|\KB) \geq \prNw(\phi|\KB)$.  Taking limits, we
        obtain that
        $$
        1 \geq \priw(\phi'|\KB) \geq \priw(\phi|\KB) = 1,
        $$
        and so necessarily $\priw(\phi'|\KB) = 1$.~\eprf
\end{description}

Besides demonstrating that $\rwent$ satisfies the minimal
standards of reasonableness for a default inference relation,
these properties, particularly the stronger
form of {\em Cut\/} and {\em Cautious Monotonicity\/} proved
in Proposition~\ref{cmc}, will prove quite useful in computing
degrees of belief, especially when combined with some other properties
we prove below (see also Section~\ref{discussion.complexity}).
In particular, many of our later results show how random-worlds
behaves for knowledge bases and queries that have certain restricted
forms. Sometimes a $\KB$ that does not satisfy these requirements can
be changed into one that does, simply by extending $\KB$ with some of
its default conclusions. We then appeal to Proposition~\ref{cmc} to
justify using the new knowledge base instead of the old one.
The other rules are also useful, as shown in the
following analysis of Poole's ``broken-arm'' example \cite{Poole89}.
\xam\label{brokenarm}
Suppose we have predicates $\UL$,
$\BL$, $\UR$, $\BR$, indicating, respectively, that the left arm is usable,
the left arm is broken, the right arm is usable, and the right arm is
broken.  Let $\KBarm'$ consist of the statements
\begin{itemize}
  \item $\prop{\UL(x)}{x} \aeq_1 1$, $\cprop{\UL(x)}{\BL(x)}{x} \aeq_2 0$
        (left arms are typically usable, but not if they are broken),
  \item $\prop{\UR(x)}{x} \aeq_3 1$, $\cprop{\UR(x)}{\BR(x)}{x} \aeq_4 0$
        (right arms are typically usable, but not if they are broken).
\end{itemize}
Now, consider $\KBarm = (\KBarm' \land (\BL(\Eric) \lor \BR(\Eric)))$;
that is, we know that Eric has a broken arm.  Poole observes that if
we use Reiter's default logic, there is precisely one extension of
$\KBarm$, and in that extension, both arms are usable.
However,
it can be shown that
$\KBarm' \land \BL(\Eric) \rwent \neg \UL(\Eric)$
(see Theorem~\ref{direct} below) and hence (using {\em Right Weakening})
that
$\KBarm' \land \BL(\Eric) \rwent \neg \UL(\Eric) \lor \neg \UR(\Eric)$;
the same conclusion is obtained from
$\KBarm' \land \BR(\Eric)$.  By the {\em Or\/} rule, it follows that
$\KBarm \rwent \neg \UL(\Eric) \lor \neg \UR(\Eric)$.  Using similar
reasoning, we can also show that $\KBarm \rwent \UL(\Eric) \lor
\UR(\Eric)$.  By applying the {\em And\/} rule, we conclude by default
{from} $\KBarm$ that exactly one of Eric's arms is usable,
but we draw no conclusions as to which one it is.
\exam

The final property mentioned in Section~\ref{KLM} is
{\em Rational Monotonicity}. Recall that {\em Rational Monotonicity\/}
asserts
that if $\KB \rwent \phi$ and $\KB \notrwent \neg \theta$ then
$(\KB \land \theta) \rwent \phi$.
Random worlds satisfies a weakened form of {\em Rational
  Monotonicity}. In particular, it satisfies {\em Rational
  Monotonicity\/} except in those situations where limits fail to
exist.%
\footnote{As we discuss later in Section~\ref{competing.results} there
  are often intuitive reasons for the non-existence of limits.}
If $\priw(\phi|\KB\land\theta)$ does exist it must be equal to 1,
i.e., we must have $(\KB \land \theta) \rwent \phi$ as desired.
Sometimes, however, this limit does not exist. Note that the
assumption that $\KB \rwent \phi$ entails that $\priw(\phi|\KB)$
exists.  But {\em Rational Monotonicity\/}'s other assumption, that
$\KB \notrwent \neg \theta$ holds if either $\priw(\theta|\KB)$ has a
value less than one {\em or\/} if this degree of belief does not
exist.  It is the latter ``incompatibility'' of $\theta$ with $\KB$
that is a potential source of problems.  In this case the combination
of $\KB$ and $\theta$ may fail to assign a
limiting degree of belief to $\phi$ even though $\KB$ by itself did.
The following theorem summarizes the status of {\em Rational
Monotonicity\/} in the random-worlds approach.
 
\thm
\label{rational.mon}
Assume that $\KB \rwent \phi$ and $\KB \notrwent \neg\theta$.  Then
$\KB \land \theta \rwent \phi$ provided that $\priw(\phi|\KB \land
\theta)$ exists.  Moreover, a sufficient condition for $\priw(\phi|
\KB \land \theta)$ to exist is that $\priw(\theta|\KB)$ exists.
\ethm
\prf Longer proofs, including the proof of this result, are in the
appendix.
\eprf

\subsection{Specificity and inheritance in random worlds}

One way of using random worlds is to
derive
conclusions about particular
individuals, based on general statistical knowledge.  This is, of course,
the type of reasoning reference-class theories were designed to deal
with.  Recall, these theories aim to discover a single
piece of data---the statistics for a single reference class---that
summarizes all the relevant information.
This idea is also useful in default reasoning, where we sometimes want
to find a single appropriate default. Random worlds rejects this
idea as a general approach, but supports it as a valuable heuristic in
special cases.
 
In this section, we give two theorems covering some of the cases
where random worlds agrees with the basic philosophy of reference
classes. Both results concern {\em specificity\/}---the idea
of using the ``smallest'' relevant reference class for which
we have statistics.  However, both results also allow some
indifference to irrelevant information.  In particular, the second
theorem also covers certain forms of {\em inheritance\/} (as described
in Section~\ref{nonmon.inheritance}).  The results cover almost all of
the noncontroversial applications of specificity and inheritance that
we are aware of, and do not seem to suffer from any of the commonly found
problems such as the disjunctive reference class problem (see
Section~\ref{identify.ref}). Because our theorems are derived
properties rather than postulates, consistency is assured and
there are no {\em ad hoc\/} syntactic restrictions on
the choice of possible reference classes.
We remark that Shastri \cite{Shastri} has also observed that irrelevance
properties hold in his framework.

Our first, and simpler, result is {\em basic direct
inference}, where we have a single reference class that is precisely
the ``right one''. That is, assume that the assertion $\psi(c)$
represents everything the knowledge base tells us about the constant
$c$.  In this case, we can view the class defined by $\psi(x)$ as the
class of all individuals who are ``just like $c$''.  If we have
adequate statistics for the class $\psi(x)$, then we should clearly
use this information.  For example, assume that all we know about
Eric is that he exhibits jaundice, and let $\psi$ represent the class of
patients with jaundice. If we know that 80\% of patients
with jaundice exhibit hepatitis, then basic direct inference would
dictate a degree of belief of $0.8$ in Eric having hepatitis.  We
would, in fact, like this to hold regardless of any other information
we might have in the knowledge base.  For example, we may know the
proportion of hepatitis among patients in general, or that patients with
jaundice and fever typically have hepatitis.  But if all we know about Eric
is that he has jaundice, we would still like to use the statistics for
the class of patients with jaundice,
regardless of the additional information.

Our result essentially asserts the following: ``If we are interested
in obtaining a degree of belief in $\phi(c)$, and the $\KB$ is of the
form $\psi(c) \land \cprop{\phi(x)}{\psi(x)}{x} \aeq \alpha \land
\KBp$, then conclude that $\priw(\phi(c)|\KB) = \alpha$.''
(Here, $\KBp$ is simply intended to denote the rest of $\KB$, whatever
it may be.)
Clearly,
in order for the result to hold, we must make certain assumptions.
The assumptions we consider can be viewed as ensuring that
$\psi(c)$ represents all the information we have about $c$.
First, for obvious reasons, we require that
$\KBp$ does not mention $c$.
However, this is not enough; we also need to assume that
$c$ does not appear in either $\phi(x)$ or $\psi(x)$. To
understand why $c$ cannot appear in $\phi(x)$, suppose that $\phi(x)$
is $Q(x) \lor x = c$, $\psi(x)$ is $\true$, and the $\KB$ is
$\cprop{\phi(x)}{\true}{x} \aeq_1 0.5$.  If the result held in this case,
we would
erroneously conclude that $\priw(\phi(c) |\KB) = 0.5$.  But
since $\phi(c)$ holds tautologically, we actually obtain
$\priw(\phi(c)|\KB) = 1$.  To see why the constant $c$ cannot appear
in $\psi(x)$, suppose that $\psi(x)$ is $(P(x) \land x \ne c) \lor
\neg P(x)$, $\phi(x)$ is $P(x)$, and the $\KB$ is $\psi(c) \land
\cprop{P(x)}{\psi(x)}{x} \aeq_2 0.5$.  Again, if the result held,
we would be able to conclude that $\priw(P(c) | \KB) = 0.5$.  But
$\psi(c)$ is equivalent to $\neg P(c)$, so in fact $\priw(P(c) | \KB)
= 0$.
 
As we now show, these assumptions suffice to guarantee the desired result.
In fact, the theorem generalizes the basic principle
to properties and classes dealing with more than one individual at a
time (as is demonstrated in some of the examples following the theorem).
In the following, let $\xtuple = \{x_1,\ldots,x_k\}$ and $\ctuple =
\{c_1,\ldots,c_k\}$ be sets of distinct variables and distinct
constants, respectively.
Furthermore, we use $\phi(\xtuple)$ to indicate that all of the free
variables in the formula $\phi$ are in $\xtuple$, and we
use $\phi(\ctuple)$ to denote the new formula formed by substituting
each $x_i$ by $c_i$ in $\phi$. Note that $\phi$ may contain other
constants not among the
$c_i$'s; these are unaffected by the substitution.
 
\thm
\label{direct}
Let $\KB$ be a knowledge base of the form $\psi(\ctuple) \land \KBp$,
and assume that for all sufficiently small tolerance vectors $\epsvec$,
$$
\KB[\epsvec] \sat \cprop{\phi(\xtuple)}{\psi(\xtuple)}{\xtuple} \in
[\alpha,\beta].
$$
If no constant in $\ctuple$ appears in $\KBp$, in $\phi(\xtuple)$, or
in $\psi(\xtuple)$, then $\priw(\phi(\ctuple)|\KB) \in
[\alpha,\beta]$,
provided the degree of belief exists.%
\footnote{The degree of belief may not exist since
$\lim_{\epsvec \tendsto \zerovec}
\liminf_{N \tendsto \infty}\, \prNw(\phi|\KB)$ may not be equal to
$\lim_{\epsvec \tendsto \zerovec}
\limsup_{N \tendsto \infty}\, \prNw(\phi|\KB)$.  However, it follows
{from} the proof of the theorem that both
these limits
lie in the interval $[\alpha,\beta]$.
A similar remark holds for
many of our later results.}
\ethm
\prf
See the appendix.
\eprf

Theorem~\ref{direct} refers to any statistical information about
$\cprop{\phi(\xtuple)}{\psi(\xtuple)}{\xtuple}$ that can be
inferred from the knowledge base.  An important special case is when
the knowledge base  contains the relevant
information explicitly.
\cor
\label{direct.range}
Let $\KBp$ be the conjunction
$$
\psi(\ctuple) \land \left( \alpha \aleq_i
\cprop{\phi(\xtuple)}{\psi(\xtuple)}{\xtuple} \aleq_j \beta \right).
$$
Let $\KB$ be a knowledge base of the form $\KBp \land \KBpp$ such
that no constant in $\ctuple$ appears in $\KBpp$, in $\phi(\xtuple)$,
or in $\psi(\xtuple)$. Then, if the degree of belief exists, we have
$$\priw(\phi(\ctuple)|\KB) \in [\alpha,\beta].$$
\ecor
\prf
Let $\epsilon > 0$, and let $\epsvec$ be sufficiently small so that
$\eps_i,\eps_j < \epsilon$.  For this $\epsvec$, the formula
$(\alpha \aleq_i \cprop{\phi(\xtuple)}{\psi(\xtuple)}{\xtuple}
\aleq_j \beta)$ implies $\cprop{\phi(\xtuple)}{\psi(\xtuple)}{\xtuple}
\in [\alpha-\epsilon,\beta+\epsilon]$.  Therefore, by
Theorem~\ref{direct}, $\priw(\phi(\ctuple)|\KB) \in
[\alpha-\epsilon,\beta+\epsilon]$.  But since this holds for any
$\epsilon > 0$, it is necessarily the case that
$\priw(\phi(\ctuple)|\KB) \in [\alpha,\beta]$.
\eprf

It is interesting to note one way in which
this result diverges from the reference-class paradigm.
Suppose we consider a query $\phi(c)$, and that our
knowledge base $\KB$ is as in the hypothesis of
Corollary~\ref{direct.range}. While we can indeed conclude that
$\priw(\phi(\ctuple)|\KB) \in [\alpha,\beta]$, the exact value of the
degree of belief within this interval
depends on the other information in the knowledge base.
Thus, while random worlds
certainly uses the
information
$\alpha \aleq_i \cprop{\phi(x)}{\psi(x)}{x} \aleq_j \beta$,
it does not necessarily ignore the rest of the knowledge base altogether.
On the other hand,
if the interval $[\alpha,\beta]$ is sufficiently small (and, in
particular, when $\alpha = \beta$),
then we may not care exactly where in the interval the degree of
belief lies.  In this case, we can ignore all the information in $\KBp$,
and use the single piece of ``local'' information for computing
the degree of belief.
We now present a number of examples that demonstrate the behavior of the
direct inference result.
\xam\label{exam.hep}
Consider a knowledge base
describing the hepatitis example discussed earlier.
In the notation of Corollary~\ref{direct.range}:
$$
\KBhep' = \Jaun(\Eric) \land \cprop{\Hep(x)}{\Jaun(x)}{x} \aeq_1 0.8,
$$
and
$$
\KBhep  =  \KBhep' \land \prop{\Hep(x)}{x} \aleq_2 0.05\; \land
  \cprop{\Hep(x)}{\Jaun(x) \land \Fever(x)}{x} \aeq_2 1.
$$
Then $\priw(\Hep(\Eric) | \KBhep) = 0.8$ as desired;
information about other reference classes
(whether more general or more specific) is
ignored. Other kinds of information are also ignored, for example,
information about other individuals.  Thus, $\priw(\Hep(\Eric) |
\KBhep \land \Hep(\Tom)) = 0.8$.~~\bbox
\end{example}

Although it is nothing but an immediate application of
Theorem~\ref{direct}, it is worth remarking that
the principle of {\em Direct Inference for Defaults\/}
(Section~\ref{nonmon.inheritance}) is satisfied
by random-worlds:
\cor
\label{dirinf.defaults}
Suppose $\KB$ implies $\cprop{\phi(\xtuple)}{\psi(\xtuple)}{\xtuple} \aeq_i 1$,
and no constant in $\ctuple$ appears in $\KB$, $\phi$, or $\psi$.
Then $\priw(\phi(\ctuple)|\KB \land \psi(\ctuple)) = 1$.
\ecor
As discussed in Section~\ref{nonmon.inheritance}, this shows that
simple forms of reasoning about classification hierarchies
are possible.
 
\xam\label{exam.fly}
The knowledge base $\KBfly$ from Section~\ref{nonmon.inheritance} is,
under our interpretation of defaults:
$$ \cprop{\Fly(x)}{\Bird(x)}{x} \aeq_1 1 \; \land
     \cprop{\Fly(x)}{\Penguin(x)}{x} \aeq_2 0 \; \land
     \forall x \, (Penguin(x) \rimp \Bird(x)).
$$
Then $\Prinf(\Fly(\Tweety)|\KBfly \land \Penguin(\Tweety)) = 0$.
That is, we conclude that Tweety the penguin does not fly, even
though he is also a bird and birds generally do fly.
\exam
 
Given this preference for the most specific reference class, one
might wonder why random worlds does not encounter the problem of {\em
disjunctive reference classes\/} (see Section~\ref{identify.ref}).  The
following example, based on the example from Section~\ref{identify.ref},
provides one answer.
\xam
\label{disjref.bad}
Recall the knowledge base $\KBhep'$ from the hepatitis example above,
and consider the disjunctive reference class $\psi(x) \eqdef \Jaun(x)
\land (\neg \Hep(x) \lor x = \Eric)$. Clearly, as the domain size grows
large, $\cprop{\Hep(x)}{\psi(x)}{x}$ becomes
arbitrarily close to 0.%
\footnote{This actually relies on the fact that, with high probability, the
proportion (as the domain size grows) of jaundiced patients
without
hepatitis is nonzero.  We do not prove this fact here; see
\cite{PV,GHK1b}.}
Therefore, for any fixed $\epsilon > 0$
$$ \Prinf \Bigl( \cprop{\Hep(x)}{\psi(x)}{x} \in [0,\epsilon] \Bigm|
    \KBhep' \Bigr) = 1.$$
We can construct a new knowledge base $\KBdishep = \KBhep' \land
\cprop{\Hep(x)}{\psi(x)}{x} \in [0,\epsilon]$. Furthermore, $\KBdishep
\sat \psi(Eric)$. Hence, $\KBdishep$ contains a more specific
reference class for $\Hep(\Eric)$ than $\Jaun(x)$ with very different
statistics. Yet, by Proposition~\ref{cmc}, we know that
$\priw(\Hep(\Eric)|\KBhep') = \priw(\Hep(\Eric)|\KBdishep)$, and in
Example~\ref{exam.hep} we showed this to be equal to 0.8. So random
worlds avoids using the spurious disjunctive class $\psi(x)$ even in a
knowledge base that explicitly includes statistics from this class.
Theorem~\ref{direct} does not apply here because the class $\psi(x)$
explicitly mentions the
constant $\Eric$.
Another way of seeing that the class $\psi(x)$ does not
affect the random-worlds computation is
to observe
that its statistics are not
informative, \ie these
statistics are true in almost all worlds. Hence $\psi(x)$'s statistics
places no constraints on the sets of worlds that determine the degree
of belief.  As we shall see in Example~\ref{disjref.good}, when we do
have informative statistics for a class, those statistics can be used,
even if the class is disjunctive.
\exam
 
As we have said, we are not limited to unary predicates, nor to
examining only one individual at a time.
\xam
In Example~\ref{elephants}, we showed how to formalize the
elephant-zookeeper example discussed in
Section~\ref{nonmon.expressivity}. As we now show, the natural
representation of $\KBel$ indeed yields the answers we expect.
We
consider two queries. First, assume that we are interested in finding
out whether Clyde likes Eric. In this case, we can use the class of
pairs $\psi(x,y)= \Elephant(x) \land
\Zookeeper(y)$.
Applying
Corollary~\ref{dirinf.defaults} to the first default in $\KBel$,
we can conclude that
$\priw(\Likes(\Clyde,\Eric)|\KBel) = 1$. Second, we examine whether or
not Clyde likes Fred.
Applying
Corollary~\ref{dirinf.defaults} to the second default in $\KBel$,
we can conclude that
$\priw(\Likes(\Clyde,\Fred)|\KBel) = 0$. Note that we cannot
apply
Corollary~\ref{dirinf.defaults} to the first default in $\KBel$
to conclude that Clyde likes Fred.
The
conditions of the corollary are violated, because the constant Fred is
used elsewhere in the knowledge base.
\exam
 
The same principles continue to hold for more complex sentences; for example,
we can mix first-order logic and statistical knowledge arbitrarily and
we can nest defaults.
 
\xam
In Example~\ref{tall.parent}, we showed how to express the default:
``People who have at least one tall parent are typically tall.''
If we have this default, and also know that $\exists y \, (\Child(\Alice,y)
\land \Tall(y))$ (Alice has a tall parent), Corollary~\ref{dirinf.defaults}
tells us that we can conclude by default that $\Tall(\Alice)$.
\exam
 
\xam
In Example~\ref{sleeplate}, we showed how the default ``Typically, people
who normally go to bed late normally rise late'' can be expressed in our
language using the knowledge base $\KBlate$.  Let $\KBlate'$ be
$$
\KBlate \ \land\;\;
\cprop{\Sleepslate(\Alice,y')}{\Day(y')}{y'} \aeq_2 1 .
$$
By Corollary~\ref{dirinf.defaults}, Alice typically rises late.  That is,
$$
\priw(\cprop{\Riseslate(\Alice,y)}{\Day(y)}{y} \aeq_1 1\ | \KBlate') = 1.
$$
By {\em Cautious Monotonicity\/} and {\em Cut\/}, we can add this conclusion
(which is itself a default) to $\KBlate'$.
By Corollary~\ref{dirinf.defaults} again, we then conclude that
Alice can be expected to rise late on any particular day,
say $\SomeMorning$.
So, for
instance:
 $$\priw(\Riseslate(\Alice,\SomeMorning)|\KBlate'
                                         \land \Day(\SomeMorning)
        ) = 1.\;\;\; \bbox$$
\end{example}
 
In all the examples presented so far in this section, we have
statistics for precisely the right reference
class to match our
knowledge about the individual(s) in question; Theorem~\ref{direct}
and its corollaries require this.
Unfortunately, in many cases our statistical information is not
detailed enough for Theorem~\ref{direct} to apply.
Consider the knowledge base $\KBhep$ from the hepatitis example.
Here we have statistics for the occurrence of hepatitis among the
class of patients who are just like Eric, so we can use these to
induce a degree of belief in $\Hep(\Eric)$.
But now consider the knowledge base $\KBhep \land \Tall(\Eric)$.
Since we do not have statistics for the
frequency of
hepatitis among tall patients, the results we have seen
so far do not apply. We would like to be able to ignore
$\Tall(\Eric)$.  But what entitles us to ignore $\Tall(\Eric)$ and
not $\Jaun(Eric)$?
{T}o solve this problem in complete generality requires a better
theory of irrelevance than we currently have.
Nevertheless, our next theorem covers many cases, including
many of the
less controversial examples found in the
default reasoning literature.
 
The theorem we present deals with a knowledge base $\KB$ that defines
a ``minimal'' reference class $\psi_0$ with respect to the query
$\phi(c)$. More precisely, assume that $\KB$ gives statistical
information regarding $\cprop{\phi(x)}{\psi_i(x)}{x}$ for a number
of different
classes $\psi_i(x)$.  Further suppose that, among these
classes, there is one class $\psi_0(x)$ that is {minimal\/}---all
other
classes are strictly larger or
entirely disjoint from it.
Our result states that if we also know $\psi_0(c)$, we can use the
statistics for
$\cprop{\phi(x)}{\psi_0(x)}{x}$ to induce a degree of belief in
$\phi(c)$. What makes this such an interesting result is that we
are allowed to know {\em more\/} about $c$ than just $\psi_0(c)$;
any extra information will be treated as being irrelevant.
This pattern of reasoning is best explained using an example:

\xam
Assume we have a knowledge base $\KBtax$ containing information about
birds and animals; in particular, $\KBtax$ contains a taxonomic
hierarchy of this domain.  Moreover, $\KBtax$ contains the following
information about the swimming ability of various types of animals:
$$  \begin{array}{llll}
\cprop{\Swims(x)}{\Penguin(x)}{x} & \aeq_1 & 0.9 & \land \\
\cprop{\Swims(x)}{\Sparrow(x)}{x} & \aeq_2 & 0.01 & \land \\
\cprop{\Swims(x)}{\Bird(x)}{x} & \aeq_3 & 0.05 & \land \\
\cprop{\Swims(x)}{\Animal(x)}{x} & \aeq_4 & 0.3 & \land \\
\cprop{\Swims(x)}{\Fish(x)}{x} & \aeq_5 & 1. &
  \end{array} $$
If we also know that Opus is a penguin, then in order to determine
whether Opus swims the best reference class is surely the
class of penguins.  The remaining classes are either larger
(in the case of birds or animals), or disjoint (in the case of
sparrows and fish). This is the case even if we know that Opus is
a black penguin with a large nose. That is, Opus {\em inherits\/} the
statistics for the minimal class $\psi_0$---penguins---even though
the class of individuals just like Opus is smaller than $\psi_0$.
\exam
That random-worlds validates this intuition is formalized
in the next theorem.
This theorem requires that
no symbol in $\phi(x)$ appear in the knowledge base other
than in statistics of the form $\cprop{\phi(x)}{\psi(x)}{x}$ for various
$\psi(x)$. This is necessary for our assumption of a
unique minimal reference class to be a practical one.
Suppose that,
in violation of this condition,
the knowledge base contains $\forall x (\psi(x) \rimp \phi(x))$. Clearly
$\psi(x)$ is in fact a reference class for $\phi(x)$ (where
the statistic is 100\%). But if we identify reference classes only by
looking for terms of the form $\cprop{\phi(x)}{\psi(x)}{x}$, we will
not notice this. Obviously the minimality assumption
needs to consider all reference
classes, irrespective of syntactic form. But because
first-order logic provides many subtle
and nonobvious ways to constrain statistics relating to $\phi(x)$,
we simplify the issue by assuming that the only
mention of information that might be related to $\phi(x)$ is contained
in explicit statistical assertions.
Of course, it would be very interesting to find a result that
addresses cases in which this assumption is not true.
\thm
\label{inherit.thm}
Let $c$ be a constant and let $\KB$ be a knowledge base satisfying the
following conditions:
\begin{enumerate}
\item[(a)]
\label{psizero}
$\KB \sat \psi_0(c)$,
\item[(b)]
\label{leastrefclass}
for any expression of the form
$\cprop{\phi(x)}{\psi(x)}{x}$ in $\KB$, it is the
case that either $\KB \sat \forall x ( \psi_0(x) \rimp \psi(x))$ or that $\KB \sat
\forall x ( \psi_0(x) \rimp \neg \psi(x))$,
\item[(c)]
\label{nophi}
the (predicate, function, and constant)
symbols in $\phi(x)$ appear in $\KB$ only on the left-hand side
of the conditionals in the
proportion expressions described in condition~(b),
\item[(d)]
\label{noc}
the constant $c$ does not appear in the formula $\phi(x)$.
\end{enumerate}
Assume that for all sufficiently small tolerance vectors $\epsvec$:
$$
\KB[\epsvec] \sat \cprop{\phi(x)}{\psi_0(x)}{x} \in [\alpha,\beta].
$$
Then $\priw(\phi(c)|\KB) \in [\alpha,\beta]$,
provided the degree of belief exists.
\ethm
\prf
See the appendix.
\eprf
 
Again, the following analogue to Corollary~\ref{direct.range} is immediate:
\cor
\label{inherit.range}
Let $\KBp$ be the conjunction
$$
\psi_0(c) \land
(\alpha \aleq_i \cprop{\phi(x)}{\psi_0(x)}{x} \aleq_j \beta).
$$
Let $\KB$ be a knowledge base of the form $\KBp \land \KBpp$ that
satisfies conditions~(b), (c), and (d)
of Theorem~\ref{inherit.thm}. Then,
if the degree of belief exists,
$$\priw(\phi(c)|\KB) \in [\alpha,\beta].$$
\ecor
 
This theorem and corollary have many useful applications.
\xam
Consider the knowledge bases $\KBhep'$ and $\KBhep$ concerning jaundice and
hepatitis from Example~\ref{exam.hep}.  In that example,
we supposed that
the only information about Eric contained in the knowledge base was that
Eric has jaundice.
It is clearly more realistic to assume that Eric's hospital records
contain more information than just this fact.  Theorem~\ref{inherit.thm}
allows us to
ignore this information in a large number of cases.
 
For example, $$\priw(\Hep(\Eric)
| \KBhep' \land \Fever(\Eric) \land \Tall(Eric)) = 0.8,$$
as desired.
On the other hand,
$$\priw(\Hep(\Eric) | \KBhep \land \Fever(\Eric) \land \Tall(Eric))= 1.$$
(Recall that $\KBhep$ includes $\cprop{\Hep(x)}{\Jaun(x) \land
\Fever(x)}{x} \aeq_2 1$,
while $\KBhep'$ does not.)
This
shows why it is important that we identify the
most specific reference class for the query $\phi$.
The most specific reference statistic for $\Hep(Eric)$ with
respect to $\KBhep' \land \Fever(\Eric) \land \Tall(\Eric)$
is $\cprop{\Hep(x)}{\Jaun(x)}{x} \aeq_1 0.8$, while with respect
to $\KBhep \land \Fever(\Eric) \land \Tall(\Eric)$ it is
$\cprop{\Hep(x)}{\Jaun(x) \land \Fever(x)}{x} \aeq_2 1$.
In the latter case, the less-specific reference classes $\Jaun$ and
$\true$ are ignored,
and in both cases Theorem~\ref{inherit.thm} allows us to ignore the
extra information $\Tall(\Eric)$.
Note that the theorem does not allow us to conclude that
$$\priw(\Hep(\Eric) | \KBhep \land \Tall(Eric))= 0.8.$$
The class $\Jaun$ is no longer the unique most specific reference
class, since we also have statistics for the more specific class
$\Jaun \land \Fever$.  Nevertheless, this conclusion is, in fact,
reached by random worlds.
\exam

As discussed in Section~\ref{nonmon.inheritance},
various inheritance properties are considered desirable in
default reasoning as well.
{T}o begin with, we note that Theorem~\ref{inherit.thm} covers
the simpler cases (which can also be seen as applications of
rational monotonicity):
\xam
In simple cases, Theorem~\ref{inherit.thm} shows that random worlds
is able to apply defaults in the presence of ``obviously irrelevant''
additional information.  For example, using the
knowledge base $\KBfly$ (see Example~\ref{exam.fly}):
 $$\priw( \Fly(\Tweety) | \KBflyp \land \Penguin(\Tweety) \land
    \Yellow(Tweety)) = 0.$$
That is, Tweety the yellow penguin
is still not able to fly.
\exam
Theorem~\ref{inherit.thm} also validates
more difficult reasoning patterns that have caused problems for
many default reasoning theories. In particular,
we validate exceptional-subclass inheritance, in which
a class that is exceptional in one respect can nevertheless
inherit other unrelated properties:
 
\xam
If we consider the property of warm-bloodedness as well as flight, we get:
$$ \priw\left( \Warmblooded(\Tweety) \left|
             \begin{array}{l}
             \KBfly \land \Penguin(\Tweety) \;\land \\
             \cprop{\Warmblooded(x)}{\Bird(x)}{x} \aeq_3 1
            \end{array} \right.
      \right) = 1. $$
Knowing that Tweety does not fly because he is a penguin does
not prevent us from assuming that he is like typical birds in other
respects.
\exam
The drowning-problem variant of the exceptional-subclass inheritance problem
is also covered by the theorem.
\xam Suppose we know, as in Section~\ref{nonmon.inheritance}, that
yellow things tend to be highly visible. Then:
$$  \priw\left( \Visible(\Tweety) \left|
            \begin{array}{l}
            \KBfly \land \Penguin(\Tweety) \land \Yellow(Tweety) \;\land \\
            \cprop{\Visible(x)}{\Yellow(x)}{x} \aeq_3 1
           \end{array} \right.
     \right) = 1. $$
Here, all that matters is that Tweety is a yellow object. The fact
that he is a bird, and an exceptional bird at that, is rightly ignored.
\exam

Notice that, unlike Theorem~\ref{direct}, the conditions of
Theorem~\ref{inherit.thm} do not extend to inferring degrees of
belief in $\phi(\ctuple)$, where $\ctuple$ is a tuple of constants.
Roughly speaking, the reason
lies in the ability of the language to create
connections between
different constants in the tuple.  For
example,
let $\KBp$ be
$\prop{\Hep(x) \land \neg \Hep(y)}{x,y} \aeq_1 0.2$. By
Theorem~\ref{direct}
(taking $\psi_0(x_1,x_2)$ to be $\true$),
$\priw(\Hep(\Tom) \land \neg \Hep(\Eric) | \KBp)
= 0.2$.  But, of course,
$\priw(\Hep(\Tom) \land \neg \Hep(\Eric) |  \KBp \land \Tom = \Eric) = 0$.
The additional information regarding $\Tom$ and $\Eric$ cannot be
ignored.
A version of Theorem~\ref{inherit.thm} where we replaced $c$ by
$\ctuple$ would incorrectly attempt to ignore this information.
This example might suggest that this is a
problem related only to the use of equality, but more complex examples
that do not mention equality can also be constructed.
 
As a final example in this section, we revisit the issue of
disjunctive reference classes.  As we saw in
Example~\ref{disjref.bad}, random worlds does not suffer from the
``disjunctive reference class'' problem.
In
Section~\ref{identify.ref},
we observed that
some systems
avoid this problem by
simply outlawing disjunctive reference classes,
which is problematic, as such classes are sometimes useful.
The next example
demonstrates that random worlds does, in fact, treat disjunctive
reference classes appropriately.
\xam
\label{disjref.good}
Recall that
in Section~\ref{identify.ref} we gave an example
involving disjunctive reference classes
for Tay-Sachs
disease.  The corresponding
statistical information was represented, in our framework, as the
knowledge base $\KB$: $$
\cprop{\TS(x)}{\EEJ(x) \lor \FC(x)}{x} \aeq_1 0.02.
$$ Given a baby \Eric\ of eastern-European extraction,
Theorem~\ref{inherit.thm} shows us that $$
\priw(\TS(\Eric) | \KB \land \EEJ(\Eric)) = 0.02.
$$ That is, random worlds is able to use the information derived from
the disjunctive reference class, and apply it to an individual known
to be in the class; indeed,
through inheritance it also deals with the case where we have
additional information determining to which of the two populations
this specific individual belongs.
Thus, disjunctive reference classes are treated
in the same manner as other potential reference classes.
\exam

The type of specificity and inheritance reasoning covered by the
results in this section are special cases of general inheritance reasoning.
While these theorems show that random worlds
does support many noncontroversial instances of such
reasoning, proving a more general theorem asserting this claim is
surprisingly subtle (partly because of the existence of numerous
divergent semantics and intuitions for inheritance reasoning
\cite{clash.intuitions}).  We are currently working towards stating and
proving such a
general claim, for the case in which we have an inheritance hierarchy
of defaults and universal implications.
On the other hand, it is easy to see
that random worlds
does not validate general inheritance reasoning in an arbitrary
statistical context (\ie where some statistical values are less than 1,
and so do not state defaults). We discuss why this happens below,
in Example~\ref{moody.magpie}, and argue that we should not want
simple inheritance in all contexts anyway.

\subsection{Competing reference classes}
\label{competing.results}
In previous sections we have always been careful to consider
examples in which there was an obviously ``best'' reference
class. In practice, we will not always be this fortunate.
Reference-class theories usually cannot give useful answers
when there are competing candidates for the best class.
However, random worlds does not have this problem, because the degrees
of belief it defines can be combinations of the values for competing
classes. In this section we examine, in very general terms,
three types of competing information. The first concerns
conflicts between specificity and accuracy, the second between
information that is too specific and information that is too general,
and the last
between incomparable reference classes, so that the specificity
principle is not applicable.
 
We discussed the conflict between specificity and accuracy in
Section~\ref{competing}. This problem was noticed by Kyburg, who
addresses this issue with his
strength rule.
In Section~\ref{competing}, we argued that,
to assign a degree
of belief to $\Chirps(\Tweety)$, we should be able to use the tighter
interval $[0.7,0.8]$ even though it is associated with a less specific
reference class.  As we observed, Kyburg's strength rule
attempts to capture this intuition.
As the following result shows, the random worlds method also captures
this intuition
(without requiring any special rules),
at least when the reference classes form a chain.%
\footnote{Kyburg's rule also
applies to cases where the reference classes do not form a chain.
The random-worlds method disagrees with the strength rule in these
cases.  For example, if we know that only 20\% of Republicans are pacifists,
that only 20\% of bankers are pacifists, and that Morgan is a Republican
banker, Kyburg's strength rule would conclude that our degree of
belief that Morgan is a pacifist is $0.2$.  On the other hand, the
random-worlds method would view this as two pieces of evidence counting
against Morgan being a pacifist; it can be shown that this would
result in a degree of belief less  than $0.2$.
}
 
\thm\label{strength}
Suppose $\KB$ has the form
$$\bigwedge_{i = 1}^m
(\alpha_i \aleq_{\ell_i} \cprop{\phi(x)}{\psi_i(x)}{x} \aleq_{r_i} \beta_i)
\; \land \; \psi_1(c) \; \land \; \KBp,$$
and, for all $i$, $\KB \sat \forall x\; (\psi_i(x) \rimp \psi_{i+1}(x))
\land \neg (\prop{\psi_1(x)}{x} \aeq_1 0)$.
Assume also that no symbol appearing in $\phi(x)$ appears in
$\KBp$ or in any $\psi_i(c)$. Further suppose that, for some $j$,
$[\alpha_j, \beta_j]$ is the tightest interval.
That is, for all $i\neq j$, $\alpha_i < \alpha_j < \beta_j < \beta_i$.
Then,
if it exists,
$$\priw(\phi(c) | \KB) \in [\alpha_j, \beta_j].$$
\ethm
\prf
See the appendix. %
\eprf
\xam
The example described in Section~\ref{competing} is essentially
captured by the following knowledge base $\KBchirps$:
$$
\begin{array}{l}
    0.7 \aleq_1 \cprop{\Chirps(x)}{\Bird(x)}{x} \aleq_2 0.8 \; \land \\
   0 \aleq_3 \cprop{\Chirps(x)}{\Magpie(x)}{x} \aleq_4 0.99 \; \land \\
    \forall x\, (\Magpie(x) \rimp \Bird(x)) \; \land \\
    \Magpie(\Tweety).
\end{array}
$$
It follows from Theorem~\ref{strength} that $\priw(\Chirps(\Tweety)|
\KBchirps) \in [0.7,0.8]$.%
\footnote{Strictly speaking, a direct application of
Theorem~\ref{strength} would require that $\KBchirps$ contains
$\neg(\prop{\Magpie(x)}{x} \aeq_i 0)$. But the maximum-entropy techniques of
Section~\ref{GMPsec} can be used to show that this actually follows
by default. Hence,
by Proposition~\ref{cmc}, we can consider this to be in $\KBchirps$.}
\exam
 
Next, we consider a different way in which competing reference
classes can arise:~when one reference class is too specific, and the
other too general.
\xam\label{moody.magpie}
We illustrate the problem with a example based on one of Goodwin's
\cite{Goodwin}. Consider $\KBmagpie$:
$$
\begin{array}{l}
  \cprop{\Chirps(x)}{\Bird(x)}{x} \aeq_1 0.9 \; \land \\
  \cprop{\Chirps(x)}{\Magpie(x) \land \Moody(x)}{x} \aeq_2 0.2 \;  \land \\
  \forall x (\Magpie(x) \rimp \Bird(x)) \; \land \\
  \Magpie(\Tweety).
\end{array}
$$
Reference class theories would typically ignore the information
about moody magpies: since Tweety is not known to be moody,
the class of moody magpies is not even a legitimate reference class
in these theories.
Using such approaches, the degree of belief would be 0.9.
Goodwin argues that this is not completely reasonable---why should
we ignore the information about moody magpies? Tweety could
be moody (the knowledge base leaves the question open). In fact, it
 is consistent with $\KBmagpie$ that magpies are generally moody.
But ignoring the second statistic in effect amounts to assuming that
magpies generally are not moody. It seems hard to see why this is
a reasonable assumption. The random-worlds approach supports Goodwin's
intuition, and the degree of belief that Tweety flies, given
$\KBmagpie$, can be shown to have a value which is less than 0.9.
\exam
 
This example illustrates a general phenomenon: if we do not have
a ``most appropriate'' reference class
(in the sense of Theorem~\ref{direct}), then
random worlds combines information from more specific
classes as well as from more general classes.
Hence, as we mentioned in the previous section, random worlds does not
always validate inheritance reasoning:
pure inheritance reasoning would always look to superclasses and
ignore subclasses.
We agree with Goodwin that this property of pure inheritance reasoning
can lead to
unintuitive conclusions, especially when we are dealing with
quantitative information.

The third and most important type of conflict is when we have
incomparable
reference classes.
As we argued in Section~\ref{competing}, this case is
likely to come up often in practice.
While the
complete characterization of the
behavior of random worlds in such
cases would appear to be complex, the following theorem illustrates
what happens when the competing references classes are essentially
disjoint.  We capture ``essentially disjoint'' here by assuming
that the overlap between these classes
consists of precisely one member:~the individual
$c$ addressed in our query.
We can generalize the following theorem to the
case where we simply assume that the overlap between competing
reference classes $\psi$ and $\psi'$ is small relative to the sizes
of the two classes; that is, where
$\cprop{\psi(x) \land \psi'(x)}{\psi(x)}{x} \aeq 0$ and
$\cprop{\psi(x) \land \psi'(x)}{\psi'(x)}{x} \aeq 0$.
For simplicity, we omit
the details of this extension here.
 
It turns out that, under this assumption, random worlds provides an
independent derivation of a well-known technique for combining
evidence: Dempster's rule of combination \cite{Shaf}.
Dempster's rule addresses the issue of combining {\em
independent\/} pieces of evidence.  Consider a query $P(c)$, and
assume we have competing reference classes that are all appropriate
for this query.  In this case, the different pieces of evidence are
the proportions of the property $P$ in the different competing
reference classes.  More precisely,
if $\psi_i(c)$ holds,
we can view the fact that
$\cprop{P(x)}{\psi_i(x)}{x} \aeq \alpha_i$ as giving evidence
of weight $\alpha_i$ in favor of $P(c)$. The fact that the intersection
between the different classes is ``small'' means that almost disjoint
samples were used to compute these pieces of evidence; thus, it
is perhaps reasonable to view them
as being independent.  Under this interpretation, Dempster's rule
tells us how to combine the different pieces of evidence to obtain an
appropriate degree of belief in $P(c)$.  The function used in
Dempster's rule is
$\dempster : [0,1]^m \rightarrow [0,1]$, defined as
follows:
$$
  \dempster(\alpha_1,\ldots,\alpha_m) =
    \frac{\prod_{i=1}^m \alpha_i}
         {\prod_{i=1}^m \alpha_i + \prod_{i=1}^m (1-\alpha_i)}.
$$
As the following theorem shows, this is also the answer obtained by
random worlds.
Since $\dempster$ is undefined if some $\alpha_i$ are equal to 1
while others are equal to 0, we assume that this is not the case in
the theorem.

\thm\label{dempster.rule}
Let $P$ be a unary predicate, and consider a knowledge base $\KB$ of the
following form:%
\footnote{Here, $\exists! x$ stands for ``there exists a unique $x$ such
that\ldots{}''.}
$$
\band_{i=1}^m
  \left(\cprop{P(x)}{\psi_i(x)}{x} \aeq_i \alpha_i \land \psi_i(c)\right)
 \ \land\
\band_{\stackrel{i,j = 1}{\scriptstyle i \neq j}}^m
       \exists! x (\psi_i(x) \land \psi_j(x)) \ ,
$$
where either $\alpha_i < 1$ for all $i = 1, \ldots, m$, or
$\alpha_i > 0$ for all $i = 1, \ldots, m$.
Then, if neither $P$ nor $c$ appear anywhere in the formulas
$\psi_i(x)$, then
$$
\priw(P(c) | \KB) = \dempster(\alpha_1,\ldots,\alpha_m).
$$
\ethm
\prf
See the appendix. %
\eprf
 
We illustrate this theorem on what is, perhaps, the most famous
example of conflicting information---the {\em Nixon Diamond\/}
\cite{ReitCris}. Suppose we are interested in assigning a degree of belief
to the assertion ``Nixon is a pacifist''.  Assume that we know that Nixon is
both a Quaker and a Republican, and we have statistical information for the
proportion of pacifists within both classes.  This is an example where we
have two incomparable reference classes for the same query.  More formally,
assume that $\KBnixon$ is
 $$\begin{array}{l}
    \|\pac(x)|\quak(x)\|_x \aeq_1 \alpha \;\land \\
    \|\pac(x)|\repub(x)\|_x \aeq_2 \beta \;\land \\
    \quak(\Nixon) \land \repub(\Nixon) \;\land \\
    \exists! x\qsep (\quak(x) \land \repub(x)) \ ,%
\end{array}
 $$
and that $\phi$ is $\pac(\Nixon)$.%
\footnote{As pointed out above, Theorem~\ref{dempster.rule} can be
generalized so that instead of asserting that Nixon is the only Quaker
Republican, it is sufficient to assert that there are very few
Quaker-Republicans.}
The degree of belief $\priw(\phi|\KBnixon)$ takes different values,
depending on the values $\alpha$ and $\beta$ for the two reference classes.
If $\{\alpha,\beta\} \ne \{0,1\}$, then
$\priw(\phi|\KBnixon)$ always exists and its
value is equal to
$\frac{\alpha \beta}{\alpha \beta + (1-\alpha)(1-\beta)}$.
If, for example,
$\beta = 0.5$, so that the information for Republicans is neutral, we get
that $\priw(\phi|\KBnixon) = \alpha$: the data for Quakers is used to
determine the degree of belief.  If the evidence given by the two reference
classes is conflicting---$\alpha > 0.5 > \beta$---then $\priw(\phi|\KBnixon)
\in [\alpha,\beta]$: some intermediate value is chosen.  If, on the other
hand, the two reference classes provide evidence in the
same direction, then the
degree of belief is
greater than both $\alpha$ and $\beta$.  For example, if $\alpha = \beta =
0.8$, then the
degree of belief
would be around $0.94$.
This has a reasonable explanation: if we have two independent bodies of
evidence, both supporting $\phi$,
when we combine them we
should get even more support for $\phi$.
 
Now, assume that our information is not entirely quantitative.  For example,
we may know that ``Quakers are typically pacifists''.  In our framework,
this corresponds to assigning $\alpha = 1$.  If our information for
Republicans is not a default, so that $\beta > 0$, then
it follows from Theorem~\ref{dempster.rule} that $\priw(\phi|\KBnixon) = 1$.
As expected, a default (\ie an ``extreme''
value) dominates.  But what happens in the case where we have conflicting
defaults for the two reference classes? It turns out that, in this case,
the limiting probability does not exist.
This is because the limit is {\em non-robust\/}: its value depends on
the way in which the tolerances $\epsvec$ tend to 0.  More precisely,  if
$\eps_1 \ll \eps_2$, so that the ``almost all'' in the statistical
interpretation of the first conjunct is much closer to ``all'' than the
``almost none'' in the second conjunct is closer to ``none'', then the
limit is 1.
We can view the magnitude of the tolerance as representing the
strength of the default.  Thus,
in this case, the first conjunct represents a default with higher
priority than the second conjunct.
Symmetrically, if $\eps_1 \gg \eps_2$,
then the limit is 0.  On the other hand, if $\eps_1 = \eps_2$, then the
limit is $1/2$.
 
The nonexistence of this limit is not simply a technical
artifact of our approach.
The fact that we obtain different limiting degrees of belief
depending on how $\epsvec$ goes to 0
is closely
related to the existence
of {\em multiple extensions\/} in many other theories of default reasoning
(for instance, in default logic \cite{reiter}).
Both non-robustness and the existence of more than one extension suggest
a certain incompleteness of our knowledge.
It is well-known that, in the presence of conflicting defaults, we often
need more information about
the strength of the different defaults in order to resolve the conflict.
Our approach has the advantage of pinpointing the type of information
that would suffice to reach a decision.
Note that our formalism does give us an explicit way to state
in this example that the two extensions are equally likely, by asserting that
the defaults that generate them
have equal strength; namely, we can use $\aeq_1$ to capture both default
statements, rather than using $\aeq_1$ and $\aeq_2$.  In this case, we get
the answer $1/2$, as expected. However, it is not always appropriate to
conclude that defaults have equal strength.
It is not difficult to extend our
language  to allow the user to prioritize defaults, by defining the
relative sizes of the components $\tau_i$ of the tolerance vector.
 
As we mentioned, Theorem~\ref{dempster.rule} tells us only how to
combine statistics from competing reference classes in the very special
case where the intersection of the different reference classes is small.
Shastri \cite[pp.~331--332]{Shastri} describes a result in the same
spirit, but for a different special case: he assumes that,
in addition to the statistics for $P$ within each reference class,
the statistics for $P$ in the general population are also known.
Shastri's result is based on maximum entropy. Maximum entropy is
in fact a very general tool for computing degrees of belief,
provided we restrict to knowledge bases that involve only unary
predicates and are well-behaved in a sense made precise in \cite{GHK2}.
(See the discussion in Section~\ref{GMPsec}.)

\subsection{Independence}
\label{independence}
As we have seen so far, random worlds captures  many of
the natural reasoning heuristics that have been proposed in the
literature.  Another heuristic is a default assumption that
all properties are probabilistically independent unless we know otherwise.
Random-worlds captures this principle as well, in many cases.
It is, in general, very hard to give simple syntactic tests for
when a knowledge base forces two properties to be dependent.
The following theorem concerns one very simple scenario where we can be
sure that no relationship is forced.
 
Consider two disjoint vocabularies $\vocab$ and
$\vocab'$, and two respective knowledge base and query pairs:
$\KB,\phi \in \cL(\vocab)$, and $\KBp,\phi' \in \cL(\vocab')$.  We
can prove that
$$
\priw(\phi \land \phi' | \KB \land \KBp) =
  \priw(\phi|\KB)
                 \times \priw(\phi'|\KBp).
$$
That is, if
there is no connection between the
symbols in the two vocabularies, the two queries will be independent:
the probability of their conjunction is the product of their
probabilities. We now prove a slightly more general case,
where the two queries are both allowed to refer to some constant $c$.
\thm
\label{independent}
Let $\vocab_1$ and $\vocab_2$ be two
subvocabularies of $\vocab$
that are disjoint except
for the constant $c$.  Consider $\KB_1, \phi_1 \in
\cL(\vocab_1)$ and $\KB_2, \phi_2 \in \cL(\vocab_2)$. Then
$$
\priw(\phi_1 \land \phi_2 | \KB_1 \land \KB_2) =
  \priw(\phi_1|\KB_1)
         \times \priw(\phi_2|\KB_2).
$$
\ethm
\prf
See the appendix. %
\eprf

Although very simple, this theorem allows us to deal with such
examples as the following:
\xam
Consider the knowledge base $\KBhep$, and a knowledge base stating
that 40\% of hospital patients are over 60 years old and that Eric
is a patient.
$$
\KB_{>60} \eqdef \cprop{\Older(x)}{\Patient(x)}{x} \aeq_5 0.4 \land
\Patient(\Eric)
$$
Then
\begin{eqnarray*}
\lefteqn{\priw(\Hep(\Eric) \land \Older(\Eric) | \KBhep \land
     \KB_{>60}) = }  \\
 & &  \priw(\Hep(\Eric)|\KBhep)
           \times  \priw(\Older(\Eric)|\KB_{>60})
      = 0.8
            \times 0.4 = 0.32. \ \ \ \bbox
\end{eqnarray*}
\end{example}

In the case of a unary vocabulary (\ie one
containing only unary predicates and constants),
Theorem~\ref{independent} can
be proved using the connection
between the
random-worlds method and maximum entropy, which we discuss in
Section~\ref{GMPsec}.
It is a well-known fact that using maximum entropy often leads to
probabilistic independence.
The result above proves that, with random-worlds, this phenomenon
appears in the non-unary case as well.

We remark that
the connection between maximum entropy and independence is sometimes
overstated.
For example, neither maximum entropy nor random worlds lead to
probabilistic independence in examples like the following:
\xam
Consider the knowledge base $\KB$, describing a domain of animals:
$$
\cprop{\Black(x)}{\Bird(x)}{x} \aeq_1 0.2 \; \land \;
  \prop{\Bird(x)}{x} \aeq_2 0.1.
$$
It is perfectly consistent for $\Bird$ and $\Black$ to be
probabilistically independent.
If this were the case, we would expect the proportion of black
animals to be the same as that of black birds.  In this case, our
degree of belief in $\Black(\Clyde)$, for some arbitrary animal Clyde,
would also be $0.2$.  However, this is not the case.
Since all the predicates here are unary we can use maximum entropy
techniques discussed in Section~\ref{GMPsec} to show that
$\priw(\Black(\Clyde)|\KB) =    0.47$.
That is, we are almost indifferent about Clyde being black, except for
a slight bias due to the fact that he might be a bird and
in that case is
unlikely to be black.
\exam

\subsection{The lottery paradox and unique names}
In Section~\ref{lottery.discuss} we discussed the
{\em lottery paradox\/}
and the challenge it poses to
theories of default reasoning. How does random-worlds perform?
 
{T}o describe the original problem in
our framework, let $\Ticket(x)$ hold if $x$ purchased a lottery ticket.
Consider the
knowledge base consisting of
$$\KB = \exists ! x\, \Winner(x) \land \forall x \, (\Winner(x) \rimp
\Ticket(x)).$$
That is, there is a unique
winner, and in order to win one must purchase a lottery ticket.  If we also
know the size of the lottery, say $N$, we can add to our knowledge base the
assertion $\exists^N x\, \Ticket(x)$ stating that there are precisely $N$
ticket holders.  (This assertion can easily be expressed in first-order
logic using equality.)  We also assume for simplicity that each individual
buys at most one lottery ticket.
Then our degree of belief that
the individual denoted by a particular constant $c$ wins the lottery is
$$\priw(\Winner(c)|\KB \land \exists^N x\, \Ticket(x) \land Ticket(c))
                     = \frac{1}{N}.$$ Our
degree of belief that {\em someone\/} wins will obviously be 1.
We would argue that these are reasonable answers. It is true
that we do not get the default conclusion that $c$ does not win
(\ie degree of belief $0$). But since our probabilistic framework can
and does express the conclusion that $c$ is very unlikely to win, this
is not a serious problem (unlike in systems which either reach a default
conclusion or not, with no possibilities in between).
 
If we do not know the exact number of ticket holders, but have only the
qualitative information that this number is ``large'', then our degree of
belief that $c$ wins the lottery is simply
$\priw(\Winner(c)|\KB \land \Ticket(c)) = 0$,
although, as before, $\priw(\exists x \, \Winner(x) | \KB) = 1$.
In this case we do conclude by default that any particular individual
will not win,
although we still have degree of belief 1 that someone does win.
This shows that the tension Lifschitz sees between concluding
a fact for any particular individual and yet not
concluding the universal does in fact have a solution in a probabilistic
setting such as ours.

Finally, we consider where random worlds fits into Poole's analysis of
the lottery paradox.
Recall, his argument concentrated on examples in which a class (such as
$\Bird(x)$) is known to be equal to the union of a number of
subclasses ($\Penguin(x), \Emu(x), \ldots{}$), each of which is exceptional
in at least one respect.
However, using our statistical interpretation of defaults,
``exceptional'' implies ``makes up a negligible fraction of the
population''.
So under our interpretation, Poole's example is inconsistent:
we cannot partition the set of birds into a finite
number of subclasses, each of which makes up a negligible
fraction of the whole set. We view the inconsistency in this case as
a feature: it alerts the user that this collection of
defaults cannot
all be true of the world (given our interpretation of defaults),
just as would the inconsistency of the default ``Birds typically fly''
with ``Birds typically do not fly''  or ``No bird flies''.

Our treatment of Poole's example clearly depends on our
interpretation of defaults.
For instance, we could interpret the default ``Birds
typically fly'' as $\cprop{\Fly(x)}{\Bird(x)}{x} \ageq \alpha$ for
some appropriately chosen $\alpha$ which is less than 1.
In this case, ``exceptional'' subclasses (such as penguins which are
nonflying birds) {\em can\/} include a nonvanishing fraction of the domain.
While allowing an interpretation of default not based on
``almost all''  does make Poole's $\KB$ consistent, it entails giving up
many of the attractive properties of the $\aeq 1$ representation (such as
having default conclusions assigned a degree of belief 1, and the
properties summarized in Theorem~\ref{KLM.properties}).
An alternative solution would be to use
the approach
presented in \cite{KH92}.
Roughly speaking, this approach interprets
``almost all''  as ``arbitrarily close to 1''
whenever such an interpretation is consistent (and
thus allows us to get the benefits associated with this interpretation).
If this interpretation is inconsistent, it takes ``almost all''
to mean ``within $\eps$ of 1'', for $\eps$ large enough to maintain
consistency.
 
We conclude this section by remarking on another property
of the random-worlds method. Applications of default reasoning
are often simplified by using the {\em unique names\/} assumption,
which says that any two constants should (but perhaps only by default)
denote different objects.
In random worlds, there is a strong {\em automatic\/} bias towards unique
names. If $c_1$ and $c_2$ are not mentioned anywhere in $\KB$, then
$\priw(c_1=c_2|\KB) = 0$
(See~\cite[Lemma~D.1]{GHK2} for a formal proof
of this fact.) Of course, when we know something
about $c_1$ and $c_2$ it is possible to find examples for which
this result fails; for instance
$\priw(c_1=c_2| (c_1=c_2) \lor (c_2=c_3) \lor (c_1=c_3)) = \frac{1}{3}$.
It is hard to give a general theorem saying precisely when the bias
towards unique names overrides other considerations. However, we note
that both of the ``benchmark'' examples that Lifschitz has given
concerning unique names \cite{Lifschitz.bench} are correctly handled
by random-worlds. For instance, Lifschitz's problem C1 is:
\begin{enumerate}
\item Different names normally denote different people.
\item The names ``Ray'' and ``Reiter'' denote the same person.
\item The names ``Drew'' and ``McDermott'' denote the same person.
\end{enumerate}
The desired conclusion here is:
\begin{itemize}
\item The names ``Ray'' and ``Drew'' denote different people.
\end{itemize}
Random worlds gives us this conclusion.
That is, $$\priw(\Ray
\ne \Drew|  \Ray = \Reiter \land \Drew = \McDermott) = 1.$$
Furthermore, we do not have to
explicitly state a unique names default.
 
\section{Random worlds and maximum entropy}
\label{GMPsec}
 
The {\em principle of maximum entropy\/} is a well-known
idea, useful for certain types of probabilistic reasoning.
Briefly,
the {\em entropy\/} of a probability distribution $\mu$ over a
finite space $\Omega$ is defined as
$H(\mu) = -\sum_{\omega \in \Omega} \mu(\omega) \ln(\mu(\omega))$.
It has been argued~\cite{ShannonWeaver} that the entropy
measures the amount of ``information'' in a probability distribution,
in the sense of information theory; note that the uniform distribution has the
maximum possible entropy. The principle of maximum entropy \cite{Jaynes}
addresses situations in which we have some
constraints on a probability distribution, which may have many solutions,
but where we must decide on one particular consistent distribution.
The principle asserts that among those distributions satisfying the
constraints, the one that should be adopted is the (hopefully unique)
distribution having maximum entropy,
because it incorporates
the least additional information
beyond the constraints themselves.
 
No explicit use of maximum entropy is made by random-worlds.%
\footnote{In fact, the postulate of uniform probability over worlds
(\ie indifference) can be seen as a degenerate application
of maximum entropy. However, in the context of this paper, this
is rather uninteresting and is unrelated to the connection we discuss
in the rest of this section.}
Indeed, although they are both tools for reasoning about probabilities,
the classes of problems considered by the two methods are seemingly
disjoint.
Nevertheless, it turns out that there is a surprising and
very close connection between the random-worlds approach and
maximum entropy provided that the language consists only of {\em
unary\/} predicates and constants. In this section we briefly describe
this connection.
This result is of considerable interest simply because it hints
at effective computational techniques for random-worlds in the unary
case.
However, as we discuss below, the connection to random-worlds is
interesting
for other reasons as well. For instance, we use the connection to
show that the maximum-entropy approach to default
reasoning, considered in \cite{GMP}, can be embedded in our framework.
 
To understand the connection to maximum entropy, suppose the
language consists of the unary predicate symbols $P_1, \ldots, P_k$
together with some constant symbols.  (Thus, we do not allow either
function symbols or higher-arity predicates.)
We can consider the $2^k$ {\em atoms\/} that can be formed from these
predicate symbols, namely, the formulas of the form $Q_1 \land \ldots
\land Q_k$, where each $Q_i$ is either $P_i$ or $\neg P_i$. Then the
knowledge base $\KB$ can be viewed as simply placing constraints on
the proportion of domain elements satisfying each atom. For example,
the formula $\cprop{P_1(x)}{P_2(x)}{x} = 1/2$ says that the proportion
of the domain satisfying some atom containing $P_1$ as a conjunct is
twice the proportion satisfying atoms containing both $P_1$ and $P_2$
as conjuncts. For unary languages (only) it can be shown that every
formula can be rewritten in a canonical form from which constraints
on the possible proportions of atoms can be simply derived.
Details of this and all other specific results can be found in \cite{GHK2},
although the general phenomenon we are about to discuss is addressed in many
places, such as \cite{PV,Shastri}
and in
statistical physics
(\eg \cite{Landau}).

The set of constraints generated by $\KB$ defines a subset of
$[0,1]^{2^k}$, which we call $S(\KB)$.
That is, each vector in $S(\KB)$, say
$\pvec = \Vector{p_1,\ldots,p_{2^k}}$, is a solution of the constraints
defined by $\KB$ (where $p_i$ is the proportion of atom $i$).
For
example, suppose our language contains only the two predicate symbols
$\{P_1, P_2\}$, so that there are four atoms $A_1 = P_1 \land P_2$,
$A_2 = P_1 \land \neg P_2$, $A_3 = \neg P_1 \land P_2$, and $A_4 =
\neg P_1 \land \neg P_2$.  Let $\KB = \forall x\, P_1(x) \land
\prop{P_1(x) \land P_2(x)}{x} \aleq_1 0.3$.
The first conjunct of $\KB$ clearly constrains both $p_3$ and $p_4$
(the proportion of domain elements satisfying atoms $A_3$ and $A_4$)
to be 0. The second conjunct forces $p_1$ to be (approximately) at
most $1/3$. Thus, $S(\KB) = \{\Vector{p_1, \ldots, p_4} \in [0,1]^4:
p_1 \le 0.3, p_3 = p_4 = 0, p_1 + p_2 = 1\}$.

The connection between maximum entropy and the random-worlds method is
based on the following observations.  With every world $\world$, we
can associate the vector $\pvec^\world$, where $p^\world_i$ is
$\prop{A_i(x)}{x}$ in $\world$.
Each vector $\pvec$ can be viewed as a probability distribution over
the space of atoms $A_1,\ldots,A_{2^k}$; we can therefore associate an
entropy with each such vector.
We define the {\em entropy\/} of
$\world$ to be the entropy of $\pvec^\world$.  Now,
consider some point $\pvec \in S(\KB)$.  What is the number of worlds
$\world \in \WN$ such that $\pvec^\world = \pvec$?  Clearly, for those
$\pvec$ where some $p_i$ is not an integer multiple of $1/N$, the
answer is 0.  However, for those $\pvec$ which are ``possible'', this
number grows asymptotically as $e^{N H(\pvec)}$.  Hence, there are
vastly more worlds $\world$ for which $\pvec^\world$ is ``near'' the
maximum entropy point of $S(\KB)$ than there are worlds elsewhere.
This allows us to prove the following result:  If, for all
sufficiently small $\epsvec$, a formula $\theta$
is true in all worlds around the maximum entropy
point(s) of $S(\KB)$, then
$\Prinf(\theta | \KB) = 1$.
 
In the above example, the maximum-entropy point in $S(\KB)$ is
$\pvec^* = (0.3,0.7,0,0)$. Our knowledge base only tells us that the size
of atom $A_1$ is (approximately) less than or equal to 1/3.  But now,
consider some small fixed $\epsilon$ and the formula $\theta[\epsilon]
= \prop{P_2(x)}{x} \in [0.3-\epsilon,0.3+\epsilon]$.  Since this
formula certainly holds at all worlds
$\world$ where $\pvec^\world$ is
sufficiently close to $\pvec^*$,
we conclude that $\Prinf(\theta[\epsilon] | \KB) = 1$.  This allows us
to use Proposition~\ref{cmc} to conclude that, for any formula $\phi$,
$\Prinf(\phi|\KB) = \Prinf(\phi|\KB \land \theta[\epsilon])$.  In
particular, this holds for $\phi = P_2(c)$.  But now, we can use direct
inference to conclude that $\Prinf(P_2(c)|\KB) \in
[0.3-\epsilon,0.3+\epsilon]$.  Since this holds for all sufficiently
small $\epsilon$, we conclude that $\Prinf(P_2(c)|\KB) = 0.3$, as
desired.  In \cite{GHK2} we formalize this argument and generalize it
to more complex examples.  These techniques allow us to use a
maximum entropy computation as a basis for computing degrees of
belief.  The resulting procedure applies to many cases not
covered by our results in Section~\ref{results}.  Furthermore, since
we can take advantage of existing algorithms for computing
maximum entropy
(see \cite{Goldman87} and the references therein),
we obtain a technique of potential practical significance.
 
The connection to maximum entropy is important for many reasons,
aside from its computational implications.
Maximum entropy has been a popular technique for
probabilistic reasoning in AI and elsewhere.
Two highly relevant works
are the application to inheritance
hierarchies by Shastri \cite{Shastri} and to default reasoning by
Goldszmidt, Morris, and Pearl \cite{GMP}.
It is desirable that such a popular technique be well-understood and
motivated, rather than be seen as an {\em ad hoc\/}
heuristic. Random worlds, resting
on the basic principle of indifference,
provides motivation
which some may find more convincing than the usual information-theoretic
justifications.

Not only does the random-worlds method provide motivation for maximum
entropy, it can be viewed as a generalization of it.
As discussed above, there is a strong connection between the
random-worlds approach and maximum entropy in the unary case (see also
\cite{GHK2}).  In fact,  restricted versions of
some of our results
{from} Section~\ref{results} can be proved using maximum
entropy (see \cite{Shastri}). But our combinatorial
proof techniques are more general (and, in fact, simpler) than
the ones based directly on entropy. The limitations of maximum entropy
are perhaps inescapable, because (as we discuss in detail in \cite{GHK2})
it is reasonable to conjecture that
maximum entropy is inherently inapplicable once we move beyond unary
predicates.
 
Finally, our results connecting random worlds to maximum entropy can
also be put to use to help clarify the connection between random
worlds and previous approaches to applying probabilistic semantics to
default reasoning. The mainstay of most of this previous work has been
the formalism of $\epsilon$-semantics
\cite{Geffner:Framework.Reasoning.With.Defaults}.
We briefly review $\epsilon$-semantics here.
 
Consider a language consisting of propositional formulas (over some
finite set of propositional variables $P_1, \ldots, P_k$) and default
rules of the form $B \rightarrow C$ (read ``$B$'s are typically
$C$'s''), where $B$ and $C$ are propositional formulas.  Let $\Omega$
be the set of $2^k$ {\em propositional worlds}, corresponding to the
possible truth assignments to
the variables. Given a probability
distribution $\mu$ on $\Omega$, we define $\mu(B)$ to be the
probability of the set of worlds where $B$ is true. We say that a
distribution $\mu$ {\em $\epsilon$-satisfies\/} a default rule $B
\rightarrow C$ if $\mu(C|B) \geq 1-\epsilon$.
 
A {\em parameterized probability distribution\/} (PPD) is a collection
$\{\mu_\epsilon\}_{\epsilon>0}$ of probability distributions over
$\Omega$, parameterized by $\epsilon$. A PPD
$\{\mu_\epsilon\}_{\epsilon>0}$ $\epsilon$-satisfies a set $\R$ of
default rules if for every $\epsilon$, $\mu_\epsilon$
$\epsilon$-satisfies every rule $r \in \R$. A set $\R$ of default
rules $\epsilon$-entails $B \rarrow C$ if for every PPD that
$\epsilon$-satisfies $\R$, $\lim_{\epsilon \tends 0}
\mu_\epsilon(C|B) = 1$.
 
As shown in
\cite{Geffner:Framework.Reasoning.With.Defaults}, $\epsilon$-entailment
possesses a number of reasonable properties typically associated with
default reasoning, including a preference for more specific
information. However, $\epsilon$-entailment is very weak. In
particular, as shown by Adams \cite{Adams}, the consequence relation
defined by $\epsilon$-entailment satisfies only the five basic
properties of default inference given in Section~\ref{KLM}. Hence,
among other limitations, it has no ability to ignore irrelevant
information, so it cannot perform any inheritance reasoning.

In order to obtain additional desirable properties, it is necessary to
restrict the class of admissible PPD's.  Goldszmidt, Morris, and
Pearl~\cite{GMP} focus attention on a single
PPD: the {\em maximum entropy PPD\/}
$\{\mu^*_{\epsilon,\R}\}_{\epsilon>0}$
(See \cite{GMP} for precise definitions and technical
details.)  A rule $B \rightarrow C$ is defined to be an {\em ME-plausible
consequence} of $\R$ if $\lim_{\epsilon \tends 0}
\mu^*_{\epsilon,\R}(C|B) = 1$. The notion of ME-plausible consequence
is analyzed in detail in \cite{GMP}, where it is shown to inherit all
the nice properties of $\epsilon$-entailment while having some
ability to
ignoring irrelevant information. Equally importantly, algorithms are
provided for computing the ME-plausible consequences of a set of rules
in certain cases (see also \cite{GMPfull}).
 
Our results relating random worlds to maximum entropy can be used
to show that the approach of \cite{GMP} can be embedded in our
framework in a straightforward manner. We simply convert all default
rules $r$ of the form $B \rightarrow C$ into formulas of the form
$\theta_r
\eqdef \cprop{\psi_C(x)}{\psi_B(x)}{x} \aeq_1 1$, where $\psi_B$ is the
formula obtained by replacing each occurrence of the propositional
variable $p_i$ in $B$ with $P_i(x)$. Note that the formulas that arise
under this conversion all use the same approximate equality relation
$\aeq_1$, since the approach of \cite{GMP} uses the same $\epsilon$
for all default rules. Note also that propositional variables become
unary predicates. Hence, default rules become statistical assertions
about classes of individuals.
Under this translation, we obtain the following theorem (which is proved,
and discussed in more detail, in \cite{GHK2}).
\thm
\label{thm.gmp}
Let $c$ be a constant symbol.  Using the translation described above,
for any set $\R$ of defeasible rules, $B \rightarrow C$ is an
ME-plausible consequence of $\R$ iff
$$\priw(\psi_C(c) | \band_{r \in \R} \theta_r \land \psi_B(c)) = 1.$$
\ethm
Hence, all the computational techniques and results described in
\cite{GMP} carry over to this special case of our approach.
Furthermore, unary versions of all of our theorems carry over to the
ME-plausible consequence relation. Examples demonstrating inheritance
were given in \cite{GMP}, but we can now use Theorem~\ref{inherit.thm}
to provide a formal characterization of some of the inheritance
properties of this consequence relation. It should also be noted that
our translation converts the default rules into statistical assertions
about classes of individuals and it converts the {\em context\/},
i.e., $B$, into information about a particular individual (whose name
we have arbitrarily chosen to be $c$). This is in keeping with the
intuitive interpretations of rules and context used by propositional
default systems (see Section~\ref{nonmon.expressivity}).

We stress that the assumption that we use the same approximate
equality relation is crucial in Theorem~\ref{thm.gmp}.
Geffner~\cite{Geffner:Conditional.Entailment} gives an example of an
anomalous conclusion obtained in the system of \cite{GMP}. Suppose the
rule set $\R$ consists of the two rules $P\land S \rightarrow Q$ and
$R \rightarrow \lnot Q$. In this case, the rule $P\land S\land R
\rightarrow Q$ is not an ME-plausible consequence of $\R$. This seems
reasonable, as we have evidence for $Q$ ($P\land S$) and against $Q$ ($R$),
and neither piece of evidence is more specific than the other.
However, if we add the new rule $P\rightarrow \lnot Q$ to $\R$, then
$P\land S\land R\rightarrow Q$ does become an ME-plausible consequence
of $\R$. This behavior seems counterintuitive, and is a consequence
of~\cite{GMP}'s use of the same $\epsilon$ for all of the rules.
Intuitively, what is occurring here is that prior to the addition of the
rule $P\rightarrow \lnot Q$, the sets $P(x)\land S(x)$ and $R(x)$ were
of comparable size. The new rule forces $P(x)\land S(x)$ to be
an $\epsilon$-small subset of $P(x)$, since almost all $P$'s are
$\lnot Q$'s, whereas almost all $P\land S$'s are $Q$'s. The size of the
set $R(x)$, on the other hand, is unaffected. Hence, the default for
the $\epsilon$-smaller class $P\land S$ now takes precedence over the
class $R$.
Once we allow a family of approximate equality connectives,
each one corresponding to a different $\epsilon$,
we are no longer forced to derive this conclusion.
An appropriate choice of $\eps_i$ can make
the default $\cprop{\lnot Q(x)}{R(x)}{x} \aeq_i 1$ so strong that
the number of $Q$'s in the set $R(x)$, and hence the number of $Q$'s in
the subset $P(x)\land S(x)\land R(x)$, is much smaller than the size
of the set $P(x)\land S(x) \land R(x)$.
In this case, the rule $R\rightarrow \lnot Q$ takes
precedence over the rule $P\land S\rightarrow Q$. With no specific
information about the relative strengths of the defaults we get
non-robustness, as in the Nixon Diamond. That is, we draw no conclusions
about $P\land S\land R \rightarrow Q$.

\section{Problems}
\label{discussion}

The principle of indifference and maximum entropy have both
been subject to criticism.
Any such criticism is, at least potentially, relevant to
random worlds.  Hence, it is important that we examine the
difficulties that people have found.
In this section, we consider problems relating to {\em causal reasoning},
{\em language dependence}, {\em acceptance}, {\em learning}, and
{\em computation}.
 
\subsection{Causal and temporal information}\label{causal}
The random-worlds method can use
knowledge bases which include statistical,
first-order, and default information.
When is this language sufficient?
We suspect that it is, in fact, adequate for most traditional
knowledge representation tasks. Nevertheless,
the question of adequacy can be subtle.
This is certainly the case for the important domain of
reasoning about actions, using causal and temporal information.  In
principle, there would seem to be no difficulty
choosing a suitable first-order vocabulary that includes the ability
to talk about time explicitly.
In the semantics appropriate to many such languages, a
world might model an entire temporal sequence of events.
However, finding a representation with sufficient expressivity is
only part of the problem: we need to know whether the degrees of
belief we derive will correctly
reflect our intuitions about causal reasoning.
It turns out that random worlds gives unintuitive results
when used with the most straightforward representations
of temporal knowledge.

This observation is not really a new one.
As we have observed, the random-worlds method is closely
related to maximum entropy (in the context of a unary knowledge base).
One significant criticism of maximum entropy techniques has been
that they seem to have difficulty dealing with causal information
\cite{Hunter89,Pearl}. Hence, it is not surprising that
the random-worlds method also gives peculiar answers if we represent
causal and temporal information naively.
On the other hand, Hunter \cite{Hunter89}
has shown that maximum entropy methods can deal with causal
information, provided it is represented appropriately.
We have recently shown that by using an appropriate representation
(related to Hunter's, but nevertheless distinct), the random-worlds method
can also deal well with causal information \cite{BGHKcausal}.  Indeed,
our representation allows us to
(a) deal with prediction and explanation problems,
(b) represent causal information of
the type implicit in Bayesian causal networks \cite{Pearl}, and (c)
provide a clean and concise solution to the frame problem in the
situation calculus \cite{MH}.  In particular, our proposal
deals well with some of the standard problems in the area, such as
the Yale Shooting Problem \cite{HM:YSP}.

The details of the proposal are beyond the scope of this paper.
However, the fact we want to emphasize here is that there may be more
than one reasonable way to represent our knowledge of a given domain.
When one formulation does not work as we expect, we can look
for other ways of representing the problem.  It will often turn out
that the new representation captures some subtle aspects of the
domain, that were ignored by the naive representation.  (We believe
that this is the case with our alternative formulation of reasoning
about actions.) We return to this issue a number of times below.

\subsection{Representation dependence}
 
As we saw above, random worlds suffers from a problem of
representation dependence: causal information is treated
correctly only if it is represented appropriately.  This shows that
choosing the ``right'' representation of our knowledge is important
in the context of the random-worlds approach.
 
In some ways, this representation dependence is a serious problem because,
in practice, how can we know whether we have chosen a good
representation or not?
Before addressing this, we note that
the situation with random worlds is actually not as bad as it might be.
As we pointed out in Section~\ref{KLMresults}, the random-worlds
approach is not sensitive to merely syntactic changes in the knowledge base:
logically equivalent knowledge bases always result in the same degrees of
belief. So if a changed representation gives different answers,
it can only be because we have changed the semantics: we might be
using a different ontology, or the new representation might model the
world with a different level of detail and accuracy.
The representation dependence exhibited by random worlds concerns
more than mere syntax.  This gives us some hope that the phenomenon
can be understood and, at least in some cases, be seen to be entirely
appropriate.

Unfortunately, it does seem as if random worlds really is too sensitive;
minor and seemingly irrelevant changes can affect things.
Perhaps the most disturbing examples concern
{\em language dependence}, or sensitivity to
definitional changes.  For instance, suppose the only
predicate in our language is $\White$ and we take $\KB$ to be
$\true$.  Then $\priw(\White(c)|\KB) = 1/2$.
On the other hand, if we refine $\neg \White$ by adding $\Red$ and
$\Blue$ to our language and having $\KB'$ assert that
$\neg \White$ is their disjoint union, then $\priw(\White(c)|\KBp) = 1/3$.
The fact that simply expanding
the language and giving a definition of an old notion ($\neg \White$) in
terms of the new notions ($\Red$ and $\Blue$) can affect
the degree of belief seems to be a serious problem.  There are
several approaches to dealing with this issue.
One approach to dealing with representation independence is to search for a
method of computing degrees of belief that does not suffer from it. To do
this, it is important to have a formal definition of representation
independence. Once we have such a definition, we can investigate whether
there are nontrivial approaches to generating degrees of belief that are
representation independent.  For example, one might think that the problem
lies with our use of point-valued degrees of belief.  After all, it is
fairly obvious that (under a few very weak assumptions) any approach that
gives point-valued degrees of belief that act like probabilities
cannot be  representation independent.  In fact, there has been considerable
debate about the ``excess precision'' forced by point-valued probabilities.
Perhaps if we generalize our concept of
``degree of belief'', say to intervals rather than point values in $[0,1]$,
we can address these concerns, as well as avoid the problem of
representation dependence.
Unfortunately, while interval-valued degrees of belief might well be
representation independent in many more circumstances than random worlds,
they do not solve the problem.
Halpern and Koller, in~\cite{HK95},
provide a
definition of representation independence in the context of probabilistic
reasoning, and show that essentially any interesting non-deductive inference
procedure cannot be representation independent in their sense.

Another response is to accept this, but to
declare that representation dependence is justified,
\ie that the choice of an appropriate vocabulary is indeed a significant
one, which does encode some of the information at our disposal.
In our example above, we can view the choice of vocabulary as
reflecting the bias of the reasoner with respect to the partition of
the world into colors. Researchers in machine learning and the
philosophy of induction have long realized that bias is an inevitable
component of effective inductive reasoning.  So we should not
be completely surprised if it turns out that the related
problem of finding
degrees of belief should also depend on the bias.
Of course, if this is the case we would hope to
have a good intuitive understanding of how the degrees of belief
depend on the bias.  In particular, we would like to give the
knowledge base designer some guidelines to selecting the
``appropriate'' representation.  This is an important and seemingly
difficult problem in the context of random worlds.
 
A third response to the problem
is to prove representation
independence with respect to a large class of queries (see also
\cite{HK95}).
To understand
this approach, consider another example. Suppose that we know that
only about half of birds can fly, Tweety is a bird, and Opus is some
other individual (who may or may not be a bird).
One obvious way to represent this information is
to have a language with predicates $\Bird$ and $\Fly$, and take the
$\KB$ to consist of the statements $\cprop{\Fly(x)}{\Bird(x)}{x}
\aeq 0.5$ and $\Bird(\Tweety)$.  It is easy to see that
$\priw(\Fly(\Tweety)|\KB) = 0.5$ and $\priw(\Bird(\Opus)|\KB) = 0.5$.
But suppose that we had chosen to use a different language,
one
that uses
the basic predicates $\Bird$ and $\FlyingBird$.  We would then
take $\KBp$ to consist of the statements
$\cprop{\FlyingBird(x)}{\Bird(x)}{x} \aeq 0.5$, $\Bird(\Tweety)$,
and $\forall x( \FlyingBird(x) \rimp \Bird(x))$.
We now get $\priw(\FlyingBird(\Tweety)|\KBp) = 0.5$ and
$\priw(\Bird(\Opus)|\KBp) = 2/3$.
Note that our degree of belief that Tweety flies is $0.5$ in both
cases.
In fact, one can give an argument that this conclusion is robust against
many ``reasonable'' representation changes.
On the other hand, our degree of belief that Opus is a bird
differs in the two representations.
Arguably, the fact that our degree of belief that Opus is a bird
is language dependent is a direct reflection of the fact that
our knowledge base does not contain sufficient
information to assign it a single ``justified'' value.
This suggests that it would be useful to characterize those queries
that are language independent,
while recognizing that not all queries will be.
 
\subsection{Acceptance and learning}
\label{learning}
The most fundamental assumption in this paper is that we are given a
knowledge base $\KB$ and wish to calculate degrees of belief relative
this knowledge.
We have not considered how one comes to know
$\KB$ in the first place.
That is, when do we {\em accept\/} information as knowledge?
We do not have a good answer to this question.  This
is unfortunate, since it seems plausible that the
processes of gaining knowledge and computing degrees of belief should
be interrelated. In particular,
Kyburg \cite{Kyburg:Full.Beliefs} has argued that perhaps we might accept
assertions that are believed sufficiently strongly.
For example, suppose we observe a block $b$ that appears to be white.
It could be that  we are is not entirely sure that the block is indeed
white; it might be some other light color.  Nevertheless, if
our confidence in $\White(b)$ exceeds some threshold, we might
accept it (and so include it in $\KB$).
 
The problem of acceptance in such examples, concerned with
what we learn directly from the senses, is well-known in philosophy
\cite{Jeffrey68}.
But the problem of acceptance we face is even more
difficult than usual, because of our statistical language.
Under what circumstances is a statement such as
$\cprop{\Fly(x)}{\Bird(x)}{x} \aeq 0.9$ accepted as knowledge?
Although we regard this as an objective statement about the
world, it is unrealistic to suppose that anyone could examine
all the birds in the world and count how many of them fly.
In practice, it seems that this statistical statement would appear in  $\KB$
if someone inspects a (presumably large) sample of birds and
about 90\% of the birds in this sample fly.  Then a leap is
made: the sample is assumed to be typical, and we then conclude that
90\% of all birds fly. This would be in the spirit of Kyburg's suggestion
so long as we believe that, with high confidence, the full
population has statistics similar to those of the sample.
 
Unfortunately, the random-worlds method by itself does not support
this leap, at least not if we represent the sampling in the most
obvious way.  That is, suppose we represent our sample using a
predicate $S$.  We could then represent the fact that 90\% of a
sample of birds fly as
$\cprop{\Fly(x)}{\Bird(x) \land S(x)}{x} \aeq 0.9$.  If the $\KB$
consists of this fact and $\Bird(\Tweety)$, we might hope that
$\priw(\Fly(\Tweety)|\KB) = 0.9$, but it is not.
In fact, random worlds treats the birds in $S$ and those outside $S$
as two unrelated populations; it maintains the default degree of belief
(1/2) that a bird not in $S$ will fly.%
\footnote{%
A related observation, that random worlds cannot do learning (although
in a somewhat different sense), was made by Carnap \cite{Carnap2},
who apparently lost his enthusiasm for (his version of) random worlds
for precisely this reason.}
Intuitively, random worlds is not treating $S$ as a
{\em random\/} sample.
 
Of course, the failure of the obvious approach does not
imply that random worlds is incapable of learning statistics. As was the
case for causal reasoning, the solution may be to find an appropriate
representation.
Perhaps we need a representation reflecting the fact
that different individuals do not acquire their properties completely
independently of each other.
If we see that an animal is tall, it may tell us something about
its genetic structure and so, by this mechanism, hint at properties
of other animals. But clearly this issue is subtle. If we see a giraffe,
this tells us much less about the height of animals in general than it
does about other giraffes, and a good representation should reflect this.

While we still hope to find ways of doing sampling within random worlds,
we can also look for other ways of coping with the problem
of learning. One idea is to
add statements about degrees of
belief to the knowledge base.
Thus, if 20\% of animals in a sample are tall,
and we believe that it is appropriate to learn this statistic,
then we might add a statement such as
$\Pr(\cprop{\Tall(x)}{\Animal(x)}{x} \aeq_1 0.2) \geq 0.9$ to the
$\KB$. Although this does not ``automate'' the sampling procedure, it
allows us to use our belief that a sample is likely to be
representative, without committing absolutely to this fact.  In
particular, this representation allows further evidence to convince
the agent that a sample is, in fact, biased.  Adding degrees of belief
would also let us deal with the problem of acceptance, mentioned at
the beginning of this subsection.  If we believe that block $b$ is
white, but are not certain, we could write $\Pr(\White(b)) \geq 0.9$.
We then do not have to fix an
arbitrary
threshold for acceptance.
 
Adding degree of belief statements to a knowledge base is a
nontrivial step.  Up to now, all the assertions we allowed in a
knowledge base were either true or false in a given world.  This is
not the case for a degree of belief statement.  Indeed, our semantics
for degrees of belief involve looking at sets of possible worlds.
Thus, in order to handle such a statement appropriately, we would need
to ensure that our probability distribution over the possible worlds
satisfies the associated constraint.
A number of
different
approaches to doing this are discussed in \cite{BGHKadddob},
and shown to be essentially equivalent.

Yet
another approach for dealing with the learning problem is to use
a variant of random worlds presented in \cite{BGHK}
called the {\em random-propensities\/} approach.
Random worlds has a strong bias
towards believing that exactly half the domain has any given
property, and this is not always reasonable. Why
should it be more likely that half of all birds fly than that a third
of them do?  Roughly speaking, the random-propensities approach
postulates the existence of a parameter denoting the ``propensity''
of a bird to fly.
Initially, all propensities are equally likely.
Observing a flying bird gives us information about the propensity of
birds to fly, and hence about the flying ability of other birds.
As shown in \cite{BGHK}, the random propensities method does,
indeed, learn from samples.
We can also show \cite{KH.irrel} that the random propensities approach
has many of the same attractive properties that we have shown for random
worlds, in particular direct inference, specificity, and irrelevance.
Unfortunately, random propensities has its own problems.  In particular,
it learns ``too often'', \ie even from arbitrary subsets that are not
representative samples. Given the assertion ``All giraffes are
tall'', random propensities would conclude that almost everything is
tall.
Addressing this problem appropriately  is an important issue
that deserves further investigation.
 
\subsection{Computational issues}
\label{discussion.complexity}
 
Our goal in this research has been to understand some of the
fundamental issues involved in first-order probabilistic and default
reasoning. Until such issues are understood, it is perhaps reasonable
to ignore or downplay concerns about computation. If an ideal
normative theory turns out to be impractical for computational
reasons, we can still use it as guidance in a search for
approximations and heuristics.
 
As we show in \cite{GHK1a}, computing degrees of belief according to
random worlds is, indeed, intractable in general.  This is not
surprising: our language extends first-order logic, for which validity
is undecidable.%
\footnote{Although, in fact, finding degrees of belief using random
  worlds is even {\em more\/} intractable than the problem of deciding
  validity in first-order logic.}
Although unfortunate, we do not view this as an insurmountable
problem.  Note that, in spite of its undecideability, first-order
logic is nevertheless viewed as a powerful and useful tool.  We
believe that the situation with random worlds is analogous.  Random
worlds is not just a computational tool; it is inherently interesting
because of what it can tell us about probabilistic reasoning.
 
But even in terms of computation, the situation with random worlds is
not as bleak as it might seem.  We have presented one class of much
more tractable knowledge bases: those using only unary predicates and
constants.  We showed in \cite{GHK2} and in Section~\ref{GMPsec} that,
in this case, we can often use maximum entropy as a computational tool
in deriving degrees of belief. While computing maximum entropy is also
hard in general, there are heuristic techniques that work efficiently
in practical cases.
As we have already claimed, this class of problems is an important
one. In general, many properties of interest can be expressed using
unary predicates, since they express properties of individuals.  For
example, in physics applications we are interested in such predicates
as quantum state (see \cite{Denbigh}). Similarly, AI applications and
expert systems typically use only unary predicates
(\cite{Cheeseman83}) such as symptoms and diseases.  In fact, a good
case can be made that statisticians tend to reformulate all problems
in terms of unary predicates, since an event in a sample space can be
identified with a unary predicate \cite{ShaferPC}.  Indeed, most cases
where statistics are used, we have a basic unit in mind (an
individual, a family, a household, etc.), and the properties
(predicates) we consider are typically relative to a single unit (\ie
unary predicates). Thus, results concerning computing degrees of
belief for unary knowledge bases are quite significant in practice.
 
Even for non-unary knowledge bases, there is hope. The intractability
proofs given in \cite{GHK1a} use knowledge bases that force the
possible worlds to mimic a Turing machine computation.
Typical knowledge bases do not usually encode Turing machines!  There
may therefore be many cases in which computation is practical.  In
particular, specific domains typically impose additional structure,
which may simplify computation. This seems quite possibly to be the
case, in particular, in certain problems that involve reasoning about
action.
 
Furthermore, as we have seen, we {\em can\/} compute degrees of belief
in many interesting cases.  In particular, we have presented a number
of theorems that tell us what the degrees of belief are for certain
important classes of knowledge bases and queries.  Most of these
theorems hold for our language in its full generality, including
non-unary predicates.  We believe that many more such results could be
found.  Particularly interesting would be more ``irrelevance'' results
that tell us when large parts of the knowledge base can be ignored.
Such results could then be used to reduce apparently complex problems
to simpler forms, to which other techniques apply.
We have already seen
in some of the examples that combining different results can often let
us compute degrees of belief in cases where no single result suffices.

Nevertheless, there are many natural knowledge bases that fail to meet
the syntactic restrictions required by the theorems we have provided.
In particular, the default-reasoning literature includes many quite
complicated examples, and we have often found that we cannot
understand random-worlds' behavior on these without
some (often nontrivial) special-purpose arguments.  (Interestingly, it
seems to us that the complexity of the required arguments is
correlated with how controversial the
example is!)
Of course, these difficulties suggest a research strategy: that of
characterizing the behavior of the random-worlds method on ever-larger
classes of examples. We close by hinting at one of the interesting
technical challenges that
confronts such a research agenda. It turns out (perhaps not that
surprisingly) that a major obstacle is simply the richness of our
language. Consider Theorem~\ref{inherit.thm}. Recall that we had to
impose rather severe syntactic restrictions, whose purpose was to
ensure that we could identify all the subformulas {\em relevant\/} to
a property $\phi(x)$. The conditions were made so strong because in
general it is easy, in a language with as much expressive power as
ours, to construct examples in which one part of the knowledge base
places nontrivial logical or probabilistic constraints on a
superficially (\ie syntactically) unrelated concept.
We certainly believe that random worlds can do inheritance and
irrelevance reasoning
in a much more comprehensive fashion
than suggested by this particular result. But it appears to be
hard
to state clean, checkable, conditions in a way that does not admit
contrived counterexamples.  Some progress in this regard is made in
\cite{KH.irrel},
where additional tools are provided for testing
whether formulas can be treated as being irrelevant.
In fact, the results of \cite{KH.irrel} can be viewed as steps towards
characterizing the properties of a prior distribution that lead
   to results such as Theorems~\ref{direct} and~\ref{inherit.thm}, and
show that such results apply to priors other than the
   uniform prior used in the random-worlds method.

\section{Summary}
 
\label{conclusions}
The random-worlds approach for probabilistic reasoning
is derived from two very intuitive ideas: possible worlds and the
principle of indifference. In spite of its simple semantics,
it has many attractive features:
\begin{itemize}
\item It can deal with very rich knowledge bases that involve
quantitative information in the form of statistics, qualitative
information in the form of defaults, and first-order information.
The language is sufficiently powerful for even fairly esoteric demands
such as the representation of nested defaults.

\item
It uses a simple and well-motivated statistical
interpretation for defaults.  The corresponding semantics allow us to
examine the reasonableness of a default with respect to our entire knowledge
base, including other default rules.
 
\item It
validates many desirable properties, like a preference for
more specific information, the ability to ignore irrelevant information,
a default assumption of unique names, the ability to combine different
pieces of evidence, and more.
Most importantly, these properties arise naturally from the very simple
semantics of random worlds. In particular, {\em ad hoc\/}
assumptions, designed to realize these properties, play no part in the
definition of the method.
 
\item It avoids many of the problems that have plagued
systems of reference-class reasoning (such as the disjunctive reference
class problem) and many of the problems that have plagued
systems of non-monotonic reasoning (such as exceptional-subclass inheritance
and the lottery paradox). Many systems have been forced to work hard to
avoid problems which, in fact, never even arise for random worlds.
\item The random-worlds approach subsumes several important reasoning
systems, and generalizes them to the case of first-order logic.  In
particular, it encompasses deductive reasoning, probabilistic reasoning,
certain theories of nonmonotonic inference, the principle of maximum
entropy, some rules of evidence combination, and more. But it is far more
powerful than any
of these individual systems.
 
\end{itemize}
 
As we saw in Section~\ref{discussion}, there are certainly some
problems with the random-worlds method.
We believe that these problems are far from insuperable.
But, even conceding these problems for the moment, the
substantial success of random-worlds supports a few general conclusions.
 
One conclusion concerns the role of statistics and
degrees of belief. The difference between these, and the problem of
relating the two, is at the heart of our work.
People have long realized that degrees of belief provide a powerful
model for understanding rational behavior (for instance, through
decision theory). The random-worlds approach shows that it is possible
to assign degrees of belief, using a principled technique, in
almost any circumstance. The ideal situation, in which we have complete
statistical knowledge concerning a domain,
is, of course, dealt with
appropriately by random worlds.  But more realistically, even
partial statistical information (which need not be precise)
can still be utilized by random worlds to give useful answers.
Likewise, completely non-numeric data, which may include defaults
and/or a rich first-order theory of some application domain, can be
used.  Probabilistic reasoning need not make unrealistic demands of
the user's knowledge base. Indeed, in a sense it makes less demands
that any other reasoning paradigm we know of.
 
This leads to our next, more general conclusion, which is that
many seemingly disparate forms of representation and reasoning
{\em can\/} (and, we believe, {\em should\/}) be unified.
The first two points listed above suggest that we can take a large step
towards this goal by simply finding a powerful language (with
clear semantics) that subsumes specialized representations.
The advantages we have found (such as a clear and general way of
using nested defaults, or combining defaults and statistics) apply
even if one rejects the random-worlds reasoning method itself.
But the language is only part of the answer. Can diverse types of
reasoning really be seen as aspects of a single more general system?
Clearly this is not always possible; for instance, there are surely some
interpretations of ``defaults'' which have no interesting connection
to statistics whatsoever.  However, we think that our work demonstrates that
the alleged gap between probabilistic reasoning and default reasoning is
much narrower than is often thought.  In fact, the success of random
worlds encourages us to hope that a synthesis between different knowledge
representation paradigms is possible in most of the interesting domains.

\appendix
\section{Proofs of results}
\label{app.results}
 
\othm{rational.mon}
Assume that $\KB \rwent \phi$ and $\KB \notrwent \neg\theta$.  Then
$\KB \land \theta \rwent \phi$ provided that $\priw(\phi|\KB \land
\theta)$ exists.  Moreover, a sufficient condition for $\priw(\phi|
\KB \land \theta)$ to exist is that $\priw(\theta|\KB)$ exists.
\eothm
\prf
Since $\KB \notrwent \neg\theta$,
$\priw(\neg\theta | \KB) \neq 1$, so that
$\priw(\theta|\KB) \ne  0$.  Therefore, there exists some $\epsilon > 0$
for which we can construct a sequence of pairs $N^i,{\epsvec\,}^i$ as follows:
$N^i$ is an increasing sequence of domain sizes, ${\epsvec\,}^i$ is a
decreasing sequence of tolerance vectors, and $\prNwi(\theta|\KB) >
\epsilon$.  For these pairs $N^i,{\epsvec\,}^i$ we can conclude
using standard probabilistic reasoning that
$$
\prNwi(\neg \phi|\KB\land\theta) = \frac{\prNwi(\neg\phi\land\theta|\KB)}
  {\prNwi(\theta|\KB)} \leq \frac{\prNwi(\neg\phi|\KB)}{\prNwi(\theta|\KB)}.
$$
Since $\priw(\neg\phi|\KB) = 0$, it
follows that
$\lim_{i\tends\infty} \prNwi(\neg\phi|\KB) = 0$.  Moreover, we know that
for all $i$, $\prNwi(\theta|\KB) > \epsilon > 0$.
We can therefore take the limit as
$i \tends \infty$, and conclude that $
\lim_{i \tends \infty}
\prNwi(\neg\phi|\KB\land\theta) = 0$.
Thus, if $\priw(\phi|\KB\land \theta)$ exists, it must be 1.
 
For the second half of the theorem,
suppose that $\priw(\theta|\KB)$ exists. Since $\KB \notrwent
\neg
\theta$,
we must have that $\priw(\theta|\KB) = p > 0$.
Therefore, for
all $\epsvec$ sufficiently small and all $N$ sufficiently large
(where ``sufficiently large'' may depend
on $\epsvec$), we can assume that
$\prNw(\theta|\KB) > \epsilon > 0$.  But now, for any such pair
$N,\epsvec$ we can again prove that
$$
\prNw(\neg \phi|\KB\land\theta) \leq
\frac{\prNw(\neg\phi|\KB)}{\prNw(\theta|\KB)}.
$$
Taking the limit, we obtain that $\priw(\neg\phi|\KB \land \theta)$
must also have a limit that must be 0. Hence
$\priw(\phi|\KB\land\theta) = 1$,
as desired.
\eprf

\othm{direct}
Let $\KB$ be a knowledge base of the form $\psi(\ctuple) \land \KBp$,
and assume that for all sufficiently small tolerance vectors $\epsvec$,
$$
\KB[\epsvec] \sat \cprop{\phi(\xtuple)}{\psi(\xtuple)}{\xtuple} \in
[\alpha,\beta].
$$
If no constant in $\ctuple$ appears in $\KBp$, in $\phi(\xtuple)$, or
in $\psi(\xtuple)$, then $\priw(\phi(\ctuple)|\KB) \in
[\alpha,\beta]$,
provided the degrees of belief exist.%
\footnote{The degree of belief may not exist since
$\lim_{\epsvec \tendsto \zerovec}
\liminf_{N \tendsto \infty}\, \prNw(\phi|\KB)$ may not be equal to
$\lim_{\epsvec \tendsto \zerovec}
\limsup_{N \tendsto \infty}\, \prNw(\phi|\KB)$.  However, it follows
{from} the proof of the theorem that both
these limits
lie in the interval $[\alpha,\beta]$.
A similar remark holds for
many of our later results.}
\eothm
\prf
First, fix any sufficiently small tolerance vector $\epsvec$, and
consider a domain size $N$ for which $\KB[\epsvec]$ is satisfiable.
The proof strategy is to partition the size $N$ worlds that
satisfy $\KB[\epsvec]$ into disjoint
clusters and then prove that, within each cluster, the probability of
$\phi(\ctuple)$
is in the range
$[\alpha,\beta]$. From this, we can show that the (unpartitioned)
probability is in this range also.

The size $N$ worlds satisfying $\KB[\epsvec]$ are partitioned so that
two worlds are in the same cluster if and only if they
agree on the denotation of all symbols
in the vocabulary $\vocab$ except for the constants in $\ctuple$.
Now consider one such cluster, and let $A \subseteq \{1,\ldots,N\}^k$
be the denotation of $\psi(\xtuple)$ inside the cluster.  That is, if
$\world$ is a world in the cluster, then
$$
A =
\{(d_1,\ldots,d_k) \in \{1,\ldots,N\}^k\ :\
(\world,\val[x_{i_1}/d_1,\ldots,x_{i_k}/d_k],\epsvec) \sat
          \psi(\xtuple) \}.
$$
Note that, since $\psi(\xtuple)$ does not mention any of the
constants in $\ctuple$, and the denotation of everything else is
fixed throughout the cluster, the set $A$ is
the same in all worlds
$\world$
of the cluster.
Similarly, let $B \subseteq A$
be the denotation of $\phi(\xtuple) \land \psi(\xtuple)$ in the
cluster. Since the worlds in the cluster all satisfy $\KB[\epsvec]$,
and $\KB[\epsvec] \sat \cprop{\phi(\xtuple)}{\psi(\xtuple)}{\xtuple}
\in [\alpha,\beta]$, we know that $|B|/|A| \in [\alpha,\beta]$.
Since none of the constants in $\ctuple$ are mentioned
in $\KB$ except for the statement $\psi(\ctuple)$, each $k$-tuple in
$A$ is a legal denotation for $\ctuple$.  There is precisely one
world in the cluster for each such denotation, and all worlds in the
cluster are of this form. Among those worlds, only those
corresponding to tuples in $B$ satisfy $\phi(\ctuple)$.  Therefore,
the fraction of worlds in the cluster satisfying $\phi(\ctuple)$ is
$|B|/|A| \in [\alpha,\beta]$.
 
The probability $\prNw(\phi(\ctuple)|\KB)$ is a weighted average of the
probabilities within the individual clusters, so it also has to be in
the range $[\alpha,\beta]$.
 
It follows that $\liminf_{N \tendsto \infty}\,
\prNw(\phi(\ctuple)|\KB)$ and $\limsup_{N \tendsto \infty}\,
\prNw(\phi(\ctuple)|\KB)$ are also in the range $[\alpha,\beta]$.
Since this holds for every sufficiently small $\epsvec$, we conclude
that if both limits
$$
\lim_{\epsvec \tendsto \vec{0}}\, \liminf_{N \tendsto \infty}\,
  \prNw(\phi(\ctuple)|\KB)
\mbox{~~and~~}
\lim_{\epsvec \tendsto \vec{0}}\, \limsup_{N \tendsto \infty}\,
   \prNw(\phi|\KB)
$$
exist and are equal, then $\priw(\phi(\ctuple)|\KB)$ has to be
in the range $[\alpha,\beta]$, as desired.
\eprf

\othm{inherit.thm}
Let $c$ be a constant and let $\KB$ be a knowledge base satisfying the
following conditions:
\begin{enumerate}
\item[(a)]
$\KB \sat \psi_0(c)$,
\item[(b)]
for any expression of the form
$\cprop{\phi(x)}{\psi(x)}{x}$ in $\KB$, it is the
case that either $\KB \sat \forall x ( \psi_0(x) \rimp \psi(x))$ or that $\KB \sat
\forall x ( \psi_0(x) \rimp \neg \psi(x))$,
\item[(c)]
the
(predicate,
function,
and constant)
symbols in $\phi(x)$ appear in $\KB$ only on the left-hand side
of the conditionals in the
proportion expressions described in condition~(b),
\item[(d)]
the constant $c$ does not appear in the formula $\phi(x)$.
\end{enumerate}
Assume that for all sufficiently small tolerance vectors $\epsvec$:
$$
\KB[\epsvec] \sat \cprop{\phi(x)}{\psi_0(x)}{x} \in [\alpha,\beta].
$$
Then $\priw(\phi(c)|\KB) \in [\alpha,\beta]$,
provided the degree of belief exists.
\eothm
 
\prf
This theorem is proved with the same general strategy we used for
Theorem~\ref{direct}. That is, for each domain size $N$ and tolerance
vector $\epsvec$, we partition the worlds of size $N$ satisfying
$\KB[\epsvec]$ into clusters and prove that, within each
cluster, the probability of $\phi(c)$ is in the interval
$[\alpha,\beta]$.  As before, this suffices to
prove the result. However the clusters are defined quite differently
in this theorem.
 
We define the clusters as maximal sets
of worlds
satisfying the following
three conditions:
\begin{enumerate}
\item All worlds in a cluster must agree on the denotation
of every vocabulary symbol except possibly those appearing in
$\phi(x)$. Note that, in particular, they agree on the
denotation of the constant $c$. They must also agree as to which elements
satisfy $\psi_0(x)$; let this set be $A_0$.
\item
The denotation of symbols
in $\phi$ must also be constant, except possibly when a member
of $A_0$ is involved.  More precisely, let $\overline{A_0}$ be the set of domain
elements $\{1,\ldots,N\} - A_0$.  Then for any predicate symbol $R$
or function symbol $f$
of arity
$r$ appearing in $\phi(x)$, and for all worlds
$\world'$ and $\world$ in the cluster,
if $d_1, \ldots{}, d_r, d_{r+1}
\in \overline{A_0}$ then $R(d_1,\ldots,d_r)$ holds
in $\world'$ iff it holds in $\world$,
and $f(d_1, \ldots, d_r) = d_{r+1}$ in $\world$ iff $f(d_1, \ldots, d_r)
= d_{r+1}$ in $\world'$.  In particular, this means that
for any constant symbol
$c'$ appearing in $\phi(x)$, if it denotes $d' \in \overline{A_0}$ in
$\world$, then it must denote $d'$ in $\world'$.
\item
All worlds in the cluster are isomorphic with respect to the
vocabulary symbols in $\phi$.  More precisely, if $\world$ and
$\world'$ are two worlds in the cluster, then there exists some
permutation $\perm$ of the domain such that for any predicate symbol
$R$
appearing in $\phi(x)$
and any domain elements $d_1,\ldots,d_r \in
\set{1,\ldots,N}$, $R(d_1,\ldots,d_r)$ holds in $\world$ iff
$R(\perm(d_1),\ldots,\perm(d_r))$ holds in $\world'$,
and similarly for function symbols.  In particular,
for any constant symbol $c'$ appearing in $\phi(x)$, if it denotes
$d'$ in $\world$, then it denotes $\perm(d')$ in $\world'$.
\end{enumerate}
It should be clear
that clusters so defined are mutually exclusive and exhaustive.

We now want to prove that each cluster is, in a precise sense,
symmetric with respect to the elements in $A_0$.  That is, let $\perm$ be
any permutation of the domain which is the identity on any element outside
of $A_0$ (\ie for any $d \not\in A_0$, $\perm(d) = d$).  Let $\world$ be
any world in our cluster, and let $\world'$ be the world
where all the symbols not appearing in $\phi$ get the same interpretation
as they do in $\world$, while
the interpretation of the symbols appearing in $\phi$ is obtained from
their interpretation in $\world$ by applying $\perm$ as described above.
We want to prove that $\world'$ is also in the
cluster.   Condition~(1) is an immediate consequence of the definition of
$\world'$; the restriction on the choice of $\perm$ implies
condition~(2);
condition~(3) holds by definition.
It remains only to prove that $\world' \sat \KB[\epsvec]$.
Because of condition~(c) in the statement of the theorem, and the
fact that vocabulary symbols not in $\phi$ have the same denotation in
$\world$ and in $\world'$, this can
fail to happen only if some expression $\cprop{\phi(x)}{\psi(x)}{x}$ has
different values in $\world$ and in $\world'$.  We show that this is
impossible.

It is easy to see that
for all domain elements $d$, we have $(\world,V,\epsvec) \sat \psi(x)$
iff $(\world',V,\epsvec) \sat \psi(x)$ (where $V$ is a valuation
mapping $x$ to $d$),  since the symbols not in $\phi$
get the same interpretation in both $\world$ and $\world'$.  On the other
hand, if $\phi'$ is a formula that mentions only the symbols appearing
in $\phi$, then a straightforward induction on the structure of $\phi'$
can be used to show that $(\world,V,\epsvec) \sat \phi'(x)$
iff $(\world',\perm\circ V,\epsvec) \sat \phi'(x)$, where
$\perm \circ V$ is the valuation that maps
$x$ to $\perm(V(x))$.  Thus, if $B$ is the set of elements satisfying
$\phi(x)$ in $\world$, then $\perm(B)$ is the set of elements satisfying
$\phi(x)$ in $\world'$.
Let $A$ be the set of domain elements satisfying $\psi(x)$
for worlds in this cluster.
We want to show that $|B \inter A|/|A| = |\perm(B) \inter A|/|A|$
or, equivalently, that $|B\inter A| = |\perm(B) \inter A|$.
By our observations above, the set of domain elements satisfying
$\phi(x) \land \psi(x)$
in $\world'$
is $\perm(B) \inter A$. By condition~(b) there are only two
cases: Either $\KB \sat \forall x(\psi_0(x) \rimp \neg \psi(x))$, in
which case $A_0$ and $A$ are disjoint, or $\KB \sat \forall
x(\psi_0(x) \rimp \psi(x))$, so that $A_0 \subseteq A$.  In the first
case, since $\perm$ is the identity off
$A_0$,
it is easy to see that $\perm(B) \inter A = B \inter A$, and we are done.
In the second case,
because $A_0 \subseteq A$,
$\perm$ is a permutation of $A$ into itself, so
we must still have $|\pi(B) \inter A| = |B \inter A|$.
We conclude that
$\world'$ does satisfy $\KB[\epsvec]$, and is therefore also in the
cluster.  Since we restricted the cluster to consist only of worlds
that are isomorphic to $\world$ in the above sense, and we have now
proved that all worlds formed in this way are in the cluster, the
cluster contains precisely all such worlds.
 
Having defined the clusters, we want to show that the degree
of belief of $\phi(c)$ is in the range $[\alpha,\beta]$ when we
look at any single cluster.
By assumption, $\KB[\epsvec] \sat \cprop{\phi(x)}{\psi_0(x)}{x} \in
[\alpha,\beta]$.  Therefore, for each world in the cluster, the subset
of the elements of $A_0$ that satisfy $\phi(x)$ is in the interval
$[\alpha,\beta]$.  Moreover, by condition~(a), $\KB$ also
entails the assertion $\psi_0(c)$.  Therefore, the denotation
of $c$ is some domain element $d$ in $A_0$. Condition~(d)
says that $c$ does not appear in $\phi$, and so the denotation
of $c$ is the same for all worlds in the cluster.
Now consider a world $\world$ in the cluster, and let $B$ be the
subset of $A_0$ whose members satisfy $\phi(x)$ in $\world$.  We have shown
that every permutation of the elements in $A_0$ (leaving the remaining
elements constant) has a corresponding world in the cluster.
In particular, all possible subsets $B'$ of size $|B|$ are possible
denotations for $\phi(x)$ in worlds in the cluster.  Furthermore, because of
symmetry, they are all equally likely.  It follows that the fixed element
$d$ satisfies $\phi(x)$ in precisely $|B|/|A_0|$ of the worlds in the
cluster. Since $|B|/|A_0| \in [\alpha, \beta]$, the probability of
$\phi(c)$ in any one cluster is in this range also.
 
As in Theorem~\ref{direct}, the truth of this fact for each cluster
implies its truth in general and at the limit. In particular, since
$\KB[\epsvec] \sat \cprop{\phi(x)}{\psi_0(x)}{x} \in [\alpha,\beta]$
for every sufficiently small $\epsvec$, we conclude that
$\priw(\phi(c)|\KB) \in [\alpha,\beta]$, if the limit exists.
\eprf
 
\othm{strength}
Suppose $\KB$ has the form
$$\bigwedge_{i = 1}^m
(\alpha_i \aleq_{\ell_i} \cprop{\phi(x)}{\psi_i(x)}{x} \aleq_{r_i} \beta_i)
\; \land \; \psi_1(c) \; \land \; \KBp,$$
and, for all $i$, $\KB \sat \forall x\; (\psi_i(x) \rimp \psi_{i+1}(x))
\land \neg (\prop{\psi_1(x)}{x} \aeq_1 0)$.
Assume also that no symbol appearing $\phi(x)$ appears in
$\KBp$ or in any $\psi_i(c)$. Further suppose that, for some $j$,
$[\alpha_j, \beta_j]$ is the tightest interval.
That is, for all $i\neq j$, $\alpha_i < \alpha_j < \beta_j < \beta_i$.
Then,
if the degree of belief exists,
$$\priw(\phi(c) | \KB) \in [\alpha_j, \beta_j].$$
\eothm
 
\prf
The proof of the theorem is based on the following result.
Consider any $\KB$ of the form
$$\neg (\prop{\psi'(x)}{x} \aeq_1 0) \;\land\;
\forall x (\psi'(x) \rimp \psi(x)) \;\land\;
\alpha \aleq_\ell \cprop{\phi(x)}{\psi(x)}{x} \aleq_{r} \beta \;\land\; \KBp
$$
where none of $\KBp$, $\psi(x), \psi'(x)$ mention any symbol appearing
in $\phi(x)$. Then, for any $\epsilon > 0$,
$$\Prinf(\alpha-\epsilon \leq \cprop{\phi(x)}{\psi'(x)}{x} \leq
\beta+\epsilon \;|\; \KB) = 1.$$
 
Note that this is quite similar in spirit to Theorem~\ref{inherit.thm},
where we proved that (under certain conditions) an individual $c$
satisfying $\psi(c)$ ``inherits'' the statistics over $\psi(x)$; that
is, the degree of belief is derived from these statistics.
Not surprisingly, the proof of the new
result is similar to that of Theorem~\ref{inherit.thm}, and we refer
the reader to that proof for many of the details.
 
We begin by clustering worlds exactly as in the earlier proof,
with $\psi(x)$ playing the role of the earlier $\psi_0(x)$.
Now consider any particular cluster and let $A$ be the corresponding
denotation of $\psi(x)$. In the cluster, the proportion of $A$ that
satisfies $\phi(x)$ is some $\gamma$ such that
$\alpha-\epscom_\ell \leq \gamma \leq \beta+\epscom_r$.  (Recall that
$\epscom_\ell$ and $\epscom_r$ are the tolerances associated with the
approximate comparisons $\aeq_\ell$ and $\aeq_r$ in $\KB$).
In this cluster, the denotation of $\phi(x)$ in $A$ ranges over subsets
of $A$ of size $\gamma |A|$.  From the proof of
Theorem~\ref{inherit.thm}, we know that there is, in fact, an equal
number of worlds in the cluster corresponding to every such subset.
 
Now let $A'$ be the denotation of $\psi'(x)$ in the cluster (recall that
it follows from the construction of the clusters that all worlds in a
cluster have the same denotation for $\psi'(x)$).
For a proportion $\gamma' \in [0,1]$, we are interested in computing the
fraction of worlds in the cluster such that the proportion of $\phi(x)$
in $A'$ is $\gamma'$.  From our discussion above, it follows that
this is a purely combinatorial question: given a set $A$ of size $n$ and a
subset $A'$ of size $n'$, how many ways are there of choosing
$\gamma n$ elements (representing the elements for which $\phi(x)$
holds) so that $\gamma' n'$ elements come from $A'$?
We estimate this using the observation that the distribution of $\gamma'
n'$ is derived from a process of sampling without replacement.%
\footnote{There are, in fact, a number of ways to solve this problem.
One alternative is to use an entropy-based technique.  We can do this
because, at this point in the proof, it no longer matters whether $\KB$
uses nonunary predicates or not; we can therefore safely apply
techniques that usually only work in the unary case.}
Hence, it behaves according to the well-known hypergeometric
distribution (see, for example, \cite{LarsenMarx}).  We can thus
conclude that $\gamma'$ is distributed with mean $\gamma$ and variance
 $$\frac{\gamma (1-\gamma) (n - n')}{(n - 1) n'} \leq
       \frac{\gamma (1-\gamma)}{n'} \leq \frac{1}{4 n'}\ .$$
Since $\KB \sat \neg(\prop{\psi'(x)}{x} \aeq_1
0)$, we know that $n' = |A'| \geq \eps_1 N$.  Thus, this variance tends
to 0 as $N$ grows large.  Now, consider the event: ``a world in the
cluster has a proportion of $\phi(x)$ within $A'$ which is not in the
interval $[\gamma-\epsilon,\gamma+\epsilon]$''. By Chebychev's
inequality, this is bounded from above by some small probability $p_N$
which depends only on $\eps_1 N$.  That is, the fraction of worlds in
each cluster that have the ``wrong'' proportion is at most $p_N$.  Since
this is the case for every cluster, it is also true in general.  More
precisely, the fraction of overall worlds for which
$\cprop{\phi(x)}{\psi'(x)}{x} \not\in [\alpha-\eps_\ell-\epsilon,
   \beta+\eps_r+\epsilon]$ is at most $p_N$.  But this probability goes
to 0 as $N$ tends to infinity.  Therefore,
  \begin{displaymath}
    \Prinfeps(\alpha-\epscom_\ell-\epsilon \leq
        \cprop{\phi(x)}{\psi'(x)}{x} \leq \beta+\epscom_r+\epsilon \;|\; \KB) = 1.
  \end{displaymath}
As $\epsvec \tendsto \zerovec$ we can simply omit $\epscom_\ell$ and
$\epscom_r$, proving the required result.
 
It is now a simple matter to prove the theorem itself.
Consider the following modification $\KBpp$ of the $\KB$ given in the
statement of the theorem:
$$\bigwedge_{i = j}^m
(\alpha_i \aleq_{\ell_i} \cprop{\phi(x)}{\psi_i(x)}{x} \aleq_{r_i} \beta_i)
\; \land \; \psi_1(c) \; \land \; \KBp,$$
where we eliminate the statistics for the reference classes that are
contained in $\psi_j$ (the more specific reference classes).  From
Theorem~\ref{inherit.thm} we can conclude that $\Prinf(\phi(c)|\KBpp)
\in [\alpha_j,\beta_j]$ (the conditions of that theorem are clearly
satisfied).  But we also know, from the result above, that for each
$\psi_i$, for $i < j$:
  \begin{displaymath}
    \Prinf(\alpha_j-\epsilon \leq \cprop{\phi(x)}{\psi'(x)}{x} \leq
          \beta_j+\epsilon \;|\; \KBpp) = 1.
  \end{displaymath}
For sufficiently small $\epsilon > 0$, the assertion that
$$ \alpha_j-\epsilon \leq \cprop{\phi(x)}{\psi'(x)}{x} \leq
    \beta_j+\epsilon $$
logically implies that
$$ \alpha_i \aleq_{\ell_i} \cprop{\phi(x)}{\psi'(x)}{x} \aleq_{r_i}
    \beta_i, $$
so that this latter assertion also has probability 1 given $\KBpp$.
We therefore also have probability 1 (given $\KBpp$) in the finite conjunction
$$ \band_{i=1}^j (\alpha_i \aleq_{\ell_i} \cprop{\phi(x)}{\psi'(x)}{x}
     \aleq_{r_i} \beta_i). $$
We can now apply Theorem~\ref{cmc} to conclude that we can add this
finite conjunction to $\KBpp$ without affecting any of the degrees of
belief.  But the knowledge base resulting from adding this conjunction
to $\KBpp$ is precisely the original $\KB$.  We conclude that
  \begin{displaymath}
    \Prinf(\phi(c)|\KB) = \Prinf(\phi(c) | \KBpp) \in [\alpha_j,\beta_j],
  \end{displaymath}
as required.
\eprf
 
\othm{dempster.rule}
Let $P$ be a unary predicate, and consider a knowledge base $\KB$ of the
following form:
$$
\band_{i=1}^m
  \left(\cprop{P(x)}{\psi_i(x)}{x} \aeq_i \alpha_i \land \psi_i(c)\right)
 \ \land\
\band_{\stackrel{i,j = 1}{\scriptstyle i \neq j}}^m
       \exists! x (\psi_i(x) \land \psi_j(x)) \ ,
$$
where either $\alpha_i < 1$ for all $i = 1, \ldots, m$, or
$\alpha_i > 0$ for all $i = 1, \ldots, m$.
Then, if neither $P$ nor $c$ appear anywhere in the formulas
$\psi_i(x)$, then
$$
\priw(P(c) | \KB) = \dempster(\alpha_1,\ldots,\alpha_m)
                  = \frac{\prod_{i=1}^m \alpha_i}
                  {\prod_{i=1}^m \alpha_i + \prod_{i=1}^m (1-\alpha_i)}.$$
\eothm
\prf
Assume without loss of generality that $\alpha_i > 0$ for $i = 1, \ldots,
m$.
As in previous theorems, we
prove the result by dividing the worlds into clusters.
More precisely, consider any $\epsvec$ such that
$\alpha_i-\epscom_i > 0$.
Let $\beta_i = \min(\alpha_i + \epscom_i, 1)$.
For any such $\epsvec$ and any domain
size $N$, we divide the worlds of size $N$ satisfying $\KB[\epsvec]$
into clusters, and prove that, within each cluster, the probability
of $\phi(c)$ is in the interval
$[\dempster(\alpha_1-\epscom_1,\ldots,\alpha_m-\epscom_m),
\dempster(\beta_1,\ldots,\beta_m)]$.  Since
$\dempster$ is a continuous function at these points,
this suffices to prove the theorem.

We partition the worlds satisfying $\KB[\epsvec]$ into maximal
clusters that satisfy the following three conditions:
\begin{enumerate}
\item
All worlds in a cluster must agree on the denotation of every
vocabulary symbol except for $P$.  In particular, the denotations of
$\psi_1(x), \ldots, \psi_m(x)$ is fixed.  For $i=1,\ldots,m$, let
$A_i$ denote the denotation of $\psi_i(x)$ in the cluster, and let
$n_i$ denote $|A_i|$.
\item
All worlds in a cluster must have the same denotation of $P$ for
elements in $\overline{A} = \{1,\ldots,N\} - \cup_{i=1}^m A_i$.
\item
For all $i = 1,\ldots,m$, all worlds in the cluster must have the
same number of elements $r_i$ satisfying $P$ within each set $A_i$.
Note that, since all worlds in the cluster satisfy $\KB[\epsvec]$, it
follows that $r_i/n_i \in [\alpha_i-\epscom_i,\beta_i]$
for $i=1,\ldots,m$.
\end{enumerate}
 
Now, consider a cluster as defined above.  The assumptions of the
theorem imply that, besides the proportion
constraints defined by the numbers $r_i$, there are no other constraints
on the denotation of $P$ within the sets $A_1,\ldots,A_m$. Therefore, all
possible denotations of $P$ satisfying these constraints are possible.
Let $d$ be the denotation of $c$ in this cluster.
Our assumptions guarantee that $d$ is the only member of $A_i \inter A_j$.
Hence, the number of elements of $A_i$ for which $P$ has not yet been
chosen is $n_i-1$.  In worlds that satisfy $P(c)$, precisely $r_i-1$
of these elements must satisfy $P$. Since the $A_i$ are disjoint
except for $d$, the choice of $P$ within each $A_i$ can be made
independently of the other choices. Therefore, the number of
worlds in the cluster where $P(c)$ holds is
$$
\prod_{i=1}^m {{n_i-1} \choose {r_i-1}}.
$$
Similarly, the number of worlds in the cluster for which $P(c)$ does not
hold is
$$
\prod_{i=1}^m {{n_i-1} \choose {r_i}}.
$$
Therefore, the fraction of worlds in the cluster satisfying $P(c)$ is:
\begin{eqnarray*}
\frac{\prod_{i=1}^m {{n_i-1} \choose {r_i-1}}}
   {\prod_{i=1}^m {{n_i-1} \choose {r_i-1}} +
       \prod_{i=1}^m {{n_i-1} \choose {r_i}}}  & = &
    \frac{\prod_{i=1}^m r_i}
         {\prod_{i=1}^m r_i + \prod_{i=1}^m (n_i - r_i)} \\
 & = & \frac{\prod_{i=1}^m r_i/n_i}
         {\prod_{i=1}^m r_i/n_i + \prod_{i=1}^m (n_i - r_i)/n_i} \\
 & = & \dempster(r_1/n_1,\ldots,r_m/n_m) \ .
\end{eqnarray*}
Since $\dempster$ is easily seen to be monotonically increasing in each
of its arguments and $r_i/n_i \in [\alpha_i - \epscom_i,
\beta_i]$, we must have that
$\dempster(r_1/n_1,\ldots,r_m/n_m)$ is in the interval
$[\dempster(\alpha_1-\epscom_1,\ldots,\alpha_m-\epscom_m),
\dempster(\beta_1,\ldots,\beta_m)]$.  Using
the same argument as in the previous theorems and the continuity of
$\dempster$, we deduce the desired result.
\eprf
 
\othm{independent}
Let $\vocab_1$ and $\vocab_2$ be two subvocabularies of $\vocab$
that are disjoint except
for the constant $c$.  Consider $\KB_1, \phi_1 \in
\cL(\vocab_1)$ and $\KB_2, \phi_2 \in \cL(\vocab_2)$. Then
$$
\priw(\phi_1 \land \phi_2 | \KB_1 \land \KB_2) =
  \priw(\phi_1|\KB_1) \times \priw(\phi_2|\KB_2).
$$
\eothm
\prf
Fix $N$, $\epsvec$, and $d$ with $1 \le d \le N$.
Given a vocabulary $\Psi$ containing $c$,
let $\worldsd{\Psi}(\xi)$ consist of all worlds in $\WN(\Psi)$
such that $(\world,\epsvec) \sat \xi$ and the denotation of
$c$ in $\world$ is $d$, and let $\nworldsd{\Psi}(\xi) = |\worldsd{\Psi}
(\xi)|$.%
\footnote{Note that we are careful to mention the vocabulary in the
superscript here, rather than suppressing it as we have up to now.
This is because the vocabulary used plays a significant role in this
proof.}
It should be clear that for each choice of $d$,
the sets $\worldsd{\vocab}(\xi)$
have equal size.
Thus, $\nworldsepsv{\vocab}(\xi) = N\nworldsd{\vocab}(\xi)$.
If $\xi_1$ is a formula in $\Phi_1$ and
$\xi_2$ is a formula in $\Phi_2$, then there is clearly a bijection
between
$\worldsd{\Phi_1 \union \Phi_2}(\xi_1 \land \xi_2)$ and
$\worldsd{\Phi_1}(\xi_1) \times \worldsd{\Phi_2}(\xi_2)$.
It follows that
$\nworldsd{\Phi_1 \union \Phi_2}(\xi_1 \land \xi_2) =
\nworldsd{\Phi_1}(\xi_1) \times \nworldsd{\Phi_2}(\xi_2)$.
Since $\nworldsepsv{\vocab}(\xi) = N\nworldsd{\vocab}(\xi)$,  we
immediately get
\begin{eqnarray*}
\lefteqn{\PrNepsv{\vocab_1 \union \vocab_2}(\phi_1 \land \phi_2|
\KB_1 \land \KB_2)}\\
& = &\frac{\nworldsepsv{\vocab_1 \union \vocab_2}(\phi_1 \land \KB_1 \land
 \phi_2 \land \KB_2)}{\nworldsepsv{\vocab_1 \union\vocab_2}(\KB_1 \land
\KB_2)}\\
& = &\frac{\nworldsd{\vocab_1 \union \vocab_2}(\phi_1 \land \KB_1 \land
 \phi_2 \land \KB_2)}{\nworldsd{\vocab_1
 \union\vocab_2}(\KB_1 \land \KB_2)}\\
& = &\frac{\nworldsd{\vocab_1}(\phi_1 \land \KB_1) \times
\nworldsd{\vocab_2}(\phi_2 \land \KB_2)}%
{\nworldsd{\vocab_1}(\KB_1) \times \nworldsd{\vocab_2}(\KB_2)}\\
& = &\frac{\nworldsepsv{\vocab_1}(\phi_1 \land \KB_1) \times
\nworldsepsv{\vocab_2}(\phi_2 \land \KB_2)}%
{\nworldsepsv{\vocab_1}(\KB_1) \times \nworldsepsv{\vocab_2}(\KB_2)}\\
& = &\PrNepsv{\vocab_1}(\phi_1|\KB_1) \times \PrNepsv{\vocab_2}(\phi_2|\KB_2).
\end{eqnarray*}
Taking limits, we get that
$\Prinfv{\vocab_1 \union \vocab_2}(\phi_1 \land \phi_2|
\KB_1 \land \KB_2) =
\Prinfv{\vocab_1}(\phi_1|\KB_1) \cdot \Prinfv{\vocab_2}(\phi_2|\KB_2)$.
As observed in \cite{GHK1a}, for all formulas $\phi$ and $\KB$, if
$\vocab \supseteq \vocab'$, then $\Prinfv{\vocab}(\phi|\KB) =
\Prinfv{\vocab'}(\phi|\KB)$.  (Intuitively, this is because the effect
of changing the vocabulary cancels out in the numerator and
denominator.)
We thus get
$\priw(\phi_1 \land \phi_2 | \KB_1 \land \KB_2) =
  \priw(\phi_1|\KB_1) \cdot \priw(\phi_2|\KB_2)$, as desired.
\eprf
 
\bibliographystyle{alpha}
\bibliography{z,bghk}
\end{document}

%% file: defn.tex
\newtheorem{THEOREM}{Theorem}[section]
\newenvironment{theorem}{\begin{THEOREM} \hspace{-.85em} {\bf :} }%
                        {\end{THEOREM}}
\newtheorem{LEMMA}[THEOREM]{Lemma}
\newenvironment{lemma}{\begin{LEMMA} \hspace{-.85em} {\bf :} }%
                      {\end{LEMMA}}
\newtheorem{COROLLARY}[THEOREM]{Corollary}
\newenvironment{corollary}{\begin{COROLLARY} \hspace{-.85em} {\bf :} }%
                          {\end{COROLLARY}}
\newtheorem{PROPOSITION}[THEOREM]{Proposition}
\newenvironment{proposition}{\begin{PROPOSITION} \hspace{-.85em} {\bf :} }%
                            {\end{PROPOSITION}}
\newtheorem{DEFINITION}[THEOREM]{Definition}
\newenvironment{definition}{\begin{DEFINITION} \hspace{-.85em} {\bf :} \rm}%
                            {\end{DEFINITION}}
\newtheorem{CLAIM}[THEOREM]{Claim}
\newenvironment{claim}{\begin{CLAIM} \hspace{-.85em} {\bf :} \rm}%
                            {\end{CLAIM}}
\newtheorem{EXAMPLE}[THEOREM]{Example}
\newenvironment{example}{\begin{EXAMPLE} \hspace{-.85em} {\bf :} \rm}%
                            {\end{EXAMPLE}}
\newtheorem{REMARK}[THEOREM]{Remark}
\newenvironment{remark}{\begin{REMARK} \hspace{-.85em} {\bf :} \rm}%
                            {\end{REMARK}}
\newcommand{\thm}{\begin{theorem}}
\newcommand{\lem}{\begin{lemma}}
\newcommand{\pro}{\begin{proposition}}
\newcommand{\dfn}{\begin{definition}}
\newcommand{\rem}{\begin{remark}}
\newcommand{\xam}{\begin{example}}
\newcommand{\cor}{\begin{corollary}}
\newcommand{\prf}{\noindent{\bf Proof:} }
\newcommand{\ethm}{\end{theorem}}
\newcommand{\elem}{\end{lemma}}
\newcommand{\epro}{\end{proposition}}
\newcommand{\edfn}{\bbox\end{definition}}
\newcommand{\erem}{\bbox\end{remark}}
\newcommand{\exam}{\bbox\end{example}}
\newcommand{\ecor}{\end{corollary}}
\newcommand{\eprf}{\bbox\vspace{0.1in}}
\newcommand{\beqn}{\begin{equation}}
\newcommand{\eeqn}{\end{equation}}
\newcommand{\bbox}{\vrule height7pt width4pt depth1pt}
\newcommand{\qed}{\eprf}
\newcommand{\clm}{\begin{claim}}
\newcommand{\eclm}{\end{claim}}
\newcommand{\rarrow}{\rightarrow}
\newcommand{\sat}{\models}

\newcommand{\rimp}{\Rightarrow}
\newcommand{\dimp}{\Leftrightarrow}
\newcommand{\bor}{\bigvee}
\newcommand{\band}{\bigwedge}
\newcommand{\union}{\cup}
\newcommand{\inter}{\cap}
\newcommand{\IR}{\mbox{$I\!\!R$}}

\renewcommand{\phi}{\varphi}
\newcommand{\R}{{\cal R}}

\newcommand{\W}{{\cal W}}
\newcommand{\X}{{\cal X}}

\newcommand{\set}[1]{\left\{ #1 \right\}}

\renewcommand{\>}{\rangle}

\newcommand{\eg}{e.g.,~}
\newcommand{\ie}{i.e.,~}

\newcommand{\ol}{\setlength{\itemsep}{0pt}\begin{enumerate}}
\newcommand{\eol}{\end{enumerate}\setlength{\itemsep}{-\parsep}}
\newcommand{\ul}{\setlength{\itemsep}{0pt}\begin{itemize}}
\newcommand{\dl}{\setlength{\itemsep}{0pt}\begin{description}}
\newcommand{\edl}{\end{description}\setlength{\itemsep}{-\parsep}}
\newcommand{\eul}{\end{itemize}\setlength{\itemsep}{-\parsep}}

\newcommand\eqdef{=_{\rm def}}
\newcommand{\true}{\mbox{{\it true}}}

\newcommand{\commentout}[1]{}

\newcommand{\bi}{\begin{itemize}}
\newcommand{\ei}{\end{itemize}}
\newcommand{\be}{\begin{enumerate}}
\newcommand{\ee}{\end{enumerate}}

%% file: bghkmac.tex
\newenvironment{oldthm}[1]{\par\noindent{\bf Theorem #1:} \em \noindent}{\par}
\newenvironment{oldlem}[1]{\par\noindent{\bf Lemma #1:} \em \noindent}{\par}
\newenvironment{oldcor}[1]{\par\noindent{\bf Corollary #1:} \em \noindent}{\par}
\newenvironment{oldpro}[1]{\par\noindent{\bf Proposition #1:} \em \noindent}{\par}
\newcommand{\othm}[1]{\begin{oldthm}{\ref{#1}}}
\newcommand{\eothm}{\end{oldthm} \medskip}
\newcommand{\olem}[1]{\begin{oldlem}{\ref{#1}}}
\newcommand{\eolem}{\end{oldlem} \medskip}
\newcommand{\ocor}[1]{\begin{oldcor}{\ref{#1}}}
\newcommand{\eocor}{\end{oldcor} \medskip}
\newcommand{\opro}[1]{\begin{oldpro}{\ref{#1}}}
\newcommand{\eopro}{\end{oldpro} \medskip}

\newcommand{\world}{W}
\newcommand{\WN}{\W_N}

\newcommand{\tends}{\rightarrow}
\newcommand{\tendsto}{\tends}

\newcommand{\nworldsv}[1]{{\it \#worlds}^{#1}}
\newcommand{\nworlds}{{{\it \#worlds}}_{N}^{\epsvec}}

\newcommand{\Vector}[1]{{\langle #1 \rangle}}
\newcommand{\Prinfv}[1]{{\Pr}^{#1}_\infty}

\newcommand{\Prinf}{{\Pr}_\infty}

\newcommand{\prNw}{{\Pr}_{N}^{\epsvec}}
\newcommand{\prNwi}{{\Pr}_{N^i}^{\epsvec^i}}
\newcommand{\priw}{{\Pr}_{\infty}}
\newcommand{\beliefprob}[1]{\Pr(#1)} %daphne1 - changed macro
\newcommand{\Prinfeps}{{{\Pr}_\infty^{\epsvec}}}

\newcommand{\vocab}{\Phi}

\newcommand{\cL}{{\cal L}}

\newcommand{\Laeq}{{\cal L}^{\aeq}}
\newcommand{\Leq}{{\cal L}^{=}}

\newcommand{\KB}{{\it KB}}
\newcommand{\KBfly}{\KB_{\mbox{\scriptsize \it fly}}}
\newcommand{\KBchirps}{\KB_{\mbox{\scriptsize \it chirps}}}
\newcommand{\KBmagpie}{\KB_{\mbox{\scriptsize \it magpie}}}
\newcommand{\KBhep}{\KB_{\mbox{\scriptsize \it hep}}}
\newcommand{\KBnixon}{\KB_{\mbox{\scriptsize \it Nixon}}}
\newcommand{\KBel}{\KB_{\mbox{\scriptsize \it likes}}}
\newcommand{\KBtax}{\KB_{\mbox{\scriptsize \it taxonomy}}}
\newcommand{\KBarm}{\KB_{\mbox{\scriptsize \it arm}}}
\newcommand{\KBlate}{\KB_{\mbox{\scriptsize \it late}}}
\newcommand{\KBp}{{\KB'}}
\newcommand{\KBflyp}{\KB_{\mbox{\scriptsize \it fly}}'}
\newcommand{\KBpp}{{\KB''}}
\newcommand{\KBdef}{\KB_{\mbox{\scriptsize \it def}}}
\newcommand{\KBdishep}{\KB_{\mbox{\scriptsize \it $\lor$hep}}}

\newcommand{\quak}{\mbox{\it Quaker\/}}
\newcommand{\repub}{\mbox{\it Republican\/}}
\newcommand{\pac}{\mbox{\it Pacifist\/}}
\newcommand{\Nixon}{\mbox{\it Nixon\/}}

\newcommand{\Winner}{\mbox{\it Winner\/}}
\newcommand{\Child}{\mbox{\it Child\/}}

\newcommand{\Tall}{\mbox{\it Tall\/}}
\newcommand{\Elephant}{\mbox{\it Elephant\/}}

\newcommand{\Yellow}{\mbox{\it Yellow\/}}
\newcommand{\Clyde}{\mbox{\it Clyde\/}}
\newcommand{\Tweety}{\mbox{\it Tweety\/}}
\newcommand{\Opus}{\mbox{\it Opus\/}}
\newcommand{\Bird}{\mbox{\it Bird\/}}
\newcommand{\Penguin}{{\it Penguin\/}}
\newcommand{\Fish}{\mbox{\it Fish\/}}
\newcommand{\Fly}{\mbox{\it Fly\/}}
\newcommand{\Warmblooded}{\mbox{\it Warm-blooded\/}}
\newcommand{\White}{\mbox{\it White\/}}
\newcommand{\Red}{\mbox{\it Red\/}}

\newcommand{\Visible}{\mbox{\it Easy-to-see\/}}
\newcommand{\Bat}{\mbox{\it Bat\/}}
\newcommand{\Blue}{\mbox{\it Blue\/}}
\newcommand{\Fever}{\mbox{\it Fever\/}}
\newcommand{\Jaun}{\mbox{\it Jaun\/}}
\newcommand{\Hep}{\mbox{\it Hep\/}}
\newcommand{\Eric}{\mbox{\it Eric\/}}
\newcommand{\Alice}{\mbox{\it Alice\/}}
\newcommand{\Tom}{\mbox{\it Tom\/}}

\newcommand{\Zookeeper}{\mbox{\it Zookeeper\/}}
\newcommand{\Fred}{\mbox{\it Fred\/}}
\newcommand{\Likes}{\mbox{\it Likes\/}}
\newcommand{\Day}{\mbox{\it Day\/}}
\newcommand{\Nextday}{\mbox{\it Next-day\/}}
\newcommand{\Sleepslate}{\mbox{\it To-bed-late\/}}
\newcommand{\Riseslate}{\mbox{\it Rises-late\/}}
\newcommand{\TS}{\mbox{\it TS\/}}
\newcommand{\EEJ}{\mbox{\it EEJ\/}}
\newcommand{\FC}{\mbox{\it FC\/}}

\newcommand{\Ab}{\mbox{\it Ab\/}}
\newcommand{\Chirps}{\mbox{\it Chirps\/}}
\newcommand{\Swims}{\mbox{\it Swims\/}}
\newcommand{\Magpie}{\mbox{\it Magpie\/}}
\newcommand{\Moody}{\mbox{\it Moody\/}}
\newcommand{\SomeMorning}{\mbox{\it Tomorrow\/}}
\newcommand{\Animal}{\mbox{\it Animal\/}}
\newcommand{\Sparrow}{\mbox{\it Sparrow\/}}

\newcommand{\Older}{\mbox{\it Over60\/}}
\newcommand{\Patient}{\mbox{\it Patient\/}}
\newcommand{\Black}{\mbox{\it Black\/}}
\newcommand{\Ray}{\mbox{\it Ray\/}}
\newcommand{\Reiter}{\mbox{\it Reiter\/}}
\newcommand{\Drew}{\mbox{\it Drew\/}}
\newcommand{\McDermott}{\mbox{\it McDermott\/}}
\newcommand{\Emu}{\mbox{\it Emu\/}}
\newcommand{\Canary}{\mbox{\it Canary\/}}

\newcommand{\FlyingBird}{\mbox{\it FlyingBird\/}}
\newcommand{\UL}{\mbox{\it LeftUsable\/}}
\newcommand{\UR}{\mbox{\it RightUsable\/}}
\newcommand{\BL}{\mbox{\it LeftBroken\/}}
\newcommand{\BR}{\mbox{\it RightBroken\/}}
\newcommand{\Ticket}{\mbox{\it Ticket\/}}

\newcommand{\HeartDisease}{{\mbox{\it Heart-disease\/}}}

\newcommand{\bxor}[1]{\dot{\bor}}

\newcommand{\eps}{\tau}
\newcommand{\vareps}{\varepsilon}

\newcommand{\xtuple}{\vec{x}}
\newcommand{\ctuple}{{\vec{c}\,}}

\newcommand{\pvec}{{\!{\vec{\,p}}}}

\newcommand{\zerovec}{\vec{0}}

\newcommand{\epscom}{\eps}
\newcommand{\epsvec}{{\vec{\eps}\/}}
\newcommand{\prop}[2]{{||{#1}||_{{#2}}}}
\newcommand{\aeq}{\approx} %``Approximate'' connectives

\newcommand{\aleq}{\preceq}

\newcommand{\ageq}{\succeq}

\newcommand{\cprop}[3]{{\|{#1}|{#2}\|_{{#3}}}}
\newcommand{\Bigcprop}[3]{{\Bigl\|{#1}\Bigm|{#2}\Bigr\|_{{#3}}}}

\newcommand{\reals}{\IR}
\newcommand{\qsep}{\,}
\newcommand{\perm}{\pi}
\newcommand{\val}{V}
\newcommand{\rwent}{\mbox{$\;|\!\!\!\sim$}_{\mbox{\scriptsize \it rw}}\;}
\newcommand{\notrwent}{\mbox{$\;|\!\!\!\not\sim$}_{\mbox{\scriptsize\it rw}}\;}
\newcommand{\dempster}{\delta}
\newcommand{\dentails}{{\;|\!\!\!\sim\;}}
\newcommand{\notdentails}{{\;|\!\!\!\not\sim\;}}
\newcommand{\dentailssub}[1]{\dentails\hspace{-0.4em}_{
          \mbox{\scriptsize{\it #1}}}\;}
\newcommand{\notdentailssub}[1]{\notdentails\hspace{-0.4em}_{
          \mbox{\scriptsize{\it #1}}}\;}
\newcommand{\default}{\rightarrow}

%% file: ijca93rjcorr.bbl
\newcommand{\etalchar}[1]{$^{#1}$}
\begin{thebibliography}{BGHK94b}

\bibitem[Ada75]{Adams}
E.~Adams.
\newblock {\em The Logic of Conditionals}.
\newblock D. Reidel, Dordrecht, Netherlands, 1975.

\bibitem[Ash93]{Asher:condworkshop}
N.~Asher.
\newblock Extensions for commonsense entailment.
\newblock In {\em Proceedings of the IJCAI Workshop on Conditionals in
  Knowledge Representation}, pages 26--41, 1993.

\bibitem[Bac90]{Bacchus}
F.~Bacchus.
\newblock {\em Representing and Reasoning with Probabilistic Knowledge}.
\newblock MIT Press, Cambridge, Mass., 1990.

\bibitem[BCD{\etalchar{+}}93]{BCDLP:IJCAI}
S.~Benferhat, C.~Cayrol, D.~Dubois, J.~Lang, and H.~Prade.
\newblock Inconsistency management and prioritized syntax-based entailment.
\newblock In {\em Proc.~Thirteenth International Joint Conference on Artificial
  Intelligence (IJCAI '93)}, pages 640--645, 1993.

\bibitem[BGHK92]{BGHK}
F.~Bacchus, A.~J. Grove, J.~Y. Halpern, and D.~Koller.
\newblock From statistics to belief.
\newblock In {\em Proc.~Tenth National Conference on Artificial Intelligence
  (AAAI '92)}, pages 602--608. 1992.

\bibitem[BGHK93]{BGHKnonmon}
F.~Bacchus, A.~J. Grove, J.~Y. Halpern, and D.~Koller.
\newblock Statistical foundations for default reasoning.
\newblock In {\em Proc.~Thirteenth International Joint Conference on Artificial
  Intelligence (IJCAI '93)}, pages 563--569, 1993.
\newblock Available by anonymous ftp from logos.uwaterloo.ca/pub/bacchus or via
  WWW at http://logos.uwaterloo.ca.

\bibitem[BGHK94a]{BGHKcausal}
F.~Bacchus, A.~J. Grove, J.~Y. Halpern, and D.~Koller.
\newblock Forming beliefs about a changing world.
\newblock In {\em Proc.~Twelfth National Conference on Artificial Intelligence
  (AAAI '94)}, pages 222--229, 1994.
\newblock Available by anonymous ftp from logos.uwaterloo.ca/pub/bacchus or via
  WWW at http://logos.uwaterloo.ca.

\bibitem[BGHK94b]{BGHKadddob}
F.~Bacchus, A.~J. Grove, J.~Y. Halpern, and D.~Koller.
\newblock Generating new beliefs from old.
\newblock pages 37--45, 1994.
\newblock Available by anonymous ftp from logos.uwaterloo.ca/pub/bacchus or via
  WWW at http://logos.uwaterloo.ca.

\bibitem[BGHK94c]{BGHKci}
F.~Bacchus, A.~J. Grove, J.~Y. Halpern, and D.~Koller.
\newblock A response to: ``{B}elieving on the basis of evidence''.
\newblock {\em Computational Intelligence}, 10(1):21--25, 1994.

\bibitem[Bou91]{Boutilier:Phd.Thesis}
C.~Boutilier.
\newblock {\em Conditional Logics for Default Reasoning and Belief Revision}.
\newblock PhD thesis, Department of Computer Science, University of Toronto,
  1991.

\bibitem[Car50]{Carnap}
R.~Carnap.
\newblock {\em Logical Foundations of Probability}.
\newblock University of Chicago Press, Chicago, 1950.

\bibitem[Car52]{Carnap2}
R.~Carnap.
\newblock {\em The Continuum of Inductive Methods}.
\newblock University of Chicago Press, Chicago, 1952.

\bibitem[Che83]{Cheeseman83}
P.~C. Cheeseman.
\newblock A method of computing generalized {B}ayesian probability values for
  expert systems.
\newblock In {\em Proc.~Eighth International Joint Conference on Artificial
  Intelligence (IJCAI '83)}, pages 198--202. 1983.

\bibitem[Chu91]{chuaqui}
R.~Chuaqui.
\newblock {\em Truth, possibility, and probability: new logical foundations of
  probability and statistical inference}.
\newblock North-Holland, Amsterdam, 1991.

\bibitem[DD85]{Denbigh}
K.~G. Denbigh and J.~S. Denbigh.
\newblock {\em Entropy in Relation to Incomplete Knowledge}.
\newblock Cambridge University Press, Cambridge, U.K., 1985.

\bibitem[Del88]{Delgrande:Conditional.Logic.Revised}
J.~P. Delgrande.
\newblock An approach to default reasoning based on a first-order conditional
  logic: Revised report.
\newblock {\em Artificial Intelligence}, 36:63--90, 1988.

\bibitem[EKP91]{EKP:scope}
D.~W. Etherington, S.~Kraus, and D.~Perlis.
\newblock Nonmonotonicity and the scope of reasoning.
\newblock {\em Artificial Intelligence}, 2:221--261, 1991.

\bibitem[Gab84]{Gabbay:nonmon.inference}
D.~Gabbay.
\newblock Theoretical foundations for nonmonotonic reasoning in expert systems.
\newblock In K.~R. Apt, editor, {\em Proceedings of the NATO Advanced Study
  Institute on logics and models of concurrent systems}. Springer-Verlag, 1984.

\bibitem[Gef92]{Geffner:thesis}
H.~Geffner.
\newblock {\em Default Reasoning: Causal and Conditional Theories}.
\newblock MIT Press, Cambridge, Mass., 1992.

\bibitem[GHK94]{GHK2}
A.~J. Grove, J.~Y. Halpern, and D.~Koller.
\newblock Random worlds and maximum entropy.
\newblock {\em Journal of A.I. Research}, 2:33--88, 1994.

\bibitem[GHK96a]{GHK1b}
A.~J. Grove, J.~Y. Halpern, and D.~Koller.
\newblock Asymptotic conditional probabilities: the unary case.
\newblock {\em SIAM Journal on Computing}, 25(1):1--51, 1996.

\bibitem[GHK96b]{GHK1a}
A.~J. Grove, J.~Y. Halpern, and D.~Koller.
\newblock Asymptotic conditional probabilities: the non-unary case.
\newblock {\em Journal of Symbolic Logic}, 61(1):250--275, 1996.

\bibitem[GMP90]{GMP}
M.~Goldszmidt, P.~Morris, and J.~Pearl.
\newblock A maximum entropy approach to nonmonotonic reasoning.
\newblock In {\em Proc.~Eighth National Conference on Artificial Intelligence
  (AAAI '90)}, pages 646--652, 1990.

\bibitem[GMP93]{GMPfull}
M.~Goldszmidt, P.~Morris, and J.~Pearl.
\newblock A maximum entropy approach to nonmonotonic reasoning.
\newblock {\em IEEE Transactions of Pattern Analysis and Machine Intelligence},
  15(3):220--232, 1993.

\bibitem[Gol87]{Goldman87}
S.~A. Goldman.
\newblock Efficient methods for calculating maximum entropy distributions.
\newblock Master's thesis, MIT EECS Department, 1987.

\bibitem[Goo92]{Goodwin}
S.~D. Goodwin.
\newblock Second order direct inference: A reference class selection policy.
\newblock {\em International Journal of Expert Systems: Research and
  Applications}, 5(3):185--210, 1992.

\bibitem[GP90]{Geffner:Framework.Reasoning.With.Defaults}
H.~Geffner and J.~Pearl.
\newblock A framework for reasoning with defaults.
\newblock In H.~E. Kyburg, Jr., R.~Loui, and G.~Carlson, editors, {\em
  Knowledge Representation and Defeasible Reasoning}, pages 245--26. Kluwer
  Academic Press, Dordrecht, Netherlands, 1990.

\bibitem[GP92]{Geffner:Conditional.Entailment}
H.~Geffner and J.~Pearl.
\newblock Conditional entailment: bridging two approaches to default reasoning.
\newblock {\em Artificial Intelligence}, 53(2--3):209--244, 1992.

\bibitem[Hac75]{Hacking2}
I.~Hacking.
\newblock {\em The Emergence of Probability}.
\newblock Cambridge University Press, Cambridge, U.K., 1975.

\bibitem[Hal90]{Hal4}
J.~Y. Halpern.
\newblock An analysis of first-order logics of probability.
\newblock {\em Artificial Intelligence}, 46:311--350, 1990.

\bibitem[HK95]{HK95}
J.~Y. Halpern and D.~Koller.
\newblock Representation dependence in probabilistic inference.
\newblock In {\em Proc.~Fourteenth International Joint Conference on Artificial
  Intelligence (IJCAI '95)}, pages 1853--1860, 1995.

\bibitem[HM87]{HM:YSP}
S.~Hanks and S.~McDermott.
\newblock Nonmonotonic logic and temporal projection.
\newblock {\em Artificial Intelligence}, 33(3):379--412, 1987.

\bibitem[Hun89]{Hunter89}
D.~Hunter.
\newblock Causality and maximum entropy updating.
\newblock {\em International Journal of Approximate Reasoning}, 3(1):379--406,
  1989.

\bibitem[Jay78]{Jaynes}
E.~T. Jaynes.
\newblock Where do we stand on maximum entropy?
\newblock In R.~D. Levine and M.~Tribus, editors, {\em The Maximum Entropy
  Formalism}, pages 15--118. MIT Press, Cambridge, Mass., 1978.

\bibitem[Jef68]{Jeffrey68}
R.~C. Jeffrey.
\newblock Probable knowledge.
\newblock In I.~Lakatos, editor, {\em International Colloquium in the
  Philosophy of Science: The Problem of Inductive Logic}, pages 157--185.
  North-Holland, Amsterdam, 1968.

\bibitem[Joh32]{Johnson}
W.~E. Johnson.
\newblock Probability: The deductive and inductive problems.
\newblock {\em Mind}, 41(164):409--423, 1932.

\bibitem[KH92]{KH92}
D.~Koller and J.~Y. Halpern.
\newblock A logic for approximate reasoning.
\newblock In {\em Proc.~Third International Conference on Principles of
  Knowledge Representation and Reasoning (KR '92)}, pages 153--164. 1992.

\bibitem[KH96]{KH.irrel}
D.~Koller and J.Y. Halpern.
\newblock Irrelevance and conditioning in first-order probabilistic logic.
\newblock In {\em Proc.~Thirteenth National Conference on Artificial
  Intelligence (AAAI '96)}, pages 569--576, 1996.

\bibitem[KLM90]{KLM}
S.~Kraus, D.~Lehmann, and M.~Magidor.
\newblock Nonmonotonic reasoning, preferential models and cumulative logics.
\newblock {\em Artificial Intelligence}, 44:167--207, 1990.

\bibitem[Kyb61]{Kyburg:Rational.Belief}
H.~E Kyburg, Jr.
\newblock {\em Probability and the Logic of Rational Belief}.
\newblock Wesleyan University Press, Middletown, Connecticut, 1961.

\bibitem[Kyb74]{Kyburg:Statistical.Inference}
H.~E. Kyburg, Jr.
\newblock {\em The Logical Foundations of Statistical Inference}.
\newblock Reidel, Dordrecht, Netherlands, 1974.

\bibitem[Kyb83]{Kyburg:Reference.Class}
H.~E. Kyburg, Jr.
\newblock The reference class.
\newblock {\em Philosophy of Science}, 50(3):374--397, 1983.

\bibitem[Kyb88]{Kyburg:Full.Beliefs}
H.~E. Kyburg, Jr.
\newblock Full beliefs.
\newblock {\em Theory and Decision}, 25:137--162, 1988.

\bibitem[Lan80]{Landau}
L.~D. Landau.
\newblock {\em Statistical Physics}, volume~1.
\newblock Pergamon Press, 1980.

\bibitem[Lif89]{Lifschitz.bench}
V.~Lifschitz.
\newblock Benchmark problems for formal non-monotonic reasoning, version 2.00.
\newblock In M.~Reinfrank, J.~de~Kleer, M.~L. Ginsberg, and E.~Sandewall,
  editors, {\em Non-Monotonic Reasoning: 2nd International Workshop (Lecture
  Notes in Artificial Intelligence 346)}, pages 202--219. Springer-Verlag,
  1989.

\bibitem[LM81]{LarsenMarx}
R.~J. Larsen and M.~L. Mark.
\newblock {\em An introduction to mathematical statistics and its
  applications}.
\newblock Prentice-Hall, Englewood Cliffs, NJ, 1981.

\bibitem[LM90]{LehmannMagidor:TARK}
D.~Lehmann and M.~Magidor.
\newblock Preferential logics: the predicate calculus case.
\newblock In R.~Parikh, editor, {\em Theoretical Aspects of Reasoning about
  Knowledge: Proc.~Third Conference}, pages 57--72, San Francisco, 1990. Morgan
  Kaufmann.

\bibitem[LM92]{LehMag}
D.~Lehmann and M.~Magidor.
\newblock What does a conditional knowledge base entail?
\newblock {\em Artificial Intelligence}, 55(1):1--60, 1992.

\bibitem[LR57]{LuceR}
R.~D. Luce and H.~Raiffa.
\newblock {\em Games and Decisions}.
\newblock Wiley, New York, 1957.

\bibitem[Mak89]{Makinson:nonmon.inference}
D.~Makinson.
\newblock General theory of cumulative inference.
\newblock In M.~Reinfrank, J.~de~Kleer, M.~L. Ginsberg, and E.~Sandewall,
  editors, {\em Non-Monotonic Reasoning: 2nd International Workshop (Lecture
  Notes in Artificial Intelligence 346)}, pages 1--18. Springer-Verlag, 1989.

\bibitem[Mak94]{Makinson:survey}
D.~Makinson.
\newblock General patterns in nonmonotonic reasoning.
\newblock In D.~Gabbay, C.~Hogger, and J.~Robinson, editors, {\em Handbook of
  Logic in Artificial Intelligence and Logic Programming}, volume~3, pages
  35--110. Oxford University Press, 1994.

\bibitem[McC80]{McC}
J.~McCarthy.
\newblock Circumscription---a form of non-monotonic reasoning.
\newblock {\em Artificial Intelligence}, 13:27--39, 1980.

\bibitem[McC86]{McCarthy:Applications.of.Circumscription}
J.~McCarthy.
\newblock Applications of circumscription to formalizing common-sense
  knowledge.
\newblock {\em Artificial Intelligence}, 28:86--116, 1986.

\bibitem[MH69]{MH}
J.~McCarthy and P.~J. Hayes.
\newblock Some philosophical problems from the standpoint of artificial
  intelligence.
\newblock In D.~Michie, editor, {\em Machine Intelligence 4}, pages 463--502.
  Edinburgh University Press, Edinburgh, 1969.

\bibitem[Moo85]{Moore85}
R.~C. Moore.
\newblock Semantical considerations on nonmonotonic logic.
\newblock {\em Artificial Intelligence}, 25:75--94, 1985.

\bibitem[Mor93]{Morreau:condworkshop}
M.~Morreau.
\newblock The conditional logic of generalizations.
\newblock In {\em Proceedings of the IJCAI Workshop on Conditionals in
  Knowledge Representation}, pages 108--118, 1993.

\bibitem[Pea88]{Pearl}
J.~Pearl.
\newblock {\em Probabilistic Reasoning in Intelligent Systems}.
\newblock Morgan Kaufmann, San Francisco, 1988.

\bibitem[Pea89]{Pearl90}
J.~Pearl.
\newblock Probabilistic semantics for nonmonotonic reasoning: a survey.
\newblock In {\em Proc.~First International Conference on Principles of
  Knowledge Representation and Reasoning (KR '89)}, pages 505--516, 1989.
\newblock Reprinted in G. Shafer and J. Pearl (Eds.), {\em Readings in
  Uncertain Reasoning}, pp.~699--710. San Francisco: Morgan Kaufmann, 1990.

\bibitem[Pea90]{Pearl:Z}
J.~Pearl.
\newblock System {Z}: A natural ordering of defaults with tractable
  applications to nonmonotonic reasoning.
\newblock In {\em Theoretical Aspects of Reasoning about Knowledge: Proc.~Third
  Conference}, pages 121--135. 1990.

\bibitem[Pol90]{Pollock:Induction}
J.~L. Pollock.
\newblock {\em Nomic Probabilities and the Foundations of Induction}.
\newblock Oxford University Press, Oxford, U.K., 1990.

\bibitem[Poo89]{Poole89}
D.~Poole.
\newblock What the lottery paradox tells us about default reasoning.
\newblock In {\em Proc.~First International Conference on Principles of
  Knowledge Representation and Reasoning (KR '89)}, pages 333--340. 1989.

\bibitem[Poo91]{Poole91}
D.~Poole.
\newblock The effect of knowledge on belief: conditioning, specificity and the
  lottery paradox in default reasoning.
\newblock {\em Artificial Intelligence}, 49(1--3):282--307, 1991.

\bibitem[PV89]{PV}
J.~B. Paris and A.~Vencovska.
\newblock On the applicability of maximum entropy to inexact reasoning.
\newblock {\em International Journal of Approximate Reasoning}, 3:1--34, 1989.

\bibitem[RC81]{ReitCris}
R.~Reiter and G.~Criscuolo.
\newblock On interacting defaults.
\newblock In {\em Proc.~Seventh International Joint Conference on Artificial
  Intelligence (IJCAI '81)}, pages 270--276, 1981.

\bibitem[Rei49]{Reichenbach:Theory.Of.Probability}
H.~Reichenbach.
\newblock {\em The Theory of Probability}.
\newblock University of California Press, Berkeley, 1949.
\newblock Translation and revision of German edition, published as {\em
  Wahrscheinlichkeitslehre}, 1935.

\bibitem[Rei80]{reiter}
R.~Reiter.
\newblock A logic for default reasoning.
\newblock {\em Artificial Intelligence}, 13:81--132, 1980.

\bibitem[Sav54]{Savage}
L.~J. Savage.
\newblock {\em Foundations of Statistics}.
\newblock Wiley, New York, 1954.

\bibitem[Sha76]{Shaf}
G.~Shafer.
\newblock {\em A Mathematical Theory of Evidence}.
\newblock Princeton University Press, Princeton, N.J., 1976.

\bibitem[Sha89]{Shastri}
L.~Shastri.
\newblock Default reasoning in semantic networks: a formalization of
  recognition and inheritance.
\newblock {\em Artificial Intelligence}, 39(3):285--355, 1989.

\bibitem[Sha93]{ShaferPC}
G.~Shafer.
\newblock Personal communication, 1993.

\bibitem[SW49]{ShannonWeaver}
C.~Shannon and W.~Weaver.
\newblock {\em The Mathematical Theory of Communication}.
\newblock University of Illinois Press, Urbana-Champaign, Ill., 1949.

\bibitem[THT87]{clash.intuitions}
D.~S. Touretzky, J.~F. Horty, and R.~H. Thomason.
\newblock A clash of intuitions: the current state of nonmonotonic multiple
  inheritance systems.
\newblock In {\em Proc.~Tenth International Joint Conference on Artificial
  Intelligence (IJCAI '87)}, pages 476--482, 1987.

\end{thebibliography}
